\documentclass{ieeeaccess}
\usepackage{cite}
\usepackage{amsmath,amssymb,amsfonts}
\usepackage{algorithm}
\usepackage{algpseudocode}
\usepackage{graphicx}
\usepackage{textcomp}
\usepackage{enumitem}
\usepackage{url}

\usepackage{bm}
\makeatletter
\AtBeginDocument{\DeclareMathVersion{bold}
\SetSymbolFont{operators}{bold}{T1}{times}{b}{n}
\SetSymbolFont{NewLetters}{bold}{T1}{times}{b}{it}
\SetMathAlphabet{\mathrm}{bold}{T1}{times}{b}{n}
\SetMathAlphabet{\mathit}{bold}{T1}{times}{b}{it}
\SetMathAlphabet{\mathbf}{bold}{T1}{times}{b}{n}
\SetMathAlphabet{\mathtt}{bold}{OT1}{pcr}{b}{n}
\SetSymbolFont{symbols}{bold}{OMS}{cmsy}{b}{n}
\renewcommand\boldmath{\@nomath\boldmath\mathversion{bold}}}
\makeatother

\usepackage[caption=false,font=footnotesize]{subfig}
\usepackage{makecell}
\usepackage{booktabs}
\usepackage{array}
\usepackage{multirow}
\usepackage{tabularx}
\usepackage[table]{xcolor}
\usepackage{siunitx}
\def\BibTeX{{\rm B\kern-.05em{\sc i\kern-.025em b}\kern-.08em
    T\kern-.1667em\lower.7ex\hbox{E}\kern-.125emX}}

\begin{document}
\bstctlcite{IEEEexample:BSTcontrol}


\onecolumn 

\begin{description}[labelindent=0cm,leftmargin=3cm,style=multiline]

\item[\textbf{Citation}]{P. Chowdhury, M. Prabhushankar, and G. AlRegib, ``BALD-SAM: Disagreement-based Active Prompting in Interactive Segmentation,'' accepted at IEEE Access}

\item[\textbf{Review}]{First submission: 02 March 2026 (Major revision 1) \\
                        Second submission: 03 June 2026 (Major revision 2) \\
                        Third submission: 29 July 2026 (Accepted 29 October, minor revision)}

\item[\textbf{Code}]{\url{https://github.com/olivesgatech/BALD-SAM-active-prompting}}

\item[\textbf{Bib}]{@ARTICLE\{Chowdhury2026BALDSAM,\\
author=\{P. Chowdhury and M. Prabhushankar and G. AlRegib\},\\
journal=\{IEEE Access\},\\
title=\{BALD-SAM: Disagreement-based Active Prompting in Interactive Segmentation\},\\
year=\{2026\},\}\\
}

\item[\textbf{Copyright}]{\textcopyright{} Creative Commons Attribution CC BY 4.0}

\item[\textbf{Contact}]{\{pchowdhury6, alregib\}@gatech.edu \\ \url{https://alregib.ece.gatech.edu/}}

\item[\textbf{Corresponding author}]{alregib@gatech.edu}

\end{description}

\thispagestyle{empty}
\newpage
\clearpage
\setcounter{page}{1}

\twocolumn

\history{Date of publication xxxx 00, 0000, date of current version xxxx 00, 0000.}
\doi{10.1109/ACCESS.2017.DOI}

\title{BALD-SAM: Disagreement-based Active Prompting in Interactive Segmentation}
\author{\uppercase{Prithwijit Chowdhury}\authorrefmark{1}, \IEEEmembership{Student Member, IEEE},
\uppercase{Mohit Prabhushankar}\authorrefmark{1}, \IEEEmembership{Member, IEEE}, AND
\uppercase{Ghassan AlRegib}\authorrefmark{1}, \IEEEmembership{Fellow, IEEE}}
\address[1]{Omni Lab for Intelligent Visual Engineering and Science (OLIVES), School of Electrical and Computer Engineering, Georgia Institute of Technology, Atlanta, GA 30332 USA}

\tfootnote{This work was supported by the ML4Seismic Industry Partners at the Georgia Institute of Technology.}

\markboth
{P. Chowdhury \headeretal: BALD-SAM: Disagreement-based Active Prompting in Interactive Segmentation}
{P. Chowdhury \headeretal: BALD-SAM: Disagreement-based Active Prompting in Interactive Segmentation}

\corresp{Corresponding author: Ghassan AlRegib (e-mail: alregib@gatech.edu).}

\begin{abstract}

Most prompting strategies for the Segment Anything Model (SAM) treat prompt generation as a one-shot problem, yet real annotation workflows are sequential: annotators inspect the current mask and place the next prompt to resolve remaining ambiguity. We formalize this as active prompting, a spatial active learning problem in which candidate image locations form the query pool and point prompts are information-seeking queries conditioned on the evolving interaction context. We propose BALD-SAM, which adapts Bayesian Active Learning by Disagreement (BALD) to this setting. BALD-SAM keeps SAM frozen and trains a lightweight prompt-aware head whose final layer is Bayesianized via a diagonal Laplace approximation, enabling tractable epistemic uncertainty estimation without modifying SAM's pretrained weights. At each step, it queries the location with maximal mutual information between the predicted label and the model parameters. On the PointPrompt benchmark, the Laplace variant of BALD-SAM improves aggregate trajectories relative to human annotators in 8 of 14 categories and ranks in the top two on final intersection over union in 13 of 16 categories. Across five medical datasets, it achieves the best full-budget intersection over union and 95th-percentile Hausdorff distance among automatic strategies (computed from SAM's native mask) and the lowest expected calibration error and negative log-likelihood among the compared strategies (computed from the Laplace probit predictive probability). Controlled shared-seed experiments indicate that the gains reflect acquisition quality rather than trajectory compounding alone. A same-head transfer probe across three SAM generations suggests that stronger initial masks do not necessarily yield larger active-prompting gains.

\end{abstract}

\begin{keywords}
Active learning, Bayesian methods, foundation models, interactive segmentation, prompting, uncertainty quantification
\end{keywords}

\titlepgskip=-21pt

\maketitle

\section{Introduction}\label{intro}

\PARstart{I}{nteractive} image segmentation enables users to delineate objects through iterative feedback, combining human semantic understanding with computational efficiency. This paradigm has proven essential across diverse applications: medical professionals annotate anatomical structures for diagnosis and treatment planning~\cite{wang2018interactive}, geoscientists identify subsurface formations in seismic surveys~\cite{waldeland2018convolutional, quesada2025large}, ecologists track species in underwater imagery~\cite{pedersen2019detection}, and computer vision researchers create training datasets for recognition systems~\cite{lin2014microsoft}. Traditional interactive segmentation methods require domain-specific models trained on labeled data from each target domain, limiting applicability and necessitating extensive retraining for new applications. The emergence of foundation models has transformed the landscape of interactive segmentation. The Segment Anything Model (SAM)~\cite{kirillov2023segment}, trained on 11 million images and 1.1 billion masks, has demonstrated unprecedented zero-shot segmentation capabilities, enabling accurate mask generation on previously unseen images and domains without task-specific fine-tuning, through a unified promptable interface. SAM accepts spatial prompts in multiple modalities, including points, boxes, and masks, and produces high-quality segmentation outputs directly at inference time. This flexibility has catalyzed widespread adoption across medical imaging~\cite{wu2025medical} and remote sensing~\cite{ren2024segment}, together with parameter-efficient methods for adapting SAM to new tasks~\cite{wang2024sam}.

\begin{figure*}[t]
    \centering
    \includegraphics[width=1.0\textwidth]{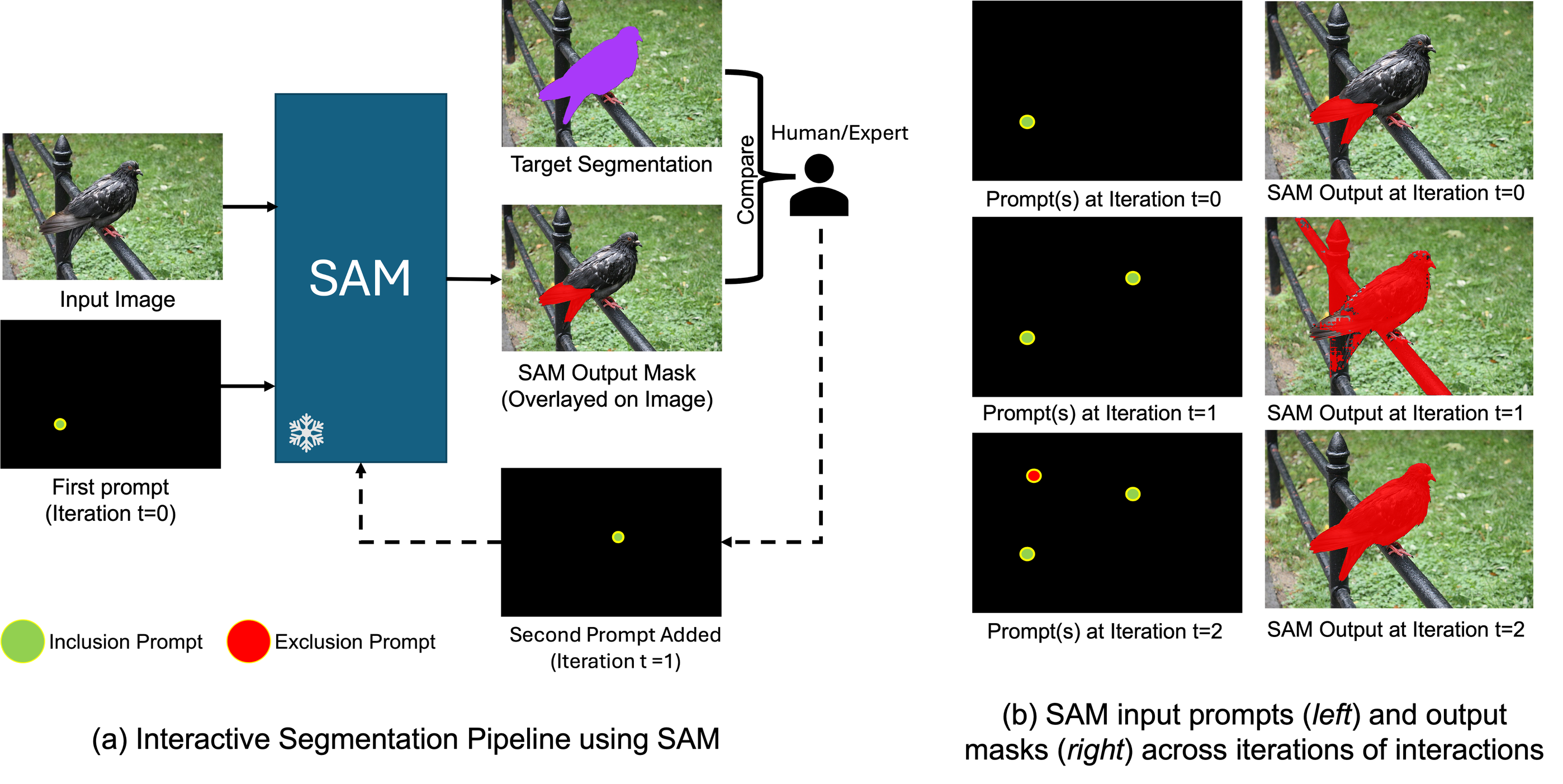}
    \caption{\textbf{Iterative prompt-based interactive segmentation using SAM.} (a) In the interactive loop, SAM receives an input image and a set of user-provided point prompts (positive/inclusion and negative/exclusion) and returns a segmentation mask. A human expert compares the predicted mask against the desired target segmentation and provides additional corrective prompts, which are fed back to SAM in the next iteration. (b) Prompt accumulation and mask evolution across iterations: the left panels show the prompt set at iterations $t=0,1,2$, and the right panels show the corresponding SAM outputs, demonstrating error correction and progressive convergence to the desired object mask.}
    \label{fig:active_prompting}
\end{figure*}

The success of promptable foundation models has naturally motivated investigation into optimal prompting strategies. Extensive research has explored automated prompt generation~\cite{liu2024grounding}, few-shot prompting techniques~\cite{gu2023systematic}, and reinforcement learning approaches for adaptive refinement~\cite{huang2024alignsam}. However, these techniques focus on automating prompting through zero-shot or minimal-prompt strategies that minimize or eliminate human involvement. This emphasis, although promising for large-scale dataset creation, fundamentally mischaracterizes the way humans use interactive segmentation systems. Humans do not generate a fixed prompt set and passively evaluate results. They observe model outputs, identify failure modes, and strategically place additional prompts to resolve ambiguities~\cite{szeto2017click}. Each prompt represents a response to the model's current understanding, creating a feedback loop where the model segments, the human evaluates, the human prompts, and the model re-segments. This cycle continues until the segmentation meets the user's quality threshold. Fig.~\ref{fig:active_prompting} illustrates an iterative SAM prompting sequence on a pigeon, where a human incrementally guides the model by adding prompts, inspecting the predicted mask, and correcting errors over successive rounds until the segmentation is satisfactory. Here, at $t=0$ a single inclusion (positive) point prompt produces an incomplete mask that captures only the pigeon's tail; at $t=1$ the user adds another inclusion prompt to encourage the full bird, but the mask overshoots and incorrectly includes background (the black railing); at $t=2$ the user adds an exclusion (negative) prompt on the wrongly included region to mark it as background, and SAM then suppresses that area and outputs a clean segmentation of the pigeon. The PointPrompt dataset~\cite{quesada2024pointprompt} is a large-scale benchmark of point-based visual prompting for interactive segmentation with SAM, created to fill the lack of publicly available datasets for systematically studying such human prompting strategies across diverse vision domains. In models like SAM, prompts are not just inputs but part of an iterative human--model dialogue, and we still do not have principled ways to characterize prompt quality in terms of how much a prompt improves the mask, reduces uncertainty, or contributes useful information to subsequent interactions. In this paper, we formalize interactive iterative prompting in SAM as active prompting.

Given a model and an unlabeled data pool, active learning asks: \textit{which examples should we query for labels to maximize model improvement under a limited annotation budget~\cite{settles2009active}?} Canonical approaches score unlabeled samples using uncertainty~\cite{gal2017deep}, diversity~\cite{sener2017active}, or hybrid criteria~\cite{benkert2023gaussian}. Importantly, even in classical pool-based active learning, informativeness is not static: after each query, the labeled set $\mathcal{D}_t$ changes, the model (or posterior) is updated, and acquisition scores must typically be recomputed for the remaining pool. We transpose this sequential selection perspective to interactive segmentation by treating candidate spatial locations within an image as the unlabeled pool~\cite{benkert2024effective} and user prompts as queries. The key insight is that not all prompts contribute equally to segmentation quality: some resolve critical ambiguities and yield substantial information gain, whereas others are redundant given the current interaction context. Adapting active learning from sample-level querying to spatial prompt selection therefore requires handling an evolving conditioning set of prompts. At iteration $t$, the model has received $\mathcal{S}_t=\{(q_1,\ell_1),\ldots,(q_t,\ell_t)\}$, and we seek the next location $q_{t+1}$ that maximizes information gain conditioned on $\mathcal{S}_t$. Unlike classical active learning where score changes are driven primarily by model updates, in active prompting the acquisition landscape can shift even with fixed parameters because the prompt set itself changes the model's conditioning context, and this must be recomputed over a vastly larger spatial candidate space at every interaction.

To address this, we propose BALD-SAM, an information-driven active prompting framework that adapts Bayesian Active Learning by Disagreement (BALD) from sample selection to spatial prompt selection in SAM. BALD-SAM is instantiated in two variants that differ only in how posterior uncertainty is approximated: BALD-SAM (Laplace), which fits a diagonal Laplace approximation over the final layer of a lightweight prompt-aware head, and BALD-SAM (MC-Dropout), which uses Monte Carlo (MC) dropout over the same head. Both variants keep SAM frozen and condition the acquisition score on the current prompt set $\mathcal{S}_t$ at every interaction step, recomputing the acquisition map after each new prompt because SAM's prediction and uncertainty both depend on the updated interaction context. The frozen SAM components provide the image representation, prompt encoding, and mask decoding, whereas the head receives the SAM logit map together with positive and negative prompt encodings. In the experiments that follow, we use BALD-SAM (Laplace) as the primary method and refer to it as BALD Laplace throughout. BALD-SAM (MC-Dropout) is evaluated as a parallel variant under the same framework, referred to as BALD-MCD. To assess whether the BALD Laplace advantage holds against a properly trained dropout baseline, we additionally evaluate BALD-MCD$^*$: a version in which the prompt-aware head is retrained with 2D spatial dropout active during both training and inference, rather than applied post-hoc to a deterministically trained head.

We evaluate BALD-SAM across the PointPrompt benchmark and five medical segmentation datasets, measuring both final segmentation quality and prompting trajectories. On PointPrompt, BALD Laplace improves aggregate prompting trajectories relative to human annotators in 8 of 14 evaluated categories and ranks among the top two methods in final intersection over union (IoU) on 13 of 16 categories. On medical data, it achieves the best full-budget IoU and 95th-percentile Hausdorff distance (HD95) (computed from SAM's native mask) and the lowest expected calibration error (ECE) and negative log-likelihood (NLL) (computed from the Laplace probit predictive probability) among automatic strategies, and is the only strategy without ground-truth (GT) access to surpass the greedy GT-error-centroid reference on multiple full-budget raw metrics. Paired tests show significant gains for IoU and HD95 against all automatic rivals, robust to multiple-comparison correction, with limited exceptions in calibration and NLL summaries against Random. A morphological strata analysis shows that BALD Laplace wins 41 of 48 stratum-metric cells, with the strongest advantages in low-contrast, small-object, moderate-boundary, and moderate-thinness conditions. A shared-seed experiment isolates the contribution of the acquisition function itself from trajectory effects and indicates that the gain is not an artifact of prompt ordering. Finally, a transfer probe across SAM generations suggests that the acquisition advantage of BALD Laplace is most pronounced when the frozen model retains meaningful epistemic uncertainty.

Our key contributions are as follows:

\begin{itemize}\item We formalize interactive iterative prompting in SAM as active prompting, where the next point prompt is selected as a prompt-conditioned information query and the acquisition score is recomputed after each interaction step.

\item We propose BALD-SAM, a framework for adapting BALD to interactive segmentation. BALD-SAM keeps SAM frozen, trains a lightweight prompt-aware head, and instantiates two practical uncertainty estimators: a diagonal Laplace approximation over the head's final layer (BALD Laplace) and Monte Carlo dropout over the same head. The MC-Dropout variant is evaluated in two forms: post-hoc dropout applied to the deterministically trained head (BALD-MCD), and a properly trained dropout head with 2D spatial dropout active during both training and inference (BALD-MCD$^*$).

\item We provide a broad evaluation across PointPrompt, medical segmentation datasets, prompt budgets, morphology strata, and baseline SAM1, SAM2, and SAM3. The comparison includes human prompts, one-shot strategies, uncertainty-based baselines, spatial coverage baselines, and a GT-based reference. BALD Laplace is the strongest automatic strategy in most settings.

\item We provide evidence that the advantage of BALD Laplace is tied to the availability of meaningful epistemic uncertainty in the frozen model, and that this advantage varies across SAM generations. This identifies active prompting as a model-dependent and interaction-dependent problem rather than a property of the acquisition rule alone.

\end{itemize}

\section{Related Work}
\label{ref}

\subsection{Segmentation and Promptable Foundation Models}

Semantic segmentation assigns class labels to every pixel and underpins dense visual understanding. Deep learning advances progressed from fully convolutional networks (FCN)~\cite{long2015fully} through U-Net~\cite{ronneberger2015u}, DeepLab~\cite{chen2017rethinking}, and PSPNet~\cite{zhao2017pyramid}, driven by benchmarks such as COCO~\cite{lin2014microsoft}, PASCAL VOC~\cite{everingham2010pascal}, and ADE20K~\cite{zhou2017scene}. Domain-specific extensions address medical imaging~\cite{isensee2021nnu,yuan2021multi}, seismic interpretation~\cite{wu2019faultseg3d,wu2020building,quesada2025large}, and remote sensing~\cite{kemker2018algorithms,maggiori2016convolutional}, each introducing challenges from limited labels, noise, and multi-scale structure. Despite strong in-domain performance, conventional segmentation remains data-hungry and poorly transferable, motivating interactive and promptable alternatives.

Foundation models address these limitations through task-agnostic pretraining with flexible prompt-based adaptation. Several systems unify segmentation modalities: SEEM~\cite{zou2023segment} supports points, boxes, scribbles, and text via a shared visual-semantic space; Semantic-SAM~\cite{li2024segment} adds granularity control; and SegGPT~\cite{wang2023seggpt} formulates segmentation as in-context learning. The Segment Anything Model (SAM1)~\cite{kirillov2023segment} established a widely adopted promptable segmentation interface for images. SAM1 comprises a vision transformer (ViT) image encoder~\cite{dosovitskiy2020image}, a prompt encoder for sparse prompts such as points and boxes and dense prompts such as masks, and a mask decoder that fuses image and prompt embeddings via cross-attention. Trained on the SA-1B dataset with 11M images and 1.1B masks, SAM1 exhibits strong zero-shot generalization and has been adapted to medical imaging~\cite{ma2024segment,zhang2023customized,wu2025medical,huang2024segment} and remote sensing~\cite{ren2024segment,zhang2024rs}, domains where task-specific networks, such as encoder--decoder models for seismic fault detection~\cite{li2019seismic}, were previously required.

Subsequent versions expanded this promptable segmentation paradigm. SAM2~\cite{ravi2024sam} extends SAM from images to videos through a streaming-memory design, enabling promptable object segmentation and tracking across temporal sequences. SAM3~\cite{carion2025sam} further broadens the interface toward concept-level segmentation, supporting language prompts, exemplar prompts, and visual prompts for detecting, segmenting, and tracking object concepts across images and videos. Together, SAM1, SAM2, and SAM3 mark a shift from fixed-task segmentation models toward general-purpose, promptable segmentation systems.

Iterative human--model interaction has driven major advances in language models through chain-of-thought prompting~\cite{wei2022chain}, in-context learning~\cite{brown2020language}, and reinforcement learning from human feedback (RLHF)~\cite{ouyang2022training}, where corrective feedback loops progressively refine outputs. Visual and multimodal models similarly benefit from iterative refinement in reasoning~\cite{rose2023visual}, instruction-based editing~\cite{brooks2023instructpix2pix}, and active example selection~\cite{diao2024active}. However, the interactive segmentation literature has not systematically adopted this perspective; existing SAM research emphasizes automation over dialogue and one-shot performance over iterative convergence. Our work bridges this gap by bringing active learning principles and iterative refinement insights into interactive segmentation, establishing a framework for human--model collaborative annotation.

\subsection{Automated and One-Shot Prompting Strategies in SAM}

SAM's interactive design has prompted extensive study focused primarily on automation and efficiency through reduced human involvement. Automated prompting and personalization methods include PerSAM~\cite{zhang2023personalize} for one-shot instance transfer and Grounded-SAM~\cite{ren2024grounded} for open-vocabulary detection followed by SAM-based mask generation. These approaches aim to synthesize effective prompts directly from images or text descriptions, bypassing iterative human refinement. In sparse prompting regimes, work has learned to generate point and box prompts automatically~\cite{liu2023matcher, chen2024rsprompter}, box prompting as a higher-information alternative to points~\cite{zhang2023segment}, and hybrid prompt combinations for robustness~\cite{mazurowski2023segment}. Sequential decision-making has been explored through reinforcement learning for prompt selection~\cite{huang2024alignsam}, and domain-specific fine-tuning has adapted SAM to targets such as colon polyps~\cite{li2024polyp}; however, these methods optimize prompts or weights offline rather than modeling real human feedback loops. Empirical studies further evaluate SAM's robustness to input perturbations~\cite{wang2024empirical} and adverse weather conditions~\cite{shan2023robustness}, as well as its zero-shot performance on polyp segmentation~\cite{zhou2023can}. While these efforts have advanced automated prompting, they largely treat prompting as a single-pass process rather than an interactive dialogue, and have not studied how to actively select prompts that maximize information gain during iterative human--model interaction.

\subsection{Active Learning and BALD}
\label{ref:bald}
Active learning studies how to choose the most informative queries so that a model improves with minimal annotation effort~\cite{settles2009active}. In the classical pool-based setting, the query strategy selects unlabeled examples for annotation using criteria such as margin sampling~\cite{scheffer2001active}, predictive entropy~\cite{shannon1948mathematical}, query-by-committee disagreement~\cite{seung1992query}, stable outputs~\cite{benkert2024targeting}, gradient-based criteria~\cite{ash2019deep, prabhushankar2022introspective}, or prediction switches~\cite{ash2019deep,benkert2023gaussian}. These strategies have been extended from images to videos~\cite{kokilepersaud2023focal} and clinical trial settings~\cite{fowler2023clinical}. These strategies typically rely on predictive uncertainty alone and may conflate epistemic uncertainty (model uncertainty due to limited knowledge) with aleatoric uncertainty (irreducible ambiguity in the data).

Bayesian Active Learning by Disagreement (BALD) addresses this by explicitly targeting epistemic uncertainty~\cite{houlsby2011bayesian}. BALD selects the query $x$ that maximizes the mutual information between the prediction $y$ and the model parameters $\bm{\theta}$ under the current posterior:
\begin{equation}
\begin{aligned}
\mathrm{BALD}(x)
&= I(y,\bm{\theta} \mid x,\mathcal{D})\\
&= H\!\left[y \mid x,\mathcal{D}\right]
- \mathbb{E}_{p(\bm{\theta} \mid \mathcal{D})}\!\left[ H\!\left[y \mid x,\bm{\theta}\right] \right].
\end{aligned}
\end{equation}
Here, $H[y \mid x,\mathcal{D}]$ is the predictive entropy under the posterior (total uncertainty), whereas $\mathbb{E}_{p(\bm{\theta} \mid \mathcal{D})}[H[y \mid x,\bm{\theta}]]$ is the expected entropy of a model sampled from the posterior (data ambiguity). Their difference isolates uncertainty caused by disagreement among plausible models, i.e., the uncertainty that can be reduced by acquiring a new label. Intuitively, BALD prioritizes queries for which different plausible models make different predictions, since labeling those examples is expected to yield the greatest information gain.

Applying BALD in deep networks requires approximate Bayesian inference. Common practical approaches include Monte Carlo dropout~\cite{gal2016dropout}, deep ensembles~\cite{beluch2018power}, and Laplace approximation around a trained solution~\cite{ritter2018scalable}. These methods have enabled Bayesian active learning in image classification~\cite{gal2017deep}, semantic segmentation~\cite{mackowiak2018cereals}, object detection~\cite{kao2018localization}, and medical imaging~\cite{yang2017suggestive}.

Prior active learning works focus on selecting images to label for supervised learning. In contrast, interactive segmentation requires selecting where within an image to query next, conditioned on an evolving set of user prompts. This spatial, sequential setting is fundamentally different from classical sample selection. Recent SAM-related work has focused on efficiently adapting SAM's parameters~\cite{wang2024sam} rather than on uncertainty-guided prompt selection, and, to our knowledge, prompt selection has not been formulated with a principled BALD objective or a Bayesian posterior. Our work builds on BALD to define a theoretically grounded criterion for active spatial prompt selection in interactive segmentation.

\subsection{PointPrompt Dataset}

PointPrompt~\cite{quesada2024pointprompt} is a large-scale benchmark of point-based visual prompting for interactive segmentation with SAM, created to fill the lack of publicly available datasets for systematically studying prompting strategies across diverse vision domains. PointPrompt contains 6,400 curated images organized into 16 datasets (400 image-mask pairs each) spanning four domains: natural images, underwater imagery, medical imaging, and seismic data. Natural images include nine COCO categories~\cite{lin2014microsoft} (dog, cat, bird, clock, bus, baseball bat, cow, tie, stop sign), covering rigid objects and highly deformable classes. The underwater domain includes two NDD20 subsets~\cite{trotter2020ndd20} capturing dolphins above and below water, with challenges from lighting variation, turbidity, and motion blur. Medical datasets include BUSI~\cite{al2020dataset} (breast ultrasound tumors), Kvasir-SEG~\cite{jha2020kvasir} (colonoscopy polyps), and ISIC~\cite{codella2019skin} (skin lesions), reflecting low-contrast boundaries, class imbalance, and clinically critical boundary precision. Seismic data are drawn from the F3 Facies dataset~\cite{alaudah2019machine} (salt dome and chalk), introducing strong domain shift from SAM's training distribution and difficulties from noise, low signal-to-noise ratio (SNR), 3D structure projected into 2D slices, and interpretation ambiguity. Prompts were collected using a SAM-based interactive annotation tool where annotators iteratively clicked inclusion points (green) and exclusion points (red), matching SAM's native interface. For each image, annotators continued prompting until achieving satisfactory segmentation under subjective judgment. The dataset records full prompting sequences, including point coordinates, intermediate SAM1 masks after each prompt, and IoU against the ground-truth mask. Multiple annotators per image enable analysis of inter-annotator variability and diverse human strategies, providing an empirical basis for studying what constitutes effective visual prompts and informing automated prompt generation. We use the PointPrompt training split for head training (Section~\ref{bay_train}) and Laplace fitting (Section~\ref{laplace}). In the primary experiments (Section~\ref{primary}), we compare BALD-SAM against the human annotator trajectories recorded in the dataset. The dataset is partitioned with 70\% for training, 15\% for validation, and 15\% for testing. To ensure consistent evaluation across experiments, we use a fixed random seed (seed=42) and maintain the same train/validation/test split throughout all experiments. Sample indices for each split are recorded and reused to eliminate variance from data partitioning.

\section{Active Prompting in Interactive Segmentation}

At iteration $t$, given the current prompt set
$\mathcal{S}_t = \{(q_1, \ell_1), \ldots, (q_t, \ell_t)\}$, where $q_i$ is a spatial location and $\ell_i \in \{0,1\}$ is an inclusion/exclusion label, a selection strategy $\pi$ assigns an informativeness score to each candidate location:
\begin{equation}
    s_q = \pi(q \mid I, \mathcal{S}_t, \bm{\theta}), \quad q_{t+1} = \arg\max_{q \in \Omega} s_q
\end{equation}
where $I$ is the input image, $\Omega$ is the set of candidate locations, and $\bm{\theta}$ denotes model parameters. Once $q_{t+1}$ is selected, its inclusion/exclusion label is assigned by reading the ground-truth mask: $\ell_{t+1} = M^*[q_{t+1}]$, with $\ell{=}1$ for foreground and $\ell{=}0$ for background. This follows the standard active learning evaluation convention in which the annotator is simulated by the ground-truth label. The scope of active prompting is \textit{location selection}: the strategy $\pi$ decides \textit{where} to prompt; the inclusion/exclusion label is determined externally and is identical across all compared strategies. The prompt history is then updated as
$\mathcal{S}_{t+1} = \mathcal{S}_t \cup \{(q_{t+1}, \ell_{t+1})\}$, and the process repeats. The full iterative procedure is summarized in Algorithm~\ref{alg:active_prompting}.

This perspective differs from standard SAM usage in three ways: (i) prompt placement is treated as query optimization rather than ad hoc interaction, (ii) selection is driven by quantitative informativeness scores rather than visual inspection alone, and (iii) each query is explicitly conditioned on the evolving prompt history.

\subsection{Active Prompting Workflow}

We formalize the active prompting loop in Algorithm~\ref{alg:active_prompting}. Starting from an optional seed prompt set $\mathcal{S}_0$, the method alternates between (i) scoring candidate locations using the selection strategy $\pi$, (ii) assigning the inclusion/exclusion label at the highest-scoring location by reading the ground-truth mask $M^*$, and (iii) updating the segmentation conditioned on the expanded prompt set. The loop terminates when a stopping criterion is met (e.g., a prompt budget, convergence of the mask, or user satisfaction). When multiple human annotators are available for an image, their IoU trajectories are aggregated by averaging per-image IoU values across annotators at each prompt step. For annotators who stopped before the maximum evaluation budget, the IoU is held constant at its terminal value for all remaining steps.

\begin{algorithm}[t]
\caption{Active Prompting for Interactive Segmentation}
\label{alg:active_prompting}
\begin{algorithmic}[1]
\Require Image $I$, selection strategy $\pi$, candidate set $\Omega$, stopping criterion $\mathcal{C}$, model parameters $\bm{\theta}$
\Require Optional seed prompts $\mathcal{S}_0$ (default: $\emptyset$)
\Ensure Final prompt set $\mathcal{S}_T$ and segmentation mask $\hat{M}_T$

\State $t \gets 0$
\State $\mathcal{S}_t \gets \mathcal{S}_0$
\State Generate initial segmentation $\hat{M}_t$ from $I$ and $\mathcal{S}_t$

\While{$\neg \mathcal{C}(\hat{M}_t, \mathcal{S}_t, t)$}
    \ForAll{$q \in \Omega$}
        \State $s_q \gets \pi(q \mid I, \mathcal{S}_t, \bm{\theta})$
    \EndFor

    \State $q_{t+1} \gets \arg\max_{q \in \Omega} s_q$
    \State $\ell_{t+1} \gets M^*[q_{t+1}]$ \Comment{GT label: $1$ = inclusion, $0$ = exclusion}
    \State $\mathcal{S}_{t+1} \gets \mathcal{S}_t \cup \{(q_{t+1}, \ell_{t+1})\}$
    \State Generate updated segmentation $\hat{M}_{t+1}$ from $I$ and $\mathcal{S}_{t+1}$
    \State $t \gets t + 1$
\EndWhile

\State \Return $\mathcal{S}_t, \hat{M}_t$
\end{algorithmic}
\end{algorithm}

\subsection{Why Active Prompting Helps}

Algorithm~\ref{alg:active_prompting} highlights that prompt selection is not a one-shot decision: informativeness scores are recomputed after every user interaction, and each new query is conditioned on the updated prompt history. This sequential, model-aware loop provides several practical advantages over intuition-driven prompting:
\begin{itemize}
    \item \emph{Principled query selection:} prompts are chosen using explicit, quantitative criteria (e.g., uncertainty, diversity, or hybrid strategies).
    \item \emph{Lower cognitive burden:} annotators no longer need to scan the entire image for failure regions.
    \item \emph{Better spatial coverage:} informative locations are explored systematically rather than based on human visual bias.
    \item \emph{Model-aware adaptation:} query locations adapt to the model's evolving uncertainty as new prompts are added.
    \item \emph{Cross-domain applicability:} the framework depends on uncertainty and informativeness, not domain-specific semantics.
\end{itemize}

In the next section, we instantiate this framework using BALD with a Laplace-approximated Bayesian head on top of frozen SAM1 features, yielding a tractable and effective active prompting strategy across diverse imaging domains.

\section{BALD-SAM: Information-Driven Active Prompt Sampling}
\label{method}

We build our method on Bayesian Active Learning by Disagreement (BALD)~\cite{houlsby2011bayesian}, which selects queries that maximize mutual information between an unknown label and uncertain model parameters under the current posterior. Intuitively, BALD favors queries where plausible models disagree most, since those queries are expected to reduce epistemic uncertainty. See Section~\ref{ref:bald} for background.

This introduces two practical challenges. First, query informativeness must be recomputed after every interaction because the acquisition score is conditioned on the evolving prompt set $\mathcal{S}_t$. As new prompts are added, the interaction context changes, so each candidate location must be reassessed. Second, BALD requires parameter uncertainty, yet full posterior inference is intractable for SAM-scale models with hundreds of millions of parameters. Direct Bayesian treatment of the full network would be prohibitively expensive, and modifying SAM's architecture could disrupt the pretrained representations and zero-shot behavior that make it effective.

To address this, we freeze SAM and fit uncertainty only over the final layer of a lightweight prompt-aware segmentation head. The full head is trained normally, and an additional diagonal Laplace approximation is fitted after training over the final $1\times 1$ convolution. This makes prompt-conditioned mutual information tractable while preserving SAM's pretrained representations and zero-shot capability.

\begin{figure*}[t]
    \centering
    \includegraphics[width=1.0\textwidth]{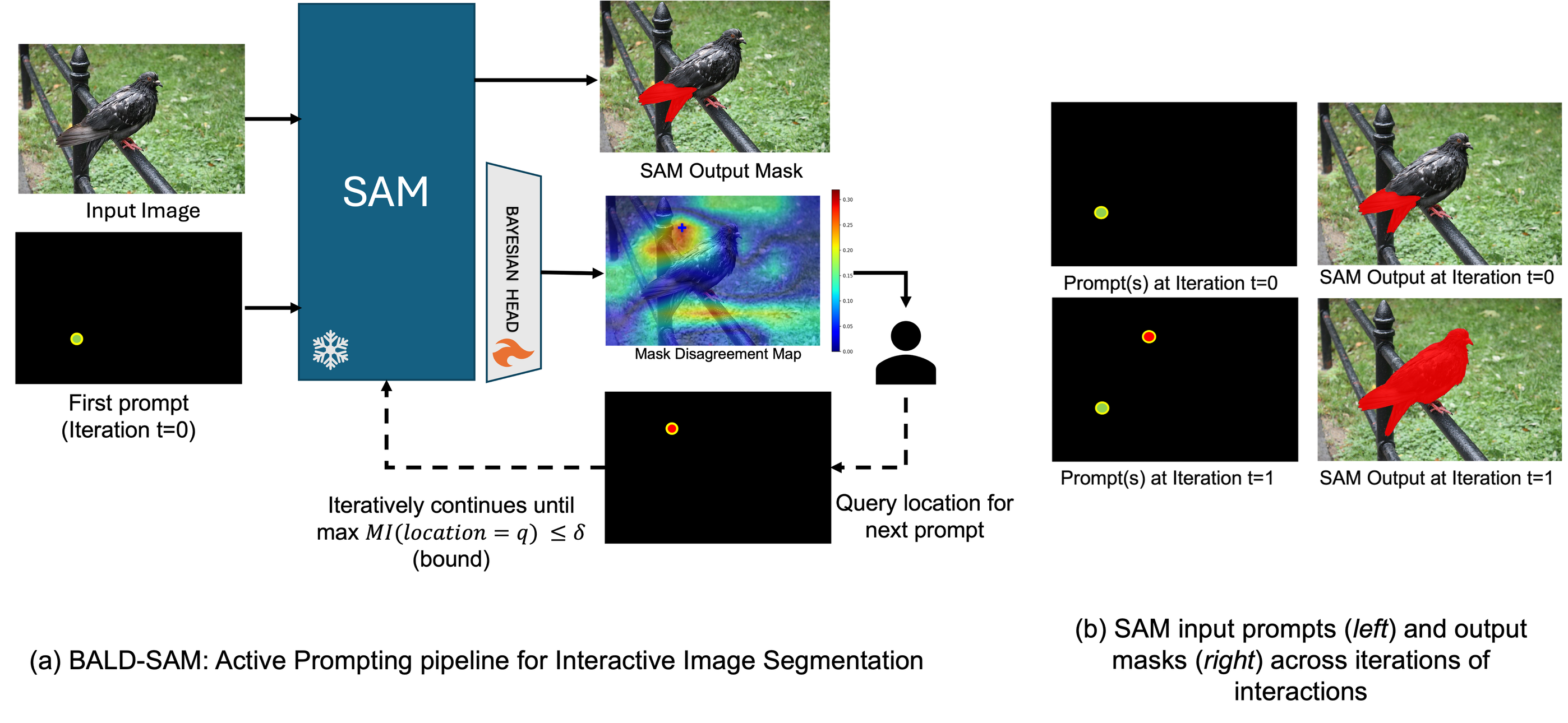}
    \caption{BALD-SAM active prompt sampling. At iteration $t$, the image $\mathcal{I}$ and current prompt set $\mathcal{S}_t$ are processed by frozen SAM components and a prompt-aware segmentation head whose final layer is sampled from a fitted Laplace posterior. Multiple posterior samples produce an ensemble of mask probability maps, from which we compute a disagreement, or mutual-information, map. The location with the highest BALD score is selected next; its inclusion/exclusion label is then read from the ground-truth mask $M^*$ (following standard active learning evaluation protocol), and the prompt set is updated. The method selects \emph{where} to prompt; label assignment is external to its scope.}
    \label{fig:bald_pipeline}
\end{figure*}

An overview of the full pipeline is illustrated in Fig.~\ref{fig:bald_pipeline}.

\subsection{Adapting BALD to Interactive Segmentation}

Standard BALD selects samples from an unlabeled dataset. Here, we select locations within a single image, conditioned on an evolving prompt set.

\subsubsection{Prompt-Conditioned Sequential Queries}

Let $M^* \in \{0,1\}^{H\times W}$ be the unknown ground-truth mask for image $\mathcal{I}$. For a candidate location $q \in \Omega_t$, define the unknown queried label as
\begin{equation}
\ell_q := M^*[q] \in \{0,1\},
\end{equation}
where $\Omega_t$ is the set of valid candidate pixels at prompting step $t$. Concretely, $\Omega_t$ consists of all pixels of the $256{\times}256$ head output, excluding a disk of radius 5 pixels around each prompt already in $\mathcal{S}_t$; selected locations are rescaled to the original image resolution before being passed to SAM. Because the head is fully convolutional, the mutual information (MI) scores for all candidates are obtained in a single batched pass per posterior sample. Given a training set $\mathcal{D}$, the next query is
\begin{equation}
q_{t+1}
=
\arg\max_{q \in \Omega_t}
I(\ell_q; \bm{\omega} \mid \mathcal{I}, \mathcal{S}_t, \mathcal{D}),
\end{equation}
where $\bm{\omega}$ denotes the uncertain final-layer parameters of the prompt-aware segmentation head.

This objective is prompt-conditioned. After each user response, $\mathcal{S}_t$ changes, so the predictive uncertainty and BALD scores must be recomputed as described in Section~\ref{sec:compute_bald}. The iterative query, label, and update process is depicted in Fig.~\ref{fig:bald_pipeline}.

\subsubsection{Partial Posterior for the Prompt-Aware Head}

Let
\begin{equation}
\bm{\theta} = \{\bm{\theta}_{\text{SAM}}, \bm{\theta}_{\text{head}}\},
\qquad
\bm{\theta}_{\text{head}}=\{\bm{\eta},\bm{\omega}\},
\end{equation}
where $\bm{\theta}_{\text{SAM}}$ denotes the frozen SAM parameters, $\bm{\eta}$ denotes the deterministic parameters of the segmentation head before the final layer, and $\bm{\omega}=(\mathbf{w},b)$ denotes the final-layer weights and bias. In our implementation, only $\bm{\omega}$ is Bayesianized. If the hidden feature map has $C$ channels, then the final $1\times 1$ convolution has $C+1$ parameters, including the bias. With $C=16$, this gives 17 Bayesian parameters.

The posterior factorization used for acquisition is
\begin{equation}
p(\bm{\theta} \mid \mathcal{D})
=
\delta(\bm{\theta}_{\text{SAM}}-\bm{\theta}_{\text{SAM}}^*)
\,
\delta(\bm{\eta}-\hat{\bm{\eta}})
\,
p(\bm{\omega}\mid\mathcal{D}),
\end{equation}
where $\bm{\theta}_{\text{SAM}}^*$ denotes the pretrained SAM weights, $\hat{\bm{\eta}}$ denotes the trained deterministic head parameters before the final layer, and $\delta(\cdot)$ is the Dirac delta. Thus, no posterior uncertainty is modeled over SAM or over the earlier layers of the segmentation head.

After training the head, we fit a diagonal Laplace approximation over $\bm{\omega}$:
\begin{equation}
p(\bm{\omega}\mid\mathcal{D})
\approx
\mathcal{N}(\hat{\bm{\omega}}, \bm{\Sigma}_{\text{Lap}}),
\end{equation}
where $\hat{\bm{\omega}}$ is the trained final-layer parameter vector and $\bm{\Sigma}_{\text{Lap}}$ is a diagonal covariance estimated from a curvature approximation after training. We use this Laplace approximation to represent uncertainty over the final layer while keeping all other model components fixed.

A natural question is why 17 global scalar parameters can produce spatially non-uniform epistemic disagreement. The final layer computes $f_{\bm{\omega}}(q) = \mathbf{w}^\top \mathbf{F}_t(q) + b = \bm{\omega}^\top \bm{\phi}_t(q)$, where $\bm{\omega}=(\mathbf{w},b) \in \mathbb{R}^{17}$ are the uncertain Bayesian parameters, $\mathbf{F}_t(q) \in \mathbb{R}^{16}$ is the feature vector at pixel $q$ produced by the deterministic backbone, and $\bm{\phi}_t(q) = [\mathbf{F}_t(q)^\top, 1]^\top \in \mathbb{R}^{17}$ is its bias-augmented version. Under the diagonal Laplace posterior, the predictive logit variance at location $q$ is
\begin{equation}
\mathrm{Var}_{\bm{\omega}}[f_{\bm{\omega}}(q)]
= \bm{\phi}_t(q)^\top \bm{\Sigma}_{\text{Lap}}\, \bm{\phi}_t(q),
\end{equation}
which varies across spatial locations because $\mathbf{F}_t(q)$ varies spatially. The same uncertainty in 17 scalars therefore induces a different epistemic signal at every candidate location, driven by the diversity of backbone feature representations across the image. This is structurally analogous to linear regression with uncertain weights: prediction intervals are spatially non-uniform because input features vary spatially, even though the uncertain parameter vector is global.

For prediction, we use a Laplace predictive correction to account for final-layer uncertainty. Let $f_{\hat{\bm{\omega}}}(q)$ be the logit at pixel $q$ under the trained final layer, and let $\mathrm{Var}_{\text{Lap}}[f(q)]$ be the corresponding predictive logit variance induced by the diagonal Laplace posterior. We compute
\begin{equation}
\kappa(q)
=
\frac{1}{\sqrt{1+\frac{\pi}{8}\mathrm{Var}_{\text{Lap}}[f(q)]}},
\quad
p_{\text{pred}}(q)
=
\sigma\!\left(\kappa(q) f_{\hat{\bm{\omega}}}(q)\right),
\end{equation}
where $\sigma(\cdot)$ is the sigmoid function. For BALD acquisition, we use a separate diagonal covariance $\bm{\Sigma}_{\text{BALD}}$ so that the acquisition map preserves epistemic disagreement.

\subsection{Disagreement-Based Sampling via BALD}

\subsubsection{Prompt-Aware Segmentation Head}

We freeze SAM's image encoder, prompt encoder, and mask decoder. At prompting step $t$, SAM produces a logit map for image $\mathcal{I}$ and prompt set $\mathcal{S}_t$. The refinement head takes three input channels:
\begin{equation}
\mathbf{X}_t = [L^{\mathrm{SAM}}_t, P^+_t, P^-_t] \in \mathbb{R}^{H\times W\times 3},
\end{equation}
where $L^{\mathrm{SAM}}_t$ is the SAM logit map, and $P^+_t$ and $P^-_t$ encode the positive and negative prompts in $\mathcal{S}_t$.

The head first maps $\mathbf{X}_t$ to a hidden feature map
\begin{equation}
\mathbf{F}_t = g_{\bm{\eta}}(\mathbf{X}_t) \in \mathbb{R}^{H\times W\times C},
\end{equation}
and then applies a final $1\times 1$ convolution to produce a logit at each pixel:
\begin{equation}
f_{\bm{\omega}}(q) = \mathbf{w}^\top \mathbf{F}_t(q) + b,
\qquad
p_{\bm{\omega}}(q) = \sigma(f_{\bm{\omega}}(q)).
\end{equation}
Here, $q\in\Omega_t$ is a valid candidate pixel, $\bm{\omega}=(\mathbf{w},b)$ denotes the final-layer parameters, and $p_{\bm{\omega}}(q)$ is the predicted foreground probability at $q$. In our implementation, $g_{\bm{\eta}}$ is a small no-image convolutional network with two $3\times3$ convolutional layers of hidden width 16 and ReLU activations, with no dropout or batch normalization; the final layer is the only Bayesianized component.

We train the head on a dataset
\begin{equation}
\mathcal{D} = \{(\mathcal{I}^n, \mathcal{S}^n, M^{*n})\}_{n=1}^{N},
\end{equation}
where $\mathcal{I}^n$ is an image, $\mathcal{S}^n$ is a prompt set for that image, and $M^{*n}$ is its ground-truth mask. The head parameters are learned by minimizing a combined binary cross-entropy (BCE) and Dice objective:
\begin{equation}
(\hat{\bm{\eta}},\hat{\bm{\omega}})
=
\arg\min_{\bm{\eta},\bm{\omega}}
\frac{1}{N}\sum_{n=1}^{N}
\mathcal{L}^{n},
\end{equation}
with the per-example loss
\begin{equation}
\mathcal{L}^{n}
=
\mathcal{L}_{\mathrm{BCE}}^{\mathrm{pos}}\!\left(p_{\bm{\omega}}^n, M^{*n}\right)
+ \lambda_{\mathrm{Dice}}\,
\mathcal{L}_{\mathrm{Dice}}\!\left(p_{\bm{\omega}}^n, M^{*n}\right),
\end{equation}
where $\lambda_{\mathrm{Dice}}=1.0$, $\mathcal{L}_{\mathrm{BCE}}^{\mathrm{pos}}$ uses a class-imbalance positive weight computed from the training masks, and $p_{\bm{\omega}}^n$ is the foreground probability map predicted by the head for example $n$.

For a test image $\mathcal{I}$ and current prompts $\mathcal{S}_t$, the posterior predictive distribution over probability maps $P \in [0,1]^{H\times W}$ is
\begin{equation}
p(P \mid \mathcal{I}, \mathcal{S}_t, \mathcal{D})
=
\int
p(P \mid \bm{\theta}_{\text{SAM}}^*, \hat{\bm{\eta}}, \bm{\omega}, \mathcal{I}, \mathcal{S}_t)
p(\bm{\omega} \mid \mathcal{D})
\, d\bm{\omega} .
\end{equation}

For acquisition, we approximate this integral with Monte Carlo sampling. We draw $K$ final-layer samples
\begin{equation}
\bm{\omega}_k \sim \mathcal{N}(\hat{\bm{\omega}}, \bm{\Sigma}_{\text{BALD}}),
\qquad k=1,\ldots,K,
\end{equation}
and compute the corresponding probability maps
\begin{equation}
P_{\bm{\omega}_k} \in [0,1]^{H\times W}.
\end{equation}
Disagreement among these $K$ maps captures epistemic uncertainty. This ensemble disagreement is visualized as the Mask Disagreement Map shown in Fig.~\ref{fig:bald_pipeline}.

\subsubsection{Computing BALD Mutual Information}
\label{sec:compute_bald}

For each candidate location $q \in \Omega_t$, define the predictive probability under final-layer sample $\bm{\omega}_k$ as
\begin{equation}
p_k(q)
:=
p(\ell_q = 1 \mid \mathcal{I}, \mathcal{S}_t, \bm{\omega}_k)
=
P_{\bm{\omega}_k}[q].
\end{equation}
The Monte Carlo predictive mean is
\begin{equation}
\bar{p}(q) = \frac{1}{K}\sum_{k=1}^{K} p_k(q).
\end{equation}

Let $h_2(p) = -p\log p - (1-p)\log(1-p)$ denote the binary entropy function. The BALD score at location $q$ is
\begin{equation}
\mathrm{MI}_{\text{Lap}}(q)
=
h_2(\bar{p}(q))
-
\frac{1}{K}\sum_{k=1}^{K} h_2(p_k(q)).
\end{equation}
The first term is the entropy of the mean prediction, and the second term is the average entropy of the sampled predictions. Their difference is high when sampled final-layer parameters disagree at location $q$.

We select the next query as
\begin{equation}
q_{t+1} = \arg\max_{q \in \Omega_t} \mathrm{MI}_{\text{Lap}}(q),
\end{equation}
and assign its inclusion/exclusion label by reading the ground-truth mask directly:
\begin{equation}
\ell_{t+1} = M^*[q_{t+1}] \in \{0,1\},
\end{equation}
where $\ell_{t+1} = 1$ denotes an inclusion (foreground) prompt and $\ell_{t+1} = 0$ denotes an exclusion (background) prompt. This follows the standard active learning evaluation convention in which the annotator is simulated by the ground-truth label. The scope of the method is location selection only: BALD-SAM decides where to place the next prompt; the inclusion/exclusion assignment is determined externally from $M^*$ and is identical for all strategies compared in this work. The prompt set is then updated as
\begin{equation}
\mathcal{S}_{t+1}
=
\mathcal{S}_t \cup \{(q_{t+1}, \ell_{t+1})\}.
\end{equation}

This query-and-update cycle is depicted in Fig.~\ref{fig:bald_pipeline}. The selected location $q_{t+1}$ is the only output of the acquisition function; its label $\ell_{t+1}$ is read from $M^*$ and folded back into $\mathcal{S}_{t+1}$ identically across all compared strategies.

\subsubsection{Stopping Criteria}

We terminate prompting when any one of the following conditions is met.

\paragraph{Global Entropy Threshold.}
Define the total predictive entropy over candidate locations as
\begin{equation}
H_{\text{total}} := \sum_{q \in \Omega_t} h_2(\bar{p}(q)).
\end{equation}
If
\begin{equation}
H_{\text{total}} \le \tau_{\text{ent}},
\end{equation}
where $\tau_{\text{ent}}$ is a preset entropy threshold, the model is deemed sufficiently certain overall.

\paragraph{Maximum Mutual-Information Threshold.}
If
\begin{equation}
\max_{q \in \Omega_t} \mathrm{MI}_{\text{Lap}}(q) \le \tau_{\text{MI}},
\end{equation}
where $\tau_{\text{MI}}$ is a preset information-gain threshold, no remaining candidate location is expected to yield substantial additional information. This condition corresponds to the convergence bound $\max\,\mathrm{MI}(q) \leq \delta$ annotated in Fig.~\ref{fig:bald_pipeline}.

\paragraph{Maximum Prompt Budget.}
\label{prompt_budget}
We additionally impose a hard cap of 20 prompts as a practical stopping criterion for all strategies. This prevents excessively long interaction sequences that may drive SAM beyond the prompting regime encountered during pretraining and helps preserve stable behavior.

\section{Experimental Setup}

To train the Bayesian prompt-aware head and fit the Laplace approximation, we use only the training images from the PointPrompt dataset. Importantly, we do not use human prompt trajectories during training. Instead, the head is trained with synthetically generated prompts only. This design prevents the method from learning annotator-specific prompting patterns and ensures that the comparison against human test annotators is not confounded by exposure to human prompt behavior. As a result, performance gains cannot be attributed to imitation of the human prompting policy.

\subsubsection{Bayesian Head Training}
\label{bay_train}

In our experiments, the Bayesian component is a lightweight \textit{no-image} refinement head trained on frozen SAM1 outputs rather than on raw images. The head takes a three-channel input consisting of the SAM1 logit map, a positive prompt map, and a negative prompt map, and predicts a refined binary logit mask. Architecturally, it uses two $3{\times}3$ convolutional layers with hidden width 16 and ReLU activations, followed by a final $1{\times}1$ convolution that produces the output logits. No dropout or batch normalization is used. Because the SAM1 outputs are precomputed and cached before head training, all SAM1 components remain fixed and only the head parameters are optimized.

Training uses AdamW with learning rate $10^{-3}$, weight decay $10^{-4}$, batch size 64, and cosine learning-rate decay. The loss is a combined BCE--Dice objective with Dice weight 1.0 and a class-imbalance weighting term computed from the training masks. Training runs for up to 50 epochs with early stopping based on validation IoU using patience 10. All inputs and targets are resized to $256 \times 256$, and the implementation uses mixed-precision training.

To expose the head to diverse prompt-conditioned SAM1 states, the training set is expanded by generating multiple synthetic prompt sets per sample. For each image, the pipeline creates 10 prompt sets for training and 4 each for validation and test. In each prompt set, the number of positive--negative point pairs is sampled uniformly from 1 to 10. Positive prompts are sampled from foreground pixels of the ground-truth mask and negative prompts are sampled from background pixels, without replacement when possible and with replacement only when a region is too small. These sampled points are then converted into Gaussian prompt maps and paired with the corresponding frozen SAM1 logits.

\subsubsection{Backbone Selection}
We first verify that the ViT-H backbone dominates smaller variants before committing to it for all subsequent ablations. Table~\ref{tab:backbone_dominance} compares ViT-H, ViT-B, and ViT-Tiny across the same hyperparameter grid. ViT-H achieves the highest IoU in every configuration, its worst setting (IoU=$0.120$) is close to ViT-B's best ($0.121$), and its mean across all configurations ($0.141$) exceeds ViT-B by $+0.029$ and ViT-Tiny by $+0.082$. The performance gap is consistent rather than concentrated in a few outlier settings, suggesting that the richer representations from the 632M-parameter encoder yield better-calibrated Bayesian posteriors. All remaining experiments use ViT-H.

\subsubsection{Laplace Posterior Ablation}
\label{laplace}

After deterministic head training, we fit a diagonal Laplace approximation only over the final prediction layer of the no-image head, i.e., the final $1{\times}1$ convolution, while keeping the preceding convolutional backbone fixed. This uncertainty model defines our BALD Laplace method: BALD acquisition is computed by sampling final-layer parameters from the fitted Laplace posterior and selecting points that maximize epistemic disagreement. This Bayesian layer contains 17 parameters in total, consisting of 16 weights and 1 bias. The Laplace approximation is estimated from a subset of the training data by accumulating a diagonal generalized Gauss--Newton approximation to the curvature.

For the NLL evaluation posterior, prior precision is selected by grid search on validation NLL over the grid $\{10^{-4}, 10^{-3}, 10^{-2}, 10^{-1}, 1.0, 10.0\}$ using the PointPrompt validation split (15\% of 6,400 images). For BALD acquisition, a separate prior precision is chosen independently from the same grid to keep the mean predictive logit variance over the validation candidate pool within the range $[0.5, 2.0]$ in logit space, ensuring the acquisition map is neither uninformatively flat (too-sharp posterior) nor uniformly saturated (too-diffuse posterior). The selected BALD acquisition prior precision is $10^{-3}$.

Table~\ref{tab:laplace_samples_interaction} ablates the two key controls of this approximation: the Laplace subset size, which determines the number of datapoints used to estimate the diagonal curvature, and the posterior sample count, which determines the number of Monte Carlo draws per image. Two trends emerge. First, subset size determines the performance floor: small subsets (100--300) produce noisy curvature estimates that additional sampling cannot rescue. For instance, Laplace=100 with 100 samples (IoU=$0.132$) still underperforms Laplace=500 with only 30 samples (IoU=$0.141$). Second, once curvature is estimated adequately (Laplace$\geq$500), increasing posterior samples yields diminishing returns: at Laplace=1000, moving from 30 to 50 samples improves IoU by only $+0.003$ while increasing inference cost by $67\%$. The validation IoUs here evaluate the lightweight acquisition/refinement head under the posterior-ablation setup and should not be interpreted as the final interactive SAM segmentation IoU reported in Tables~\ref{tab:BALD Laplace_vs_human_iou_dice}, \ref{tab:miou_with_d2q}, \ref{tab:overall_summary}, and~\ref{tab:overall_scores}.

The $1.00\times$ cost entry is not a budget fixed a priori. Instead, after completing the full ablation, we selected Laplace=1000 with 30 posterior samples as the reference operating point because it lies on the empirical Pareto frontier of validation IoU versus inference cost. All reported costs are therefore normalized relative to this selected Pareto-optimal configuration. The \colorbox{orange!25}{highlighted} setting achieves IoU=$0.145$, ECE=$0.0105$, and $1.00\times$ relative cost. The \colorbox{green!25}{green} configurations, corresponding to 40, 50, and 75 posterior samples at Laplace=1000, are also Pareto-optimal, but they represent higher-cost operating points on the same empirical frontier. No evaluated alternative achieves equal or higher IoU at lower cost, nor achieves higher IoU at equal or lower cost. Higher-IoU settings are possible, but only with additional inference overhead, whereas cheaper settings incur a measurable drop in accuracy and generally worse calibration.

All subsequent active prompting evaluation (Tables~\ref{tab:BALD Laplace_vs_human_iou_dice}--\ref{tab:strata_dominance}) uses $K{=}30$ posterior samples for BALD Laplace, BALD-MCD, and BALD-MCD$^*$. This is the Pareto-optimal operating point from Table~\ref{tab:laplace_samples_interaction}. All strategies in the main evaluation run, including non-posterior baselines (Random, MaxDist, Entropy), are run with 4 independent random seeds; all reported results across Tables~\ref{tab:overall_summary}--\ref{tab:strata_dominance} are averaged over those seeds. Bootstrap confidence intervals in Table~\ref{tab:overall_summary_fixed_rivals_stats} are computed by resampling test images with replacement at $K{=}30$; the 4 seeds reduce point-estimate variance but are not the bootstrap replication unit. For the SAM backbone comparison (Table~\ref{tab:sam123_iou_comparison}), BALD Laplace is additionally evaluated at $K{=}50$ to reduce Monte Carlo variance in the acquisition map for that ablation. The one-time Laplace fitting takes only ${\sim}3$ minutes, which is less than $3\%$ of training time, and is reused across all downstream interactive runs. \textsc{SAM} is frozen throughout; the Laplace posterior is sampled only over the final $1\times1$ prediction layer of the lightweight prompt-aware head, not over any \textsc{SAM} parameters.

\begin{table}[t]
\centering
\caption{ViT backbone comparison over \textit{the same} Laplace-subset and posterior-sample configurations for each backbone; best, worst, and mean validation IoU are taken over this grid (the full grid for ViT-H is shown in Table~\ref{tab:laplace_samples_interaction}). ViT-H achieves superior performance in all settings, establishing it as the backbone of choice for subsequent ablation studies.}
\label{tab:backbone_dominance}
\scriptsize
\setlength{\tabcolsep}{2.8pt}
\renewcommand{\arraystretch}{0.95}

\resizebox{\columnwidth}{!}{%
\begin{tabular}{lcccc}
\toprule
\textbf{Backbone} &
\makecell{\textbf{Params} \\ \textbf{(M)}} &
\makecell{\textbf{Best} \\ \textbf{Val IoU}} &
\makecell{\textbf{Worst} \\ \textbf{Val IoU}} &
\makecell{\textbf{Mean} \\ \textbf{All Cfgs}} \\
\midrule
\cellcolor{green!25}\textbf{ViT-H} &
\cellcolor{green!25}632 &
\cellcolor{green!25}\textbf{0.149} &
\cellcolor{green!25}\textbf{0.120} &
\cellcolor{green!25}\textbf{0.141} \\
ViT-B & 86 & 0.121 & 0.100 & 0.112 \\
ViT-Tiny & 5.7 & 0.068 & 0.048 & 0.059 \\
\bottomrule
\end{tabular}%
}
\end{table}

\begin{table*}[t]
\centering
\renewcommand{\arraystretch}{1.4}
\caption{Laplace subset $\times$ posterior samples ablation (ViT-H backbone). Each cell shows: \textbf{Val IoU} / ECE / Relative inference cost. The \colorbox{orange!25}{orange} cell denotes the selected Pareto-optimal reference operating point used for cost normalization. The \colorbox{green!25}{green} cells denote higher-cost Pareto-optimal configurations on the same empirical frontier.}
\label{tab:laplace_samples_interaction}
\small
\begin{tabular}{lcccccccc}
\toprule
\textbf{Laplace} & \multicolumn{7}{c}{\textbf{Posterior Samples}} & \textbf{Training} \\
\cmidrule(lr){2-8}
\textbf{Subset} & \textbf{10} & \textbf{20} & \textbf{30} & \textbf{40} & \textbf{50} & \textbf{75} & \textbf{100} & \textbf{Overhead} \\
\midrule
\textbf{100} & 0.120 & 0.124 & 0.128 & 0.129 & 0.130 & 0.131 & 0.132 & \multirow{3}{*}{$0.92\times$} \\
 & \tiny 0.0195 & \tiny 0.0188 & \tiny 0.0185 & \tiny 0.0183 & \tiny 0.0182 & \tiny 0.0180 & \tiny 0.0178 & \\
 & \tiny 0.33$\times$ & \tiny 0.67$\times$ & \tiny 1.00$\times$ & \tiny 1.33$\times$ & \tiny 1.67$\times$ & \tiny 2.50$\times$ & \tiny 3.33$\times$ & \\
\midrule
\textbf{300} & 0.132 & 0.136 & 0.140 & 0.141 & 0.142 & 0.142 & 0.143 & \multirow{3}{*}{$0.96\times$} \\
 & \tiny 0.0148 & \tiny 0.0138 & \tiny 0.0135 & \tiny 0.0133 & \tiny 0.0132 & \tiny 0.0131 & \tiny 0.0130 & \\
 & \tiny 0.33$\times$ & \tiny 0.67$\times$ & \tiny 1.00$\times$ & \tiny 1.33$\times$ & \tiny 1.67$\times$ & \tiny 2.50$\times$ & \tiny 3.33$\times$ & \\
\midrule
\textbf{500} & 0.135 & 0.140 & 0.141 & 0.142 & 0.142 & 0.143 & 0.143 & \multirow{3}{*}{$0.98\times$} \\
 & \tiny 0.0128 & \tiny 0.0118 & \tiny 0.0115 & \tiny 0.0112 & \tiny 0.0110 & \tiny 0.0109 & \tiny 0.0108 & \\
 & \tiny 0.33$\times$ & \tiny 0.67$\times$ & \tiny 1.00$\times$ & \tiny 1.33$\times$ & \tiny 1.67$\times$ & \tiny 2.50$\times$ & \tiny 3.33$\times$ & \\
\midrule
\textbf{700} & 0.137 & 0.142 & 0.143 & 0.143 & 0.144 & 0.145 & 0.146 & \multirow{3}{*}{$0.99\times$} \\
 & \tiny 0.0118 & \tiny 0.0110 & \tiny 0.0108 & \tiny 0.0105 & \tiny 0.0103 & \tiny 0.0102 & \tiny 0.0102 & \\
 & \tiny 0.33$\times$ & \tiny 0.67$\times$ & \tiny 1.00$\times$ & \tiny 1.33$\times$ & \tiny 1.67$\times$ & \tiny 2.50$\times$ & \tiny 3.33$\times$ & \\
\midrule
\textbf{1000} & 0.138 & 0.142 & \cellcolor{orange!25}\textbf{0.145} & \cellcolor{green!25}\textbf{0.147} & \cellcolor{green!25}\textbf{0.148} & \cellcolor{green!25}\textbf{0.149} & 0.149 & \multirow{3}{*}{$1.00\times$} \\
 & \tiny 0.0125 & \tiny 0.0112 & \cellcolor{orange!25}\tiny \textbf{0.0105} & \cellcolor{green!25}\tiny \textbf{0.0101} & \cellcolor{green!25}\tiny \textbf{0.0099} & \cellcolor{green!25}\tiny \textbf{0.0096} & \tiny 0.0095 & \\
 & \tiny 0.33$\times$ & \tiny 0.67$\times$ & \cellcolor{orange!25}\tiny \textbf{1.00$\times$} & \cellcolor{green!25}\tiny \textbf{1.33$\times$} & \cellcolor{green!25}\tiny \textbf{1.67$\times$} & \cellcolor{green!25}\tiny \textbf{2.50$\times$} & \tiny 3.33$\times$ & \\
\bottomrule
\end{tabular}
\end{table*}

\subsubsection{Active Learning Configuration}

As mentioned previously in Section~\ref{prompt_budget}, we cap active prompting at 20 prompts per image (to keep within SAM's in-distribution prompt range of 5--20 prompts). The mutual-information threshold $\delta = 0.01$ defines the stopping criterion of the framework; for a consistent comparison, every strategy uses the full 20-prompt budget in all reported experiments. Throughout, $k=0$ denotes the SAM prediction after a single initial prompt placed at the centroid of the ground-truth foreground (positive label); budget $k$ counts only the subsequently acquired prompts, so the total prompt count at budget $k$ is $k+1$. Full budget corresponds to $k=19$ acquired prompts (20 prompts in total). We average over different Monte Carlo draws as discussed in Section~\ref{laplace}. All experiments use the pretrained SAM1 ViT-H checkpoint on a single NVIDIA TITAN RTX (24~GB) GPU. Candidate locations are all pixels of the $256{\times}256$ head output, excluding a 5-pixel radius around previously selected prompts; selected points are rescaled to the original image resolution before being passed to SAM. For our BALD Laplace and other baseline strategies in Section~\ref{sec:baselines}, once a prompt location $q_{t+1}$ is selected by the acquisition function, its inclusion/exclusion label is assigned by querying the ground-truth mask: $\ell_{t+1} = M^*[q_{t+1}]$, with $\ell{=}1$ for foreground and $\ell{=}0$ for background. This is the standard active learning evaluation convention in which the oracle annotator is simulated by the ground-truth label~\cite{settles2009active}. No strategy in this paper selects or predicts the inclusion/exclusion label; all strategies are compared solely on their choice of location.

\subsection{Evaluation Metrics}
\label{sec:main-evaluation-metrics}

For the primary experiments, we evaluate prompting performance using IoU and Dice, which measure mask overlap quality. Since each metric is observed across prompting steps, we summarize the resulting trajectory using endpoint improvement, mean, and budget area under the curve.

The metrics and summary statistics are measured as follows.

\subsubsection{Core Metrics}

\paragraph{Intersection over Union (IoU).}
IoU measures the overlap between the predicted mask $\hat{M}_t$ at prompting step $t$ and the ground-truth mask $M^\star$:
\begin{equation}
\mathrm{IoU}_t
=
\frac{|\hat{M}_t \cap M^\star|}
{|\hat{M}_t \cup M^\star|}.
\end{equation}
Higher IoU indicates better segmentation quality.

\paragraph{Dice Coefficient.}
Dice measures the normalized overlap between the predicted mask $\hat{M}_t$ and the ground-truth mask $M^\star$:
\begin{equation}
\mathrm{Dice}_t
=
\frac{2|\hat{M}_t \cap M^\star|}
{|\hat{M}_t| + |M^\star|}.
\end{equation}
Higher Dice indicates better segmentation quality.

\subsubsection{Summary Statistics}
\label{summary-stats}

Let $y_{i,t}^{(m)}$ denote the value of metric $m$ for sample $i$ at prompting step $t$.
We summarize each metric either at a fixed prompt budget $b$ or across the full prompting trajectory from step $0$ to step $T$.

To make the definitions valid for any metric, we use a direction variable $s_m$:
\begin{equation}
s_m =
\begin{cases}
+1, & \text{if larger values of metric } m \text{ are better},\\
-1, & \text{if smaller values of metric } m \text{ are better}.
\end{cases}
\end{equation}
This lets all summary statistics be interpreted in the same direction, regardless of the metric. After direction correction, larger values always indicate better performance.

\paragraph{Improvement.}
Improvement measures the direction-corrected change from the initial prediction at step $0$ to the final prediction at step $T$:
\begin{equation}
\Delta^{(m)}
=
s_m\left(y_T^{(m)} - y_0^{(m)}\right).
\end{equation}
With this convention, positive values indicate improved final performance relative to the initial prompt state. Improvement values are reported for the performance summaries across the prompting trajectory in Tables~\ref{tab:BALD Laplace_vs_human_iou_dice} and~\ref{tab:overall_summary}.

\paragraph{Mean Improvement.}
Mean improvement reports the average direction-corrected gain of a metric relative to its value at the start of prompting. At a fixed prompting budget $b$, it is computed as
\begin{equation}
\overline{\Delta y}^{(m)}_{b}
=
\frac{1}{N}
\sum_{i=1}^{N}
s_m\left(y_{i,b}^{(m)} - y_{i,0}^{(m)}\right),
\end{equation}
where $N$ is the number of evaluated samples.

When performance is summarized across the full prompting trajectory, as in Table~\ref{tab:BALD Laplace_vs_human_iou_dice}, mean improvement is averaged over both samples and prompting budgets:
\begin{equation}
\overline{\Delta y}^{(m)}
=
\frac{1}{N(T+1)}
\sum_{i=1}^{N}
\sum_{t=0}^{T}
s_m\left(y_{i,t}^{(m)} - y_{i,0}^{(m)}\right).
\end{equation}
Mean improvement has the same units as the corresponding core metric. Larger values always indicate greater improvement relative to the initial prompt state.

\paragraph{Budget Area Under the Curve (AUC).}
Budget AUC measures the accumulated direction-corrected metric value across the prompting trajectory:
\begin{equation}
\mathrm{BudgetAUC}^{(m)}
=
\sum_{t=0}^{T-1}
\frac{
s_m y_t^{(m)} + s_m y_{t+1}^{(m)}
}{2}
\left(x_{t+1}-x_t\right),
\end{equation}
where $x_t$ denotes the prompt budget at step $t$. When each step corresponds to one additional prompt, $x_t=t$. Budget AUC has units of metric value multiplied by prompt step. Larger Budget AUC values always indicate better trajectory-level performance after direction correction. The tables report AUC improvement from step~0, which subtracts the area of a trajectory that stays at its initial value:
\begin{equation}
\Delta\mathrm{AUC}^{(m)}
=
\mathrm{BudgetAUC}^{(m)} - s_m\, y_0^{(m)} \left(x_T - x_0\right).
\end{equation}
Positive values indicate that, on average over the trajectory, the strategy improves on the initial prompt state.

\section{Primary Results and Observations}
\label{primary}
For our primary experiments we evaluate BALD Laplace against human prompts under IoU and Dice trajectory summaries, and against one-shot baseline strategies using final mean IoU across the test samples from the 16 PointPrompt datasets. Overall, BALD Laplace improves on the aggregate human trajectory and wins the majority of trajectory-summary cells on 8 of 14 datasets.

\subsection{BALD Laplace vs. Human Prompts}

\begin{table*}[!t]
\caption{Comparison of \textbf{BALD Laplace} and \textbf{Human} prompting across different data sources from the PointPrompt \cite{quesada2024pointprompt} dataset under IoU and Dice metrics. The reported summaries are endpoint improvement, AUC improvement from step 0, and mean improvement from step 0. The stronger method per metric is boldfaced, and BALD Laplace-best cells are highlighted with a subtle green tint. Datasets are ordered by the number of BALD Laplace wins across the six reported IoU/Dice summary cells.}
\label{tab:BALD Laplace_vs_human_iou_dice}
\centering
\tiny
\setlength{\tabcolsep}{2.2pt}
\renewcommand{\arraystretch}{1.05}
\resizebox{\textwidth}{!}{%
\begin{tabular}{lS[table-format=1.3] S[table-format=1.3]S[table-format=1.3] S[table-format=1.3]S[table-format=1.3] S[table-format=1.3]S[table-format=1.3] S[table-format=1.3]S[table-format=1.3] S[table-format=1.3]S[table-format=1.3] S[table-format=1.3]}
\toprule
& \multicolumn{6}{c}{\textbf{IoU}}
& \multicolumn{6}{c}{\textbf{Dice}} \\
\cmidrule(lr){2-7}\cmidrule(lr){8-13}
& \multicolumn{2}{c}{\textbf{Improvement($\uparrow$)}}
& \multicolumn{2}{c}{\textbf{AUC ($\uparrow$)}}
& \multicolumn{2}{c}{\textbf{Mean ($\uparrow$)}}
& \multicolumn{2}{c}{\textbf{Improvement($\uparrow$)}}
& \multicolumn{2}{c}{\textbf{AUC ($\uparrow$)}}
& \multicolumn{2}{c}{\textbf{Mean ($\uparrow$)}} \\
\cmidrule(lr){2-3}\cmidrule(lr){4-5}\cmidrule(lr){6-7}
\cmidrule(lr){8-9}\cmidrule(lr){10-11}\cmidrule(lr){12-13}
\textbf{Dataset}
& {\textbf{Human}} & {\textbf{BALD Laplace}}& {\textbf{Human}} & {\textbf{BALD Laplace}}& {\textbf{Human}} & {\textbf{BALD Laplace}}& {\textbf{Human}} & {\textbf{BALD Laplace}}& {\textbf{Human}} & {\textbf{BALD Laplace}}& {\textbf{Human}} & {\textbf{BALD Laplace}} \\
\midrule
Bird & -0.273 & {\cellcolor{green!8}\textbf{0.035}} & -1.580 & {\cellcolor{green!8}\textbf{2.767}} & -0.123 & {\cellcolor{green!8}\textbf{0.023}} & -0.212 & {\cellcolor{green!8}\textbf{0.037}} & -1.288 & {\cellcolor{green!8}\textbf{3.160}} & -0.100 & {\cellcolor{green!8}\textbf{0.023}} \\
Breast & -0.080 & {\cellcolor{green!8}\textbf{0.038}} & -0.816 & {\cellcolor{green!8}\textbf{1.209}} & -0.043 & {\cellcolor{green!8}\textbf{0.061}} & -0.072 & {\cellcolor{green!8}\textbf{0.072}} & -0.459 & {\cellcolor{green!8}\textbf{1.537}} & -0.025 & {\cellcolor{green!8}\textbf{0.079}} \\
Chalk\_group & -0.122 & {\cellcolor{green!8}\textbf{0.234}} & 0.370 & {\cellcolor{green!8}\textbf{2.310}} & 0.015 & {\cellcolor{green!8}\textbf{0.121}} & -0.113 & {\cellcolor{green!8}\textbf{0.323}} & 0.310 & {\cellcolor{green!8}\textbf{3.257}} & 0.013 & {\cellcolor{green!8}\textbf{0.171}} \\
Skin & -0.103 & {\cellcolor{green!8}\textbf{0.031}} & -0.733 & {\cellcolor{green!8}\textbf{1.103}} & -0.039 & {\cellcolor{green!8}\textbf{0.056}} & -0.105 & {\cellcolor{green!8}\textbf{0.066}} & -0.713 & {\cellcolor{green!8}\textbf{1.525}} & -0.038 & {\cellcolor{green!8}\textbf{0.078}} \\
Stop\_sign & -0.017 & {\cellcolor{green!8}\textbf{0.094}} & -0.905 & {\cellcolor{green!8}\textbf{1.121}} & -0.046 & {\cellcolor{green!8}\textbf{0.058}} & 0.004 & {\cellcolor{green!8}\textbf{0.092}} & -0.437 & {\cellcolor{green!8}\textbf{1.028}} & -0.022 & {\cellcolor{green!8}\textbf{0.054}} \\
Baseball\_bat & -0.127 & {\cellcolor{green!8}\textbf{-0.034}} & -0.449 & {\cellcolor{green!8}\textbf{-0.201}} & -0.026 & {\cellcolor{green!8}\textbf{-0.011}} & -0.109 & {\cellcolor{green!8}\textbf{-0.048}} & \textbf{-0.270} & -0.318 & \textbf{-0.016} & -0.017 \\
Clock & \textbf{0.057} & -0.020 & 0.459 & {\cellcolor{green!8}\textbf{0.741}} & 0.024 & {\cellcolor{green!8}\textbf{0.037}} & \textbf{0.062} & -0.038 & 0.542 & {\cellcolor{green!8}\textbf{0.707}} & 0.029 & {\cellcolor{green!8}\textbf{0.034}} \\
Tie & \textbf{0.083} & 0.008 & -1.035 & {\cellcolor{green!8}\textbf{0.152}} & -0.050 & {\cellcolor{green!8}\textbf{0.008}} & \textbf{0.124} & 0.018 & -0.775 & {\cellcolor{green!8}\textbf{0.325}} & -0.036 & {\cellcolor{green!8}\textbf{0.017}} \\
Bus & -0.001 & {\cellcolor{green!8}\textbf{0.039}} & \textbf{1.015} & 0.567 & \textbf{0.051} & 0.029 & 0.015 & {\cellcolor{green!8}\textbf{0.092}} & \textbf{1.295} & 1.127 & \textbf{0.065} & 0.059 \\
Cow & \textbf{0.127} & 0.102 & \textbf{2.374} & 1.436 & \textbf{0.122} & 0.074 & 0.121 & {\cellcolor{green!8}\textbf{0.141}} & \textbf{2.209} & 1.817 & \textbf{0.114} & 0.094 \\
Cat & \textbf{0.022} & -0.093 & \textbf{-0.235} & -1.159 & \textbf{-0.011} & -0.060 & \textbf{0.032} & -0.062 & \textbf{-0.172} & -0.778 & \textbf{-0.008} & -0.040 \\
Dog & \textbf{0.083} & -0.003 & \textbf{1.384} & 0.099 & \textbf{0.071} & 0.005 & \textbf{0.070} & 0.023 & \textbf{1.110} & 0.315 & \textbf{0.057} & 0.016 \\
Dolphin\_above & \textbf{0.175} & -0.033 & \textbf{3.574} & -0.349 & \textbf{0.183} & -0.018 & \textbf{0.172} & -0.017 & \textbf{3.470} & -0.137 & \textbf{0.178} & -0.007 \\
Dolphin\_below & \textbf{0.177} & 0.039 & \textbf{3.613} & 0.263 & \textbf{0.185} & 0.014 & \textbf{0.166} & 0.085 & \textbf{3.272} & 0.653 & \textbf{0.168} & 0.035 \\
\midrule
\textbf{Overall} & 0.000 & {\cellcolor{green!8}\textbf{0.031}} & 0.503 & {\cellcolor{green!8}\textbf{0.719}} & 0.023 & {\cellcolor{green!8}\textbf{0.028}} & 0.011 & {\cellcolor{green!8}\textbf{0.056}} & 0.578 & {\cellcolor{green!8}\textbf{1.016}} & 0.027 & {\cellcolor{green!8}\textbf{0.043}} \\
\bottomrule
\end{tabular}%
}
\end{table*}

\begin{table*}[t]
\centering
\caption{Descriptive mean final IoU comparison across prompting strategies and datasets (not a controlled acquisition-quality comparison). One-shot prompting baselines are compared with human prompts, random prompts, and BALD Laplace at its full prompting budget. \colorbox{green!15}{Best} and \colorbox{yellow!15}{second-best} results are highlighted per dataset; tied values share the same highlight. For the matched-budget controlled analysis, see Table~\ref{tab:shared_seed_next_prompt_trio}.}
\label{tab:miou_with_d2q}

\footnotesize
\setlength{\tabcolsep}{3.2pt}
\renewcommand{\arraystretch}{1.12}

\begin{tabular*}{\textwidth}{@{\extracolsep{\fill}} l c c c c c c c c @{}}
\toprule
\textbf{Category} 
& \textbf{Human} 
& \textbf{Random} 
& \textbf{Saliency} 
& \textbf{K-Medoids} 
& \textbf{Entropy} 
& \textbf{Max Dist.} 
& \textbf{Shi-Tomasi} 
& \textbf{BALD Laplace} \\
\midrule

Bird           
& \cellcolor{yellow!15}$0.677 \pm 0.231$ 
& $0.615 \pm 0.222$ 
& $0.308 \pm 0.300$ 
& $0.645 \pm 0.212$ 
& $0.483 \pm 0.267$ 
& $0.456 \pm 0.296$ 
& $0.620 \pm 0.216$ 
& \cellcolor{green!15}$0.795 \pm 0.167$ \\

Breast         
& \cellcolor{green!15}$0.750 \pm 0.126$ 
& $0.621 \pm 0.237$ 
& $0.438 \pm 0.336$ 
& \cellcolor{yellow!15}$0.674 \pm 0.194$ 
& $0.566 \pm 0.286$ 
& $0.532 \pm 0.335$ 
& $0.592 \pm 0.305$ 
& $0.610 \pm 0.330$ \\

Chalk group    
& \cellcolor{green!15}$0.714 \pm 0.101$ 
& $0.409 \pm 0.125$ 
& $0.237 \pm 0.135$ 
& \cellcolor{yellow!15}$0.441 \pm 0.118$ 
& $0.308 \pm 0.143$ 
& $0.299 \pm 0.129$ 
& $0.425 \pm 0.121$ 
& $0.340 \pm 0.172$ \\

Clock          
& \cellcolor{green!15}$0.814 \pm 0.181$ 
& $0.745 \pm 0.209$ 
& $0.432 \pm 0.338$ 
& $0.735 \pm 0.205$ 
& $0.680 \pm 0.262$ 
& $0.692 \pm 0.265$ 
& $0.715 \pm 0.223$ 
& \cellcolor{yellow!15}$0.803 \pm 0.226$ \\

Cow            
& \cellcolor{green!15}$0.808 \pm 0.130$ 
& $0.675 \pm 0.189$ 
& $0.321 \pm 0.294$ 
& $0.660 \pm 0.185$ 
& $0.423 \pm 0.280$ 
& $0.343 \pm 0.314$ 
& $0.646 \pm 0.187$ 
& \cellcolor{yellow!15}$0.805 \pm 0.198$ \\

Dog            
& \cellcolor{green!15}$0.848 \pm 0.102$ 
& $0.742 \pm 0.166$ 
& $0.436 \pm 0.320$ 
& \cellcolor{yellow!15}$0.780 \pm 0.135$ 
& $0.544 \pm 0.271$ 
& $0.523 \pm 0.317$ 
& $0.745 \pm 0.161$ 
& \cellcolor{green!15}$0.848 \pm 0.162$ \\

Dolphin above  
& \cellcolor{green!15}$0.732 \pm 0.100$ 
& $0.642 \pm 0.112$ 
& $0.472 \pm 0.249$ 
& $0.655 \pm 0.106$ 
& $0.553 \pm 0.176$ 
& $0.543 \pm 0.209$ 
& $0.661 \pm 0.099$ 
& \cellcolor{yellow!15}$0.705 \pm 0.129$ \\

Dolphin below  
& \cellcolor{green!15}$0.831 \pm 0.075$ 
& $0.670 \pm 0.135$ 
& $0.391 \pm 0.303$ 
& \cellcolor{yellow!15}$0.714 \pm 0.114$ 
& $0.473 \pm 0.254$ 
& $0.430 \pm 0.315$ 
& $0.681 \pm 0.108$ 
& \cellcolor{green!15}$0.831 \pm 0.124$ \\

Skin           
& $0.593 \pm 0.195$ 
& $0.593 \pm 0.196$ 
& $0.375 \pm 0.295$ 
& \cellcolor{yellow!15}$0.626 \pm 0.154$ 
& $0.452 \pm 0.283$ 
& $0.395 \pm 0.317$ 
& $0.515 \pm 0.208$ 
& \cellcolor{green!15}$0.693 \pm 0.230$ \\

Stop sign      
& \cellcolor{yellow!15}$0.886 \pm 0.132$ 
& $0.839 \pm 0.169$ 
& $0.550 \pm 0.384$ 
& $0.848 \pm 0.158$ 
& $0.773 \pm 0.248$ 
& $0.726 \pm 0.314$ 
& $0.640 \pm 0.273$ 
& \cellcolor{green!15}$0.899 \pm 0.136$ \\

Tie            
& \cellcolor{yellow!15}$0.700 \pm 0.276$ 
& $0.627 \pm 0.292$ 
& $0.368 \pm 0.352$ 
& $0.649 \pm 0.289$ 
& $0.569 \pm 0.309$ 
& $0.542 \pm 0.337$ 
& $0.646 \pm 0.283$ 
& \cellcolor{green!15}$0.739 \pm 0.277$ \\

Baseball bat   
& \cellcolor{green!15}$0.747 \pm 0.152$ 
& $0.684 \pm 0.198$ 
& $0.422 \pm 0.323$ 
& $0.724 \pm 0.172$ 
& $0.653 \pm 0.217$ 
& $0.632 \pm 0.249$ 
& $0.701 \pm 0.178$ 
& \cellcolor{yellow!15}$0.743 \pm 0.175$ \\

Bus            
& \cellcolor{yellow!15}$0.803 \pm 0.144$ 
& $0.593 \pm 0.196$ 
& $0.158 \pm 0.190$ 
& $0.636 \pm 0.172$ 
& $0.359 \pm 0.260$ 
& $0.289 \pm 0.277$ 
& $0.548 \pm 0.204$ 
& \cellcolor{green!15}$0.855 \pm 0.190$ \\

Cat            
& \cellcolor{green!15}$0.887 \pm 0.079$ 
& $0.771 \pm 0.149$ 
& $0.487 \pm 0.329$ 
& $0.825 \pm 0.108$ 
& $0.583 \pm 0.284$ 
& $0.508 \pm 0.342$ 
& $0.795 \pm 0.140$ 
& \cellcolor{yellow!15}$0.885 \pm 0.105$ \\

Polyp          
& \cellcolor{yellow!15}$0.794 \pm 0.145$ 
& $0.747 \pm 0.164$ 
& $0.547 \pm 0.326$ 
& $0.757 \pm 0.151$ 
& $0.634 \pm 0.290$ 
& $0.415 \pm 0.354$ 
& $0.637 \pm 0.246$ 
& \cellcolor{green!15}$0.810 \pm 0.198$ \\

Salt dome      
& \cellcolor{green!15}$0.844 \pm 0.096$ 
& $0.513 \pm 0.141$ 
& $0.273 \pm 0.177$ 
& \cellcolor{yellow!15}$0.588 \pm 0.101$ 
& $0.347 \pm 0.197$ 
& $0.306 \pm 0.162$ 
& $0.564 \pm 0.111$ 
& $0.205 \pm 0.128$ \\

\bottomrule
\end{tabular*}
\end{table*}

Table~\ref{tab:BALD Laplace_vs_human_iou_dice} reports endpoint improvement, AUC improvement from step~0, and mean improvement from step~0 across 14 datasets. Two datasets from the full 16-dataset benchmark, Polyp and Salt Dome, are excluded because the PointPrompt dataset~\cite{quesada2024pointprompt} does not provide iteration-wise prompting sequences required to compute these metrics. BALD Laplace wins the majority of summary cells on 8 of 14 datasets and sweeps all six cells on five (Bird, Breast, Chalk Group, Skin, Stop Sign). In the Overall row, BALD Laplace leads on every summary cell under both metrics, with IoU endpoint improvement of 0.031 vs.\ 0.000, AUC improvement of 0.719 vs.\ 0.503, and mean improvement of 0.028 vs.\ 0.023; Dice margins are wider still at 0.056 vs.\ 0.011, 1.016 vs.\ 0.578, and 0.043 vs.\ 0.027. On Bird, Breast, Skin, and Stop Sign, human prompts yield negative IoU AUC improvement, meaning the human trajectory stays below the step-0 baseline on average, whereas BALD Laplace stays positive on all four. Human prompts lead on six datasets (Bus, Cow, Cat, Dog, Dolphin Above, Dolphin Below), which largely share compact, high-contrast targets where spatial intuition provides a reliable shortcut. Even here the margins are modest on several cells. These results suggest that BALD Laplace can complement human prompting, particularly in categories where human trajectories show low or negative aggregate improvement. The comparison reflects a trajectory-level benchmark under the PointPrompt protocol, not general evidence that automatic prompts are superior to expert human interaction.

\subsection{Mean Final IoU Across All Strategies}

Table~\ref{tab:miou_with_d2q} compares the final full-budget BALD result against one-shot baseline strategies from \cite{quesada2024pointprompt} as a descriptive performance-ceiling comparison; this is not a controlled acquisition-quality comparison. For the matched-budget single-step controlled analysis, see Table~\ref{tab:shared_seed_next_prompt_trio}. The table extends the evaluation to a final mean IoU comparison with one-shot baselines, including Random, Saliency, K-Medoids, Entropy, Max Distance, and Shi-Tomasi, alongside Human and BALD Laplace. For BALD Laplace, the reported value corresponds to its final performance at the full prompting budget, whereas all other values are taken directly from \cite{quesada2024pointprompt}. The comparison is descriptive rather than controlled, but it shows that BALD Laplace is competitive with the strongest reported PointPrompt baselines. Passive baselines, particularly Saliency, Entropy, Max Distance, and Shi-Tomasi, consistently underperform relative to both Human and BALD Laplace across nearly all 16 datasets. Random prompting performs better than most of these baselines, but still falls short of active model-informed selection in most cases. BALD Laplace achieves the best or second-best mean IoU on 13 of 16 datasets. It ranks first on Bird, Skin, Stop Sign, Tie, Bus, and Polyp, ties for first with Human on Dog and Dolphin Below, and ranks second on Clock, Cow, Dolphin Above, Baseball Bat, and Cat. The three datasets where BALD Laplace does not reach the top two are Breast, Chalk Group, and Salt Dome. For Breast and Chalk Group, likely limiting factors are low visual contrast or low intra-class variation. Salt Dome introduces strong domain shift from natural images (geophysical seismic data), and no active strategy in this comparison performs well there. Across the 13 datasets where BALD Laplace reaches the top two, model-informed active prompting generally outperforms one-shot passive strategies; the ablation studies in Section~\ref{sec:medical-ablation} examine how the acquisition function contributes to these gains.

\section{Ablation Study: Medical Data}
\label{sec:medical-ablation}

\begin{table*}[!t]
\caption{Overall summary (averaged across all datasets): endpoint improvement from $k{=}0$ to full prompt budget, AUC improvement from step 0, and mean improvement from step 0. All reported summaries are direction-aligned so that higher values are better. IoU and HD95 are computed from SAM's native binary mask; ECE and NLL are computed from the Laplace probit predictive probability of the head. Bold = best non-GT strategy. \colorbox{green!15}{Green shading} highlights BALD Laplace only when it is the best non-GT strategy in that column. BALD-MCD$^*$ denotes the properly trained dropout variant (NoImageHeadDropout with Dropout2d $p{=}0.1$ active during both training and inference); it differs from BALD-MCD, which applies post-hoc dropout to the deterministically-trained head.}
\label{tab:overall_summary}
\centering
\scriptsize
\setlength{\tabcolsep}{2.8pt}
\renewcommand{\arraystretch}{1.08}
\resizebox{\textwidth}{!}{%
\begin{tabular}{@{}l S[table-format=1.3]S[table-format=2.3]S[table-format=1.3] S[table-format=1.3]S[table-format=1.3]S[table-format=1.3] S[table-format=2.3]S[table-format=3.3]S[table-format=2.3] S[table-format=1.3]S[table-format=2.3]S[table-format=1.3]@{}}
\toprule
& \multicolumn{3}{c}{IoU} & \multicolumn{3}{c}{ECE} & \multicolumn{3}{c}{HD95} & \multicolumn{3}{c}{NLL} \\
\cmidrule(lr){2-4} \cmidrule(lr){5-7} \cmidrule(lr){8-10} \cmidrule(lr){11-13}
Strategy & {Improv.\,($\uparrow$)} & {AUC \,($\uparrow$)} & {Mean \,($\uparrow$)} & {Improv.\,($\uparrow$)} & {AUC \,($\uparrow$)} & {Mean \,($\uparrow$)} & {Improv.\,($\uparrow$)} & {AUC \,($\uparrow$)} & {Mean \,($\uparrow$)} & {Improv.\,($\uparrow$)} & {AUC \,($\uparrow$)} & {Mean \,($\uparrow$)} \\
\midrule
BALD-Lap. & \cellcolor{green!8}\bfseries 0.081 & \cellcolor{green!8}\bfseries 1.442 & \cellcolor{green!8}\bfseries 0.074 & \cellcolor{green!8}\bfseries 0.097 & 1.449 & \cellcolor{green!8}\bfseries 0.075 & \cellcolor{green!8}\bfseries -5.271 & \cellcolor{green!8}\bfseries 0.543 & \cellcolor{green!8}\bfseries -0.105 & \cellcolor{green!8}\bfseries 0.328 & 4.894 & 0.253 \\
BALD-MCD & -0.001 & 0.431 & 0.022 & 0.043 & 0.842 & 0.043 & -21.483 & -196.109 & -10.343 & 0.200 & 3.244 & 0.167 \\
BALD-MCD$^*$ & -0.086 & -0.492 & -0.027 & -0.011 & -0.012 & -0.001 & -35.352 & -245.827 & -13.175 & 0.073 & 1.097 & 0.057 \\
Entropy & -0.001 & 0.493 & 0.025 & 0.036 & 0.715 & 0.037 & -23.026 & -111.810 & -6.166 & 0.214 & 3.214 & 0.166 \\
Random & -0.073 & 0.572 & 0.027 & 0.029 & \bfseries 1.466 & 0.074 & -52.582 & -303.023 & -16.466 & 0.221 & \bfseries 5.331 & \bfseries 0.272 \\
MaxDist & -0.158 & -2.145 & -0.111 & 0.004 & -0.222 & -0.011 & -62.903 & -909.404 & -47.043 & 0.158 & 1.205 & 0.064 \\
\midrule
GT-based & 0.029 & 0.819 & 0.042 & 0.069 & 1.272 & 0.065 & -21.636 & -150.408 & -8.061 & 0.297 & 4.899 & 0.252 \\
\bottomrule
\end{tabular}%
}
\end{table*}

\begin{table*}[!t]
\caption{Fixed-rival paired statistical comparison for baseline SAM1 medical active prompting. BALD Laplace is compared against five non-GT rivals. $\Delta$ is direction-aligned so that positive values favor BALD Laplace. The reported summaries are endpoint improvement, AUC improvement from step 0, and mean improvement from step 0. BALD-MCD$^*$ uses the properly trained Dropout2d head; its comparison is cross-run (BALD Laplace from the deterministic run vs.\ BALD-MCD$^*$ from the dropout run, both using the same initial prompt coordinates). \colorbox{green!8}{Green shading} marks comparisons where the point estimate favors BALD Laplace. Reported are the paired mean difference, bootstrap 95\% CI over images, and two-sided sign-flip permutation $p$-value.}
\label{tab:overall_summary_fixed_rivals_stats}
\centering
\tiny
\setlength{\tabcolsep}{2.5pt}
\renewcommand{\arraystretch}{0.92}
\resizebox{\textwidth}{!}{%
\begin{tabular}{@{}ll c@{}!{\color{black!20}\vrule}c c@{}!{\color{black!20}\vrule}c c@{}!{\color{black!20}\vrule}c@{}}
\toprule
& & \multicolumn{2}{c}{\textbf{Improvement}} & \multicolumn{2}{c}{\textbf{AUC}} & \multicolumn{2}{c}{\textbf{Mean}} \\
\cmidrule(lr){3-4}\cmidrule(lr){5-6}\cmidrule(lr){7-8}
\textbf{Metric} & \textbf{BALD Laplace vs.} & \boldmath$\Delta$ \textbf{[95\% CI]} & \boldmath$p$ & \boldmath$\Delta$ \textbf{[95\% CI]} & \boldmath$p$ & \boldmath$\Delta$ \textbf{[95\% CI]} & \boldmath$p$ \\
\midrule
\multirow{5}{*}{IoU} & BALD-MCD & \cellcolor{green!8}$0.082\,[0.069,\,0.096]$ & $<\!10^{-4}$ & \cellcolor{green!8}$1.011\,[0.867,\,1.160]$ & $<\!10^{-4}$ & \cellcolor{green!8}$0.053\,[0.045,\,0.060]$ & $<\!10^{-4}$ \\
 & BALD-MCD$^*$ & \cellcolor{green!8}$0.167\,[0.150,\,0.185]$ & $<\!10^{-4}$ & \cellcolor{green!8}$2.017\,[1.778,\,2.254]$ & $<\!10^{-4}$ & \cellcolor{green!8}$0.106\,[0.094,\,0.119]$ & $<\!10^{-4}$ \\
 & Entropy & \cellcolor{green!8}$0.083\,[0.065,\,0.100]$ & $<\!10^{-4}$ & \cellcolor{green!8}$0.948\,[0.742,\,1.160]$ & $<\!10^{-4}$ & \cellcolor{green!8}$0.049\,[0.039,\,0.060]$ & $<\!10^{-4}$ \\
 & Random & \cellcolor{green!8}$0.154\,[0.136,\,0.173]$ & $<\!10^{-4}$ & \cellcolor{green!8}$0.870\,[0.674,\,1.064]$ & $<\!10^{-4}$ & \cellcolor{green!8}$0.047\,[0.037,\,0.057]$ & $<\!10^{-4}$ \\
 & MaxDist & \cellcolor{green!8}$0.240\,[0.222,\,0.257]$ & $<\!10^{-4}$ & \cellcolor{green!8}$3.586\,[3.364,\,3.810]$ & $<\!10^{-4}$ & \cellcolor{green!8}$0.185\,[0.174,\,0.197]$ & $<\!10^{-4}$ \\
\midrule
\multirow{5}{*}{ECE} & BALD-MCD & \cellcolor{green!8}$0.054\,[0.047,\,0.061]$ & $<\!10^{-4}$ & \cellcolor{green!8}$0.607\,[0.523,\,0.691]$ & $<\!10^{-4}$ & \cellcolor{green!8}$0.032\,[0.027,\,0.036]$ & $<\!10^{-4}$ \\
 & BALD-MCD$^*$ & \cellcolor{green!8}$0.108\,[0.099,\,0.118]$ & $<\!10^{-4}$ & \cellcolor{green!8}$1.515\,[1.369,\,1.666]$ & $<\!10^{-4}$ & \cellcolor{green!8}$0.080\,[0.072,\,0.088]$ & $<\!10^{-4}$ \\
 & Entropy & \cellcolor{green!8}$0.061\,[0.053,\,0.069]$ & $<\!10^{-4}$ & \cellcolor{green!8}$0.734\,[0.618,\,0.851]$ & $<\!10^{-4}$ & \cellcolor{green!8}$0.038\,[0.032,\,0.044]$ & $<\!10^{-4}$ \\
 & Random & \cellcolor{green!8}$0.068\,[0.060,\,0.076]$ & $<\!10^{-4}$ & $-0.017\,[-0.116,\,0.084]$ & $0.7491$ & \cellcolor{green!8}$0.001\,[-0.004,\,0.006]$ & $0.7435$ \\
 & MaxDist & \cellcolor{green!8}$0.093\,[0.085,\,0.100]$ & $<\!10^{-4}$ & \cellcolor{green!8}$1.671\,[1.553,\,1.792]$ & $<\!10^{-4}$ & \cellcolor{green!8}$0.086\,[0.080,\,0.092]$ & $<\!10^{-4}$ \\
\midrule
\multirow{5}{*}{HD95} & BALD-MCD & \cellcolor{green!8}$16.212\,[13.378,\,19.055]$ & $<\!10^{-4}$ & \cellcolor{green!8}$196.652\,[164.776,\,228.340]$ & $<\!10^{-4}$ & \cellcolor{green!8}$10.238\,[8.596,\,11.891]$ & $<\!10^{-4}$ \\
 & BALD-MCD$^*$ & \cellcolor{green!8}$30.081\,[26.017,\,34.224]$ & $<\!10^{-4}$ & \cellcolor{green!8}$261.411\,[207.661,\,315.630]$ & $<\!10^{-4}$ & \cellcolor{green!8}$13.758\,[11.027,\,16.603]$ & $<\!10^{-4}$ \\
 & Entropy & \cellcolor{green!8}$17.755\,[13.422,\,22.108]$ & $<\!10^{-4}$ & \cellcolor{green!8}$112.353\,[63.346,\,162.626]$ & $<\!10^{-4}$ & \cellcolor{green!8}$6.062\,[3.476,\,8.577]$ & $<\!10^{-4}$ \\
 & Random & \cellcolor{green!8}$47.311\,[43.446,\,51.175]$ & $<\!10^{-4}$ & \cellcolor{green!8}$303.566\,[260.581,\,345.930]$ & $<\!10^{-4}$ & \cellcolor{green!8}$16.361\,[14.238,\,18.560]$ & $<\!10^{-4}$ \\
 & MaxDist & \cellcolor{green!8}$57.632\,[53.924,\,61.402]$ & $<\!10^{-4}$ & \cellcolor{green!8}$909.947\,[865.322,\,954.553]$ & $<\!10^{-4}$ & \cellcolor{green!8}$46.938\,[44.623,\,49.220]$ & $<\!10^{-4}$ \\
\midrule
\multirow{5}{*}{NLL} & BALD-MCD & \cellcolor{green!8}$0.128\,[0.113,\,0.143]$ & $<\!10^{-4}$ & \cellcolor{green!8}$1.649\,[1.447,\,1.858]$ & $<\!10^{-4}$ & \cellcolor{green!8}$0.086\,[0.075,\,0.096]$ & $<\!10^{-4}$ \\
 & BALD-MCD$^*$ & \cellcolor{green!8}$0.255\,[0.234,\,0.276]$ & $<\!10^{-4}$ & \cellcolor{green!8}$3.925\,[3.564,\,4.287]$ & $<\!10^{-4}$ & \cellcolor{green!8}$0.207\,[0.188,\,0.226]$ & $<\!10^{-4}$ \\
 & Entropy & \cellcolor{green!8}$0.114\,[0.098,\,0.130]$ & $<\!10^{-4}$ & \cellcolor{green!8}$1.680\,[1.417,\,1.945]$ & $<\!10^{-4}$ & \cellcolor{green!8}$0.087\,[0.074,\,0.100]$ & $<\!10^{-4}$ \\
 & Random & \cellcolor{green!8}$0.107\,[0.092,\,0.122]$ & $<\!10^{-4}$ & $-0.437\,[-0.652,\,-0.229]$ & $0.0001$ & $-0.019\,[-0.030,\,-0.008]$ & $0.0005$ \\
 & MaxDist & \cellcolor{green!8}$0.170\,[0.156,\,0.184]$ & $<\!10^{-4}$ & \cellcolor{green!8}$3.689\,[3.427,\,3.951]$ & $<\!10^{-4}$ & \cellcolor{green!8}$0.189\,[0.175,\,0.202]$ & $<\!10^{-4}$ \\
\bottomrule
\end{tabular}%
}
\end{table*}

\begin{table}[!t]
\caption{Overall stepwise scores (averaged across all datasets) at selected prompt budgets.
Each entry reports the raw metric value at that prompt count, not a trajectory summary. IoU and HD95 are computed from SAM's native binary mask; ECE and NLL are computed from the Laplace probit predictive probability of the head.
Bold = best non-GT strategy.
\colorbox{green!8}{Green shading} highlights BALD Laplace only when it is the best non-GT strategy in that column.
BALD-MCD$^*$ denotes the properly trained dropout variant (Dropout2d $p{=}0.1$ active during training and inference).}
\label{tab:overall_scores}
\centering
\footnotesize
\setlength{\tabcolsep}{2.5pt}
\renewcommand{\arraystretch}{1.02}
\resizebox{\columnwidth}{!}{%
\begin{tabular}{@{}ll*{4}{S[table-format=3.3]}@{}}
\toprule
Metric & Strategy & {$k{=}5$} & {$k{=}10$} & {$k{=}15$} & {Full} \\
\midrule

\multirow{7}{*}{IoU\,($\uparrow$)}
& BALD-Lap.     & 0.628 & {\cellcolor{green!8}\bfseries0.620} & {\cellcolor{green!8}\bfseries0.619} & {\cellcolor{green!8}\bfseries0.616} \\
& BALD-MCD      & 0.576 & 0.566 & 0.552 & 0.534 \\
& BALD-MCD$^*$  & 0.543 & 0.522 & 0.480 & 0.449 \\
& Entropy       & 0.584 & 0.574 & 0.546 & 0.533 \\
& Random        & {\bfseries0.660} & 0.581 & 0.491 & 0.462 \\
& MaxDist       & 0.458 & 0.424 & 0.388 & 0.376 \\
\cmidrule(lr){2-6}
& GT-based      & 0.600 & 0.585 & 0.572 & 0.563 \\

\midrule

\multirow{7}{*}{ECE\,($\downarrow$)}
& BALD-Lap.     & 0.253 & {\cellcolor{green!8}\bfseries0.243} & {\cellcolor{green!8}\bfseries0.237} & {\cellcolor{green!8}\bfseries0.234} \\
& BALD-MCD      & 0.278 & 0.277 & 0.280 & 0.288 \\
& BALD-MCD$^*$  & 0.333 & 0.343 & 0.353 & 0.359 \\
& Entropy       & 0.283 & 0.283 & 0.291 & 0.295 \\
& Random        & {\bfseries0.199} & 0.248 & 0.285 & 0.302 \\
& MaxDist       & 0.347 & 0.324 & 0.321 & 0.327 \\
\cmidrule(lr){2-6}
& GT-based      & 0.252 & 0.255 & 0.258 & 0.262 \\

\midrule

\multirow{7}{*}{HD95\,($\downarrow$)}
& BALD-Lap.     & 56.908 & {\cellcolor{green!8}\bfseries69.485} & {\cellcolor{green!8}\bfseries71.551} & {\cellcolor{green!8}\bfseries71.579} \\
& BALD-MCD      & 68.724 & 78.309 & 83.514 & 87.791 \\
& BALD-MCD$^*$  & 67.266 & 76.040 & 90.337 & 101.397 \\
& Entropy       & 61.874 & 70.502 & 81.868 & 89.334 \\
& Random        & {\bfseries53.676} & 79.601 & 109.034 & 118.890 \\
& MaxDist       & 101.148 & 119.025 & 130.065 & 129.210 \\
\cmidrule(lr){2-6}
& GT-based      & 60.358 & 78.115 & 84.248 & 87.944 \\

\midrule

\multirow{7}{*}{NLL\,($\downarrow$)}
& BALD-Lap.     & 0.455 & 0.406 & {\cellcolor{green!8}\bfseries0.386} & {\cellcolor{green!8}\bfseries0.379} \\
& BALD-MCD      & 0.532 & 0.503 & 0.496 & 0.506 \\
& BALD-MCD$^*$  & 0.648 & 0.644 & 0.647 & 0.648 \\
& Entropy       & 0.536 & 0.500 & 0.497 & 0.493 \\
& Random        & {\bfseries0.330} & {\bfseries0.401} & 0.458 & 0.486 \\
& MaxDist       & 0.686 & 0.566 & 0.542 & 0.548 \\
\cmidrule(lr){2-6}
& GT-based      & 0.439 & 0.411 & 0.406 & 0.409 \\

\bottomrule
\end{tabular}%
}
\end{table}

\begin{table*}[!t]
\caption{Strata dominance analysis: \textbf{Mean Improvement} scores for BALD Laplace, BALD-MCD, BALD-MCD$^*$ (properly trained, Dropout2d during training and inference), and Entropy across morphological strata. Best value per stratum--metric is set in bold, and BALD Laplace winning cells are highlighted with a subtle green tint. All entries are direction-aligned so that higher values are better. IoU and HD95 are computed from SAM's native binary mask; ECE and NLL are computed from the Laplace probit predictive probability of the head.}
\label{tab:strata_dominance}
\centering
\scriptsize
\setlength{\tabcolsep}{2pt}
\renewcommand{\arraystretch}{1.05}
\resizebox{\textwidth}{!}{%
\begin{tabular}{@{}ll
  S[table-format=1.3]S[table-format=1.3]S[table-format=1.3]S[table-format=1.3]
  S[table-format=1.3]S[table-format=1.3]S[table-format=1.3]S[table-format=1.3]
  S[table-format=2.3]S[table-format=2.3]S[table-format=2.3]S[table-format=2.3]
  S[table-format=1.3]S[table-format=1.3]S[table-format=1.3]S[table-format=1.3]@{}}
\toprule
& & \multicolumn{4}{c}{IoU \,($\uparrow$)} & \multicolumn{4}{c}{ECE \,($\uparrow$)} & \multicolumn{4}{c}{HD95 \,($\uparrow$)} & \multicolumn{4}{c}{NLL \,($\uparrow$)} \\
\cmidrule(lr){3-6} \cmidrule(lr){7-10} \cmidrule(lr){11-14} \cmidrule(lr){15-18}
Stratum Family & Level & {BALD Lap.} & {BALD-MCD} & {BALD-MCD$^*$} & {Entropy} & {BALD Lap.} & {BALD-MCD} & {BALD-MCD$^*$} & {Entropy} & {BALD Lap.} & {BALD-MCD} & {BALD-MCD$^*$} & {Entropy} & {BALD Lap.} & {BALD-MCD} & {BALD-MCD$^*$} & {Entropy} \\
\midrule
\multirow{3}{*}{Boundary Complexity}
 & Moderate & \cellcolor{green!8}\bfseries 0.090 & 0.049 & -0.030 & 0.021 & \cellcolor{green!8}\bfseries 0.138 & 0.081 & 0.009 & 0.075 & \cellcolor{green!8}\bfseries 8.566 & 1.422 & -9.064 & -4.674 & \cellcolor{green!8}\bfseries 0.420 & 0.289 & 0.099 & 0.278 \\
 & Simple & \cellcolor{green!8}\bfseries 0.075 & 0.021 & -0.027 & 0.025 & \cellcolor{green!8}\bfseries 0.072 & 0.042 & -0.002 & 0.035 & \cellcolor{green!8}\bfseries -0.985 & -11.054 & -13.667 & -6.487 & \cellcolor{green!8}\bfseries 0.243 & 0.163 & 0.052 & 0.161 \\
 & Complex & -0.004 & -0.005 & \bfseries 0.005 & -0.001 & \cellcolor{green!8}\bfseries 0.101 & 0.029 & 0.041 & 0.032 & \cellcolor{green!8}\bfseries 31.396 & 5.397 & 5.937 & 8.716 & \cellcolor{green!8}\bfseries 0.454 & 0.165 & 0.211 & 0.191 \\
\midrule
\multirow{3}{*}{Contrast}
 & Low & \cellcolor{green!8}\bfseries 0.098 & 0.050 & -0.017 & 0.031 & \cellcolor{green!8}\bfseries 0.109 & 0.066 & 0.010 & 0.057 & \cellcolor{green!8}\bfseries 12.231 & -0.222 & -10.416 & -3.932 & \cellcolor{green!8}\bfseries 0.365 & 0.254 & 0.111 & 0.253 \\
 & Medium & \cellcolor{green!8}\bfseries 0.062 & 0.001 & -0.025 & 0.023 & \cellcolor{green!8}\bfseries 0.056 & 0.031 & -0.003 & 0.026 & -9.189 & -18.907 & -14.150 & \bfseries -7.658 & \cellcolor{green!8}\bfseries 0.188 & 0.117 & 0.032 & 0.117 \\
 & High & \cellcolor{green!8}\bfseries 0.029 & 0.008 & -0.101 & -0.001 & \cellcolor{green!8}\bfseries 0.017 & 0.004 & -0.047 & -0.005 & \cellcolor{green!8}\bfseries -2.597 & -3.729 & -21.972 & -7.813 & \cellcolor{green!8}\bfseries 0.090 & 0.049 & -0.079 & 0.033 \\
\midrule
\multirow{3}{*}{Thinness}
 & Compact & \cellcolor{green!8}\bfseries 0.075 & 0.021 & -0.027 & 0.025 & \cellcolor{green!8}\bfseries 0.072 & 0.042 & -0.002 & 0.035 & \cellcolor{green!8}\bfseries -1.002 & -11.091 & -13.704 & -6.484 & \cellcolor{green!8}\bfseries 0.243 & 0.162 & 0.052 & 0.161 \\
 & Moderate & \cellcolor{green!8}\bfseries 0.093 & 0.059 & -0.022 & 0.022 & \cellcolor{green!8}\bfseries 0.142 & 0.087 & 0.013 & 0.077 & \cellcolor{green!8}\bfseries 9.246 & 2.739 & -7.940 & -4.715 & \cellcolor{green!8}\bfseries 0.432 & 0.309 & 0.110 & 0.285 \\
 & Thin & -0.004 & -0.005 & \bfseries 0.005 & -0.001 & \cellcolor{green!8}\bfseries 0.101 & 0.029 & 0.041 & 0.032 & \cellcolor{green!8}\bfseries 31.396 & 5.397 & 5.937 & 8.716 & \cellcolor{green!8}\bfseries 0.454 & 0.165 & 0.211 & 0.191 \\
\midrule
\multirow{3}{*}{Object Size}
 & Medium & \cellcolor{green!8}\bfseries 0.117 & 0.055 & -0.032 & 0.017 & \cellcolor{green!8}\bfseries 0.102 & 0.062 & -0.003 & 0.040 & \cellcolor{green!8}\bfseries 12.413 & -1.423 & -13.897 & -7.253 & \cellcolor{green!8}\bfseries 0.343 & 0.232 & 0.071 & 0.202 \\
 & Small & \cellcolor{green!8}\bfseries 0.040 & 0.015 & -0.095 & -0.070 & \cellcolor{green!8}\bfseries 0.035 & 0.020 & -0.036 & -0.027 & \cellcolor{green!8}\bfseries 8.302 & 2.795 & -23.057 & -16.865 & \cellcolor{green!8}\bfseries 0.111 & 0.074 & -0.043 & -0.009 \\
 & Large & 0.019 & -0.028 & -0.001 & \bfseries 0.062 & 0.045 & 0.021 & 0.011 & \bfseries 0.049 & -21.197 & -27.251 & -9.489 & \bfseries -1.715 & 0.154 & 0.094 & 0.062 & \bfseries 0.158 \\
\bottomrule
\end{tabular}%
}
\end{table*}

\begin{table}[t]
\centering
\caption{Single next-prompt evaluation under a fixed shared seed of size $k$ for the three uncertainty-based strategies only. All strategies start from the same $k$ shared-seed prompts; each then selects one additional location according to its own acquisition rule. \textbf{Mean} IoU, ECE, HD95, and NLL are reported over the medical test set. \textbf{Bold} = best non-GT strategy. Green shading highlights BALD Laplace only when it is the best non-GT strategy in that column.}
\label{tab:shared_seed_next_prompt_trio}
\setlength{\tabcolsep}{5pt}
\begin{tabular}{llcccc}
\toprule
$k$ & Strategy & IoU$\uparrow$ & ECE$\downarrow$ & HD95$\downarrow$ & NLL$\downarrow$ \\
\midrule
\multirow{4}{*}{3}
  & BALD Laplace    & {\cellcolor{green!8}\bfseries 0.432} & {\cellcolor{green!8}\bfseries 0.429} & 97.6 & {\cellcolor{green!8}\bfseries 0.946} \\
  & BALD-MCD & 0.413                            & 0.441                            & 100.6 & 0.980 \\
  & Entropy         & 0.420                            & 0.431                            & {\bfseries 96.2} & 0.953 \\
\midrule
  & GT-based        & 0.460                            & 0.396                            & 87.6 & 0.842 \\
\midrule
\multirow{4}{*}{5}
  & BALD Laplace    & {\cellcolor{green!8}\bfseries 0.400} & 0.455                            & {\cellcolor{green!8}\bfseries 106.7} & 1.003 \\
  & BALD-MCD & 0.389                            & 0.462                            & 109.1 & 1.022 \\
  & Entropy         & 0.393                            & {\bfseries 0.452}                & 107.3 & {\bfseries 0.994} \\
\midrule
  & GT-based        & 0.425                            & 0.423                            & 99.6 & 0.903 \\
\midrule
\multirow{4}{*}{8}
  & BALD Laplace    & {\cellcolor{green!8}\bfseries 0.319} & {\cellcolor{green!8}\bfseries 0.491} & {\cellcolor{green!8}\bfseries 126.3} & {\cellcolor{green!8}\bfseries 1.073} \\
  & BALD-MCD & 0.310                            & 0.497                            & 127.0 & 1.090 \\
  & Entropy         & 0.302                            & 0.495                            & 127.9 & 1.074 \\
\midrule
  & GT-based        & 0.333                            & 0.472                            & 124.9 & 1.006 \\
\bottomrule
\end{tabular}
\end{table}

\begin{table*}[t]
\centering
\caption{IoU-based comparison of SAM1, SAM2, and SAM3 after the first prompt and after BALD-based interactive prompting. All three SAM versions use the same head and Laplace posterior trained on SAM1-precomputed logits; no retraining or adaptation is performed for SAM2 or SAM3. This table measures out-of-distribution transfer of the SAM1 acquisition map to later-generation backbones. The first prompt is placed at the centroid of the ground-truth foreground, so IoU@0 is identical before either BALD strategy starts. Results are from a single seed per backbone. For consistency with the other gain-based summary tables, trajectory summaries are reported as endpoint improvement and AUC improvement from step 0, whereas the raw endpoint score is reported explicitly as final IoU at the full prompting budget. Higher is better for all reported quantities. Best values within each row are shown in bold.}
\label{tab:sam123_iou_comparison}

\footnotesize
\setlength{\tabcolsep}{8pt}
\renewcommand{\arraystretch}{1.15}

\begin{tabular}{llccc}
\toprule
\textbf{Prompting Setting}
& \textbf{Metric}
& \textbf{SAM1}
& \textbf{SAM2}
& \textbf{SAM3} \\
\midrule

\textbf{First prompt}
& IoU@0 \textbf{(before strategy starts)}
& 0.534
& 0.684
& \textbf{0.718} \\
\midrule

\multirow{3}{*}{\textbf{BALD Laplace}}
& Final IoU
& 0.621
& \textbf{0.697}
& 0.427 \\
& Improvement (IoU)
& \textbf{0.087}
& 0.013
& -0.291 \\
& AUC Improvement (IoU)
& \textbf{1.392}
& -0.075
& -2.821 \\
\midrule

\multirow{3}{*}{\textbf{BALD-MCD}}
& Final IoU
& 0.534
& \textbf{0.631}
& 0.333 \\
& Improvement (IoU)
& \textbf{-0.001}
& -0.052
& -0.385 \\
& AUC Improvement (IoU)
& \textbf{0.431}
& -1.377
& -4.130 \\

\bottomrule
\end{tabular}
\end{table*}

The ablation study evaluates BALD Laplace on medical imaging data against a broader set of competing strategies and metrics than the main PointPrompt evaluation. The datasets used for this analysis are described in Section~\ref{sec:medical_datasets}. Rather than simply comparing human prompts to BALD Laplace at the trajectory level and final IoU, the ablation probes deeper questions: how competing active strategies behave across the full prompting trajectory in Section~\ref{sec:overall-trajectory}, whether early budget performance is a reliable predictor of long-term quality in Section~\ref{sec:fixed-prompt-budgets}, whether the gains of BALD Laplace are statistically reliable in Section~\ref{sec:statistical-significance}, how strategies fare on structurally difficult images such as thin or low-contrast objects in Section~\ref{sec:morphological-strata}, whether the advantage of BALD Laplace holds in a single-step controlled setting that isolates the acquisition function from trajectory effects in Section~\ref{sec:shared-seed-next-prompt}, and whether these findings are consistent across different SAM backbone versions in Section~\ref{sec:sam-version-generalization}. The comparison set includes BALD-MCD, BALD-MCD$^*$, Entropy, Random, MaxDist, and a GT-based strategy, as defined in Section~\ref{sec:baselines}, and these methods are evaluated jointly under IoU, ECE, HD95, and NLL, as specified in Section~\ref{sec:evaluation-metrics}. Table~\ref{tab:ablation_takeaways} consolidates the key finding from each ablation experiment. Together, these experiments provide a more granular characterization of uncertainty-based active prompting beyond the primary experiments.

\subsection{Medical Datasets for Ablation Studies}
\label{sec:medical_datasets}

We use five medical imaging datasets spanning breast ultrasound, chest radiography, retinal imaging, dermatology, and gastrointestinal endoscopy: BUSI~\cite{al2020dataset} (780 breast ultrasound images with lesion boundary masks across normal, benign, and malignant categories), CheXmask~\cite{gaggion2024chexmask} (over 676{,}000 chest X-ray anatomical segmentation masks covering lungs and heart, validated by expert physicians), DRIVE~\cite{staal2004ridge} (40 retinal fundus photographs with manually traced vessel segmentation masks), ISIC 2018~\cite{codella2019skin,tschandl2018ham10000} (2{,}594 dermoscopic images with lesion boundary masks from the HAM10000 collection), and Kvasir-SEG~\cite{jha2020kvasir} (1{,}000 colonoscopy polyp images with masks annotated by a medical doctor and verified by a gastroenterologist). Together these datasets cover a wide range of object morphologies, contrast levels, and boundary complexities representative of real clinical imaging conditions. Image-level overlap between PointPrompt training images and the medical evaluation sets was not explicitly audited. A direct audit is not possible because PointPrompt does not release the official filenames or sample indices that map its curated images back to the original databases. It was assembled by curating 400 images per category from publicly available sources, without identifying those specific files in the release. PointPrompt contains at most 400 images per medical category, whereas our evaluation uses the full published test splits (630 for BUSI, 247 for CheXmask, 200 each for ISIC 2018 and Kvasir-SEG, and 20 for DRIVE). Under worst-case assumptions, up to 800 of our 1{,}297 test images (up to 400 BUSI, 200 ISIC 2018, and 200 Kvasir-SEG images) could therefore coincide with PointPrompt images. Image-level duplicates cannot be ruled out for BUSI, ISIC 2018, and Kvasir-SEG; absolute performance values on those datasets should therefore be interpreted with caution and are not claimed to reflect contamination-free generalization. Because BALD Laplace, BALD-MCD, and Entropy use the same trained head, and BALD-MCD$^*$ uses a head retrained on the same data with identical hyperparameters, overlap is unlikely to explain relative differences among these head-based acquisition rules. Overlap remains a limitation for absolute medical performance estimates and for comparisons against head-free strategies (Random, MaxDist, GT-based). For evaluation, we use fixed evaluation splits from our preprocessing pipeline, yielding test-set sizes of 630 (BUSI), 247 (CheXmask), 20 (DRIVE), 200 (ISIC 2018), and 200 (Kvasir-SEG), totalling 1{,}297 test images. These splits are fixed across all strategies. Unless otherwise stated, step~0 denotes the SAM prediction after a single initial positive point prompt placed at the centroid of the ground-truth foreground; active acquisition strategies select prompts from step~1 onward. Evaluation scripts and trained model artifacts are available upon request from the corresponding author.

\subsection{Baselines}
\label{sec:baselines}

The medical ablation benchmarks BALD Laplace against five automatic prompt-selection baselines spanning uncertainty-driven, diversity-based, sampling-based, and reference prompting strategies, as defined below.

All strategies except BALD-MCD$^*$ are drawn from the same experimental run using 4 independent random seeds. BALD Laplace and BALD-MCD each use $K{=}30$ posterior samples, the Pareto-optimal operating point identified in Table~\ref{tab:laplace_samples_interaction}. The same 4 seeds are used for all non-posterior baselines (Random, MaxDist, Entropy). BALD-MCD$^*$ comes from a separate run using the properly trained dropout head, with 4 seeds and $K{=}30$ posterior samples; the two runs share the same initial prompt coordinates.

The baselines used in this evaluation are defined below.

\paragraph{Entropy.}
This baseline selects the pixel with the largest predictive entropy. For each candidate pixel $i$, let $p(i)$ be the predicted foreground probability. The entropy score is
\begin{equation}
\mathrm{Ent}(i)=H(p(i)).
\end{equation}
The selected point is
\begin{equation}
a_t=\arg\max_{i\in\Omega_t} \mathrm{Ent}(i).
\end{equation}
Unlike BALD Laplace, this criterion uses only predictive uncertainty and does not separate epistemic uncertainty from aleatoric uncertainty.

\paragraph{Random.}
This baseline selects a valid candidate pixel uniformly at random:
\begin{equation}
a_t \sim \mathrm{Uniform}(\Omega_t).
\end{equation}
It provides a non-adaptive reference for measuring whether a prompting strategy improves over uninformed point selection.

\paragraph{MaxDist.}
This baseline selects the valid candidate pixel farthest from all existing prompt points. Let $P_t$ be the set of prompt locations already used by step $t$. The score for candidate pixel $i$ is
\begin{equation}
d(i,P_t)=\min_{p\in P_t} \|i-p\|_2,
\end{equation}
where $\|\cdot\|_2$ denotes Euclidean distance. The selected point is
\begin{equation}
a_t=\arg\max_{i\in\Omega_t} d(i,P_t).
\end{equation}
This baseline encourages spatial coverage rather than uncertainty-based targeting.

\paragraph{BALD-MCD (Post-Hoc MC Dropout).}
This baseline uses the same trained prompt-aware segmentation head as BALD Laplace, but changes the source of uncertainty used for acquisition. Instead of fitting and sampling from a Laplace posterior over the final layer, it keeps the trained head parameters fixed and applies stochastic dropout before the final convolution during acquisition.

For a candidate pixel $q\in\Omega_t$, we run $K$ stochastic forward passes with dropout enabled. Let
\begin{equation}
p^{\mathrm{drop}}_k(q)
=
p(\ell_q=1\mid \mathcal{I},\mathcal{S}_t,\hat{\bm{\theta}}_{\mathrm{head}},\xi_k),
\qquad k=1,\ldots,K,
\end{equation}
where $\hat{\bm{\theta}}_{\mathrm{head}}$ denotes the trained head parameters and $\xi_k$ denotes the dropout mask used in the $k$-th stochastic forward pass. The Monte Carlo predictive mean is
\begin{equation}
\bar{p}^{\mathrm{drop}}(q)
=
\frac{1}{K}\sum_{k=1}^{K}p^{\mathrm{drop}}_k(q).
\end{equation}
Using the same binary entropy function
\begin{equation}
h_2(p)=-p\log p-(1-p)\log(1-p),
\end{equation}
the dropout BALD score is
\begin{equation}
\mathrm{MI}_{\mathrm{drop}}(q)
=
h_2\!\left(\bar{p}^{\mathrm{drop}}(q)\right)
-
\frac{1}{K}\sum_{k=1}^{K}
h_2\!\left(p^{\mathrm{drop}}_k(q)\right).
\end{equation}
The selected point is
\begin{equation}
q_{t+1}
=
\arg\max_{q\in\Omega_t}
\mathrm{MI}_{\mathrm{drop}}(q).
\end{equation}
Thus, BALD-MCD uses the same BALD mutual-information objective and the same trained segmentation head as our method. The difference is only the uncertainty model used to form the acquisition map: BALD Laplace samples final-layer parameters from a fitted diagonal Laplace posterior, whereas BALD-MCD uses post-hoc stochastic dropout passes and does not fit an explicit parameter posterior.

\paragraph{BALD-MCD$^*$ (Properly Trained MC Dropout).}
The BALD-MCD baseline above applies dropout post-hoc to a head trained without dropout during training. To provide a properly trained MC-Dropout comparison, we retrained the head with 2D spatial dropout ($p{=}0.1$) inserted after each ReLU activation, using identical hyperparameters (AdamW, lr$=10^{-3}$, BCE+Dice loss, cosine scheduler, up to 50 epochs, early stopping patience 10 on val IoU). This properly trained head achieves best val IoU of 0.6794, matching the original deterministic head. (This val IoU measures raw segmentation quality of the head on the validation set and is distinct from the acquisition-improvement IoU values in Table~\ref{tab:laplace_samples_interaction}, which measure prompt-selection gain relative to a random baseline.) Dropout is kept active at both training and inference time; acquisition uses the same MC mutual-information formula as BALD-MCD. The result is denoted BALD-MCD$^*$ throughout. Key results: BALD-MCD$^*$ achieves IoU improvement $-0.086$ at full budget versus $+0.081$ for BALD Laplace. At all fixed budgets ($k\,{=}\,5$: 0.543; $k\,{=}\,10$: 0.522; $k\,{=}\,15$: 0.480; full: 0.449), BALD-MCD$^*$ falls below both BALD Laplace and the post-hoc BALD-MCD. We hypothesize that internal Dropout2d modulates feature activations rather than inducing posterior uncertainty over final-layer weights, producing acquisition maps less well calibrated for epistemic disagreement in this setting. This outcome is consistent with the BALD Laplace advantage being robust to a properly trained dropout comparison.

\paragraph{GT-Based.}
This baseline uses the ground-truth mask to identify the largest current error region and selects its centroid. Let $M^\star$ be the ground-truth mask and $\hat{M}_t$ be the predicted mask at step $t$. The error mask is
\begin{equation}
E_t = \hat{M}_t \triangle M^\star,
\end{equation}
where $\triangle$ denotes symmetric difference. Let $C_t$ be the largest connected component in $E_t$. The selected point is
\begin{equation}
a_t = \mathrm{centroid}(C_t).
\end{equation}
GT-based is not a globally optimal oracle; it is a greedy heuristic that prompts the centroid of the largest current error region. We interpret its occasional underperformance through three plausible limitations. First, the centroid need not lie inside the object or even the error region for non-convex or annular shapes, making the prompt geometrically uninformative. Second, the strategy is purely reactive: it targets where the mask is wrong but ignores model uncertainty, so it may repeatedly prompt the same boundary \textsc{SAM} cannot resolve rather than querying a structurally distinct region. Third, \textsc{SAM}'s response depends on the full prompt context $\mathcal{S}_t$, not the error map alone; maximally correcting the current pixel error may not reduce \textsc{SAM}'s internal uncertainty, especially when prior prompts have saturated attention heads or collapsed the embedding toward a local mode. \textsc{BALD-SAM} targets epistemic disagreement directly, so cases where \textsc{BALD} exceeds GT-based should be read not as an oracle failure, but as evidence that greedy error-centroid targeting is an imperfect proxy for informativeness under \textsc{SAM}'s architecture. A stronger reference would be a one-step lookahead oracle that selects the candidate maximizing the resulting IoU; evaluating such an oracle is beyond the scope of this study.

\subsection{Evaluation Metrics}
\label{sec:evaluation-metrics}

The inference path differs by metric. IoU and Dice (Tables~\ref{tab:BALD Laplace_vs_human_iou_dice}, \ref{tab:miou_with_d2q}, \ref{tab:overall_summary}, and~\ref{tab:overall_scores}) are computed from SAM's native mask decoder output conditioned on the accumulated prompt set $\mathcal{S}_t$; the head's output does not contribute to these reported segmentation masks. When SAM returns multiple mask hypotheses, the mask with the highest SAM-predicted quality score is selected; the same rule is used for SAM1, SAM2, and SAM3. ECE and NLL (Tables~\ref{tab:overall_summary} and~\ref{tab:overall_scores}) are computed from the Laplace probit predictive probability of the head, since SAM's binary mask output is not a calibrated probability. These two metrics therefore jointly measure Laplace calibration and head refinement quality. HD95 (Tables~\ref{tab:overall_summary} and~\ref{tab:overall_scores}) is computed from SAM's binary mask, the same source as IoU.

For our extended ablation studies, we additionally report ECE, HD95, and NLL along with IoU (already defined in Section \ref{sec:main-evaluation-metrics}) to assess calibration, boundary accuracy, and probabilistic prediction quality. Similar to our primary experiments these metrics are also observed across iterative prompting steps, hence we summarize the resulting trajectory using endpoint improvement, mean, and budget area under the curve as previously defined in Section \ref{summary-stats}. In the ablation studies, we further quantify uncertainty around the ranking comparisons using paired statistical analysis. Since all prompting strategies are evaluated on the same test images, comparisons are made using paired per-image differences. For each metric-summary pair, we report the direction-aligned paired mean difference $\Delta$, where positive values favor BALD Laplace. We estimate 95\% confidence intervals (CIs) using paired bootstrap resampling over test images and report two-sided sign-flip permutation $p$-values to assess whether the paired differences are distinguishable from zero.

The extended metrics measured as follows.

\subsubsection{Extended Core Metrics}

\paragraph{Expected Calibration Error (ECE).}
ECE measures the mismatch between predicted confidence and empirical correctness. We divide predictions into $B{=}10$ equal-width confidence bins. Let $\mathcal{B}_b$ denote the set of pixels whose confidence falls in bin $b$, and let $N$ be the total number of evaluated pixels. For bin $b$, let $\mathrm{acc}(\mathcal{B}_b)$ denote the empirical accuracy and $\mathrm{conf}(\mathcal{B}_b)$ denote the average predicted confidence. ECE at prompting step $t$ is
\begin{equation}
\mathrm{ECE}_t
=
\sum_{b=1}^{B}
\frac{|\mathcal{B}_b|}{N}
\left|
\mathrm{acc}(\mathcal{B}_b)
-
\mathrm{conf}(\mathcal{B}_b)
\right|.
\end{equation}
Lower ECE indicates better calibration. Because ECE is computed over all pixels, it is dominated by the many confidently predicted background pixels in foreground-sparse images; we therefore use it to compare strategies relative to one another rather than as an absolute measure of calibration quality.

\paragraph{95th-Percentile Hausdorff Distance (HD95).}
HD95 measures boundary error while reducing sensitivity to extreme outliers. Let $\partial \hat{M}_t$ and $\partial M^\star$ denote the boundaries of the predicted and ground-truth masks. For a boundary point $u$, define its distance to a boundary set $A$ as
\begin{equation}
d(u,A)=\min_{v\in A}\|u-v\|_2.
\end{equation}
HD95 is computed as the 95th percentile of the bidirectional boundary distances between $\partial \hat{M}_t$ and $\partial M^\star$. Lower HD95 indicates better boundary alignment.

\paragraph{Negative Log-Likelihood (NLL).}
NLL measures probabilistic prediction quality. Let $y_i\in\{0,1\}$ be the ground-truth label for pixel $i$, and let $p_i$ be the predicted foreground probability. For $N$ evaluated pixels, NLL at prompting step $t$ is
\begin{equation}
\mathrm{NLL}_t
=
-\frac{1}{N}
\sum_{i=1}^{N}
\left[
y_i\log p_i
+
(1-y_i)\log(1-p_i)
\right].
\end{equation}
Probabilities are clipped to $[10^{-7},\,1-10^{-7}]$ before computing NLL. Lower NLL indicates better probabilistic prediction quality.

\subsection{Overall Performance Across the Prompting Trajectory}
\label{sec:overall-trajectory}

Table~\ref{tab:overall_summary} reports endpoint improvement, AUC improvement from step~0, and mean improvement from step~0 averaged across all medical datasets. BALD Laplace is the best overall method in 9 of 12 columns and exceeds the GT-based strategy in 11 of 12 columns. Under IoU, BALD Laplace leads all strategies on every summary: improvement 0.081, AUC 1.442, and mean 0.074, against GT-based values of 0.029, 0.819, and 0.042. For HD95, BALD Laplace leads all three summary cells. For ECE, it leads improvement and mean, whereas Random is slightly higher on AUC. Under NLL, BALD Laplace leads GT-based on improvement (0.328 vs.\ 0.297) and mean (0.253 vs.\ 0.252), whereas Random edges ahead of BALD Laplace on ECE AUC (1.466 vs.\ 1.449), NLL AUC (5.331 vs.\ 4.894), and NLL Mean (0.272 vs.\ 0.253). These Random leads are consistent with an early-budget effect that diminishes as the prompt budget grows, as shown by the per-budget breakdown in Table~\ref{tab:overall_scores}. MaxDist performs worst across nearly every metric and budget level, indicating that spatial diversity alone is not a reliable criterion for prompt selection in medical imaging. BALD-MCD$^*$ (properly trained dropout) achieves IoU improvement $-0.086$, worse than both BALD-MCD ($-0.001$) and BALD Laplace ($+0.081$), indicating that the BALD Laplace advantage is not attributable to a weak post-hoc dropout comparison. Per-dataset IoU endpoint improvements and the per-dataset SAM-version breakdown are reported in Appendix~\ref{app:per-dataset-results}.

\subsection{Performance at Fixed Prompt Budgets}
\label{sec:fixed-prompt-budgets}

Table~\ref{tab:overall_scores} tracks mean performance at prompt budgets of $k = 5$, $k = 10$, $k = 15$, and full budget. Random prompting achieves the best IoU at $k = 5$, but its performance degrades sharply with each additional prompt, falling from 0.660 at $k = 5$ to 0.462 at full budget. BALD Laplace, by contrast, remains stable across budget levels, finishing at 0.616 IoU at full budget while remaining the best non-GT strategy from $k = 10$ onward. The relative IoU change for BALD Laplace from $k=5$ to full budget is $-1.9\%$ (0.628 to 0.616); competing strategies degrade far more severely. Random falls 30\% (0.660 to 0.462) and Entropy falls 9\% (0.584 to 0.533). The $k{=}5$ Random lead is consistent with an early-budget effect in active learning: random spatial coverage can outperform uncertainty-based selection when the model has insufficient prompt context to estimate informative disagreement. This advantage disappears by $k{=}10$ (Random: 0.581 vs.\ BALD Laplace: 0.620). For ECE, BALD Laplace improves as the budget increases. For HD95, BALD Laplace does not improve monotonically, but it remains the best non-GT strategy from $k = 10$ onward and degrades less than the competing strategies. The pooled HD95 worsening for BALD Laplace is almost entirely attributable to CheXmask: on the four remaining datasets, HD95 improvement is positive for BALD Laplace at all budget levels (BUSI $+21.3$, DRIVE $+55.6$, ISIC 2018 $+11.2$, Kvasir-SEG $+11.7$). In deployment, the MI stopping criterion ($\max \mathrm{MI}(q) \leq \delta = 0.01$) provides a natural exit once the model's epistemic uncertainty is resolved; for a consistent comparison, every strategy in these tables uses the full 20-prompt budget.

\subsection{Statistical Significance of BALD Laplace Gains}
\label{sec:statistical-significance}

Table~\ref{tab:overall_summary_fixed_rivals_stats} provides paired comparisons between BALD Laplace and each competing strategy across all metrics. Every comparison involving BALD-MCD, Entropy, and MaxDist is statistically significant at $p < 10^{-4}$ across all three summaries and all four metrics. The comparison against BALD-MCD$^*$ is also significant at $p < 10^{-4}$ for all 12 metric-summary pairs and yields the largest IoU endpoint gap among the uncertainty-based rivals ($\Delta = 0.167$, 95\% CI $[0.150,\,0.185]$), consistent with a substantially weaker acquisition signal from the properly trained dropout head. Against Random, BALD Laplace is significantly better on every IoU and HD95 summary and on NLL endpoint improvement. On NLL AUC ($p = 0.0001$) and NLL Mean ($p = 0.0005$), the difference is significant in favor of Random, reflecting its early-budget advantage. The only non-significant comparisons are ECE AUC vs.\ Random ($p = 0.749$) and ECE Mean vs.\ Random ($p = 0.744$), both of which correspond to the early-budget Random advantage in ECE discussed above. Table~\ref{tab:overall_summary_fixed_rivals_stats} contains 60 paired tests; after Holm correction for multiple comparisons (equivalently, under a Bonferroni threshold of $0.05/60 \approx 8.3\times10^{-4}$), the same 58 comparisons remain significant and the same two ECE comparisons against Random remain non-significant, so no conclusion changes. Paired bootstrap confidence intervals are computed by resampling test images with replacement at the fixed operating point $K{=}30$. Point estimates are averaged over 4 independent random seeds; the seeds reduce point-estimate variance but are not the bootstrap replication unit. $K$-sensitivity is reported separately in Table~\ref{tab:laplace_samples_interaction}.

\subsection{Performance Across Morphological Strata}
\label{sec:morphological-strata}

Table~\ref{tab:strata_defs} in the appendix defines the four strata families with exact formulas, thresholds, and test-set sample counts. Perimeter $P$ is approximated by binary erosion; area $A$ is the foreground pixel count; all measurements are on the $256{\times}256$ resized mask or image. The Complex-boundary and Thin strata each contain only 22 images (1.7\% of the test set), so results in those cells should be read with caution.

Table~\ref{tab:strata_dominance} reports mean improvement stratified by four image properties: boundary complexity, contrast, thinness, and object size. Representative examples for each stratum are provided in Appendix~\ref{app:strata-examples}. BALD Laplace wins 41 of 48 stratum--metric cells; excluding the two 22-image strata (Complex boundary and Thin), it wins 35 of the remaining 40 cells. Of the seven cells it does not win, BALD-MCD$^*$ wins two (Complex-boundary IoU and Thin-object IoU) and Entropy wins five (Medium contrast HD95 and all four Large object metrics). In the Complex-boundary and Thin-object strata under IoU, BALD-MCD$^*$ achieves a marginally positive improvement whereas BALD Laplace is marginally negative ($+0.005$ vs.\ $-0.004$); BALD Laplace leads on ECE, HD95, and NLL within both of these strata by a wide margin, so the IoU deficit does not reflect a general failure. The Medium contrast HD95 loss is Entropy at $-7.658$ vs.\ BALD Laplace at $-9.189$. In the Large object stratum, Entropy leads on all four metrics. Large objects present relatively easy segmentation targets where a single well-placed early prompt already captures most of the target, reducing the information available for uncertainty-guided refinement; Entropy's advantage here is consistent with that interpretation. Outside of these cases, BALD Laplace leads in every stratum, with particularly large margins in the Low contrast, Moderate boundary, Moderate thinness, and Small object strata, settings where target structure is ambiguous and uncertainty estimation can be expected to help most over uninformed selection.

\subsection{Next-Prompt Selection Under a Shared Seed}
\label{sec:shared-seed-next-prompt}

Motivated by \cite{chakrabarty2025comparing}, Table~\ref{tab:shared_seed_next_prompt_trio} evaluates a controlled single-step scenario: all three uncertainty-based strategies (BALD Laplace, BALD-MCD, Entropy) begin from the same $k$ shared seed prompts and each selects exactly one additional prompt according to its acquisition rule. BALD-MCD$^*$ is excluded from this table because the controlled shared-seed experiment was run before the dropout head was trained; including BALD-MCD$^*$ would require a separate controlled run with matching seed conditions. BALD Laplace is the best non-GT strategy in 9 of 12 cells across seed sizes $k \in \{3, 5, 8\}$. At $k = 8$, it leads all four metrics simultaneously. This controlled comparison isolates the quality of the acquisition function itself from trajectory effects.

Seed prompts are generated deterministically from the $256{\times}256$ ground-truth mask using furthest-point sampling: $\lceil k/2 \rceil$ points from the foreground region serve as positive (inclusion) prompts and $\lfloor k/2 \rfloor$ points from the background region serve as negative (exclusion) prompts. All compared strategies receive the identical seed context; the same seed integer (seed$=0$) is used for all images and all strategies. A single seed configuration per image is evaluated (a known limitation), at $k \in \{3, 5, 8\}$. This differs from model-independent benchmarking protocols in that the seed points are image-specific (drawn from each image's ground-truth mask) and the comparison isolates only the next-step decision, not the full trajectory.

\subsection{Generalization Across SAM Versions}
\label{sec:sam-version-generalization}

All three SAM versions use the same head and Laplace posterior trained on SAM1-precomputed logits; no retraining or adaptation is performed for SAM2 or SAM3. Table~\ref{tab:sam123_iou_comparison} is therefore a single-seed, out-of-distribution transfer probe measuring how well the SAM1 acquisition map transfers to later-generation backbones, not an optimized per-version evaluation.

Table~\ref{tab:sam123_iou_comparison} compares SAM1, SAM2, and SAM3 under the same first-prompt and BALD-based prompting setup. The first prompt is placed at the centroid of the ground-truth foreground for all backbones, so IoU@0 measures the starting quality before either BALD strategy begins. SAM3 has the best starting point, with IoU@0 of $0.718$, compared with $0.684$ for SAM2 and $0.534$ for SAM1. However, better initial performance does not lead to better active-prompting gains in this probe. With BALD Laplace, SAM1 improves from $0.534$ to $0.621$, giving the largest endpoint gain of $0.087$ and the largest AUC gain of $1.392$. SAM2 reaches the best final IoU, $0.697$, but improves only slightly from its starting IoU of $0.684$. SAM3 degrades sharply, dropping from $0.718$ to $0.427$, with an endpoint change of $-0.291$ and an AUC change of $-2.821$. BALD-MCD shows the same trend but performs worse overall. SAM1 remains almost unchanged, SAM2 drops from $0.684$ to $0.631$, and SAM3 drops further to $0.333$. This suggests that a newer or stronger SAM model is not automatically better for active prompting with a fixed acquisition head. Per-dataset results in Table~\ref{tab:per_dataset_sam123} show that the SAM3 collapse concentrates in BUSI ($-0.239$), CheXmask ($-0.510$), and Kvasir-SEG ($-0.352$), whereas DRIVE is the only dataset where SAM3 gains ($+0.022$), possibly because its low starting IoU leaves epistemic uncertainty for BALD to exploit. Overall, BALD Laplace helps most on SAM1, gives only a small final gain on SAM2, and hurts SAM3 despite SAM3 having the best first-prompt IoU. One possible explanation is that the SAM1-trained head and Laplace posterior are mismatched to SAM3's logit geometry; another is that SAM3's stronger first-prompt masks leave less exploitable epistemic variation. Because the probe uses a single seed and a single configuration per backbone, we treat the SAM3 result as an observation to be confirmed rather than an optimized SAM3 evaluation; these observations are consistent with findings on backbone-specific epistemic uncertainty in medical imaging contexts~\cite{sengupta2025sam, chakrabarty2025comparing}.

\begin{table*}[t]
\centering
\caption{Summary of ablation findings. Each row corresponds to one experiment and states the principal conclusion.}
\label{tab:ablation_takeaways}
\small
\begin{tabularx}{\linewidth}{@{} l X @{}}
\toprule
\textbf{Experiment} & \textbf{Finding} \\
\midrule
Overall trajectory performance
  & BALD Laplace is the best overall method in 9 of 12 trajectory-summary columns and exceeds the greedy GT-error-centroid reference in 11 of 12, with BALD Laplace as the strongest automatic strategy across most trajectory summaries, whereas Random remains competitive on some early-budget calibration summaries. BALD-MCD$^*$ (properly trained dropout) achieves IoU improvement $-0.086$, showing the BALD Laplace advantage holds against a proper MC-Dropout baseline. \\
\addlinespace
Fixed prompt budgets
  & BALD Laplace remains stable across budget levels and is the best non-GT strategy from $k{=}10$ onward, whereas Random leads at $k{=}5$ but degrades sharply; this makes BALD Laplace a suitable choice when the prompt budget is unknown or variable at deployment time. \\
\addlinespace
Statistical significance
  & BALD Laplace significantly outperforms BALD-MCD, BALD-MCD$^*$, Entropy, and MaxDist across all metric-summary pairs at $p < 10^{-4}$, and conclusions are unchanged after Holm correction. The BALD Laplace vs.\ BALD-MCD$^*$ gap is the largest IoU endpoint difference among the uncertainty-based rivals ($\Delta{=}0.167$). BALD Laplace also significantly outperforms Random on IoU and HD95; Random is significantly better on NLL AUC/Mean, and ECE AUC/Mean vs.\ Random are not significant. \\
\addlinespace
Per-dataset results
  & BALD Laplace is positive on 3 of 5 datasets (BUSI, ISIC 2018, Kvasir-SEG). BALD-MCD$^*$ is positive only on DRIVE and Kvasir-SEG. CheXmask is negative for every strategy, likely reflecting a SAM performance ceiling (IoU@0$=0.909$) rather than a BALD-specific failure. BALD Laplace also leads under equal-weight dataset aggregation ($+0.056$); the BALD-MCD$^*$ equal-weight mean is $-0.056$. \\
\addlinespace
Morphological strata
  & BALD Laplace wins 41 of 48 stratum--metric cells (35 of 40 excluding the two 22-image strata); BALD-MCD$^*$ wins 2 (Complex-boundary IoU and Thin-object IoU) and Entropy wins 5 (Medium-contrast HD95 and all four Large-object metrics). The losses do not indicate a systematic weakness: BALD Laplace leads BALD-MCD$^*$ on ECE, HD95, and NLL within those two strata. \\
\addlinespace
Next-prompt selection (shared seed)
  & The Laplace approximation produces the strongest single-step prompt decisions overall, winning 9 of 12 shared-seed comparisons and all four metrics at $k = 8$, suggesting that acquisition function quality, rather than trajectory compounding alone, underlies the overall BALD Laplace advantage. \\
\addlinespace
SAM backbone generalization
  & In this single-seed transfer probe, stronger first-prompt performance does not guarantee better active prompting. SAM3 starts with the highest IoU@0 but degrades under BALD-based prompting; SAM1 gains the most from BALD Laplace, and SAM2 achieves the best final IoU with only a small endpoint gain. \\
\bottomrule
\end{tabularx}
\end{table*}

\section{Future Work}

A natural next direction is to understand how epistemic uncertainty evolves as the frozen backbone is adapted to the target domain. Our SAM1 versus SAM3 result already hints at this: a stronger first prompt leaves less informative uncertainty for BALD to exploit, and the same effect should be even more pronounced for medically fine-tuned backbones such as MedSAM~\cite{ma2024segment} and MedSAM2~\cite{ma2025medsam2}, where large in-domain corpora have absorbed much of the epistemic signal into the weights. A controlled study that tracks the BALD acquisition map at successive checkpoints of a SAM backbone being fine-tuned on medical data would make this trade-off measurable and could guide when active prompting is worth deploying. Beyond fine-tuning, three further directions stand out. The first is supporting mixed prompt modalities, in particular starting from a bounding box that restricts the region of interest and then iterating with informative point queries, which the current framework should accommodate without architectural change but has not been benchmarked. The second is extending the evaluation to additional out-of-distribution domains such as seismic imagery, camouflaged object segmentation, and a broader set of medical modalities, where strong domain shift from SAM's pretraining distribution should make the epistemic signal richer and the gains from active prompting more pronounced. The third, and the one we view as most important, is closing the loop with real human annotators. The current protocol simulates the annotator by reading the inclusion or exclusion label from the ground-truth mask, which is the standard active learning evaluation convention but leaves open whether the same gains hold when a human is in the loop. A user study that replaces the ground-truth labels with real annotator clicks would directly validate the deployment claim and would also surface effects that the current protocol cannot measure, such as how the BALD-suggested locations interact with human visual attention, how often annotators accept or override the suggested point, and whether the per-prompt latency (under 0.1~s per acquisition step on a single NVIDIA TITAN RTX GPU, including the SAM forward pass) is acceptable in practice. Finally, a more detailed quantification of the epistemic uncertainty itself, including how it is distributed across the candidate grid at each step, how it correlates with image features such as boundary complexity and contrast, and how it changes as the prompt set grows, would give a clearer mechanistic picture of why the acquisition rule works when it does and fails when it does, and would tie the empirical SAM1 versus SAM3 contrast to a measurable property of the frozen backbone.

Our evaluation is limited to prompt-aware acquisition over frozen SAM backbones. Direct comparison with iterative click-based methods (e.g., RITM-style networks) and domain-adapted backbones (e.g., MedSAM~\cite{ma2024segment} and SAMed~\cite{zhang2023customized}) under a shared sequential prompting protocol is a natural extension. Such a comparison would clarify whether the epistemic advantage of BALD Laplace is specific to SAM's frozen pretrained representations or extends to fine-tuned architectures where epistemic uncertainty has been partially absorbed into the model weights.

\section{Conclusion}\label{conclusion}

This paper introduced active prompting, a formulation of interactive segmentation as a sequential spatial query problem, and proposed BALD-SAM, an uncertainty-driven method for selecting the next prompt location. BALD-SAM keeps SAM frozen and Bayesianizes only the final layer of a lightweight prompt-aware head using a diagonal Laplace approximation, enabling tractable prompt selection without modifying SAM's pretrained representations. Across PointPrompt and five medical datasets, our experiments support three findings. First, BALD Laplace is competitive with or better than existing prompting strategies: it achieves stronger aggregate improvement relative to recorded PointPrompt human trajectories in the majority of tested categories under this benchmark protocol, ranks among the top two final-IoU methods on 13 of 16 categories, is best overall in 9 of 12 medical trajectory-summary columns, and exceeds the greedy GT-error-centroid reference in 11 of 12 columns. Second, the Laplace posterior yields strong acquisition decisions that do not depend only on trajectory compounding: in the shared-seed experiment, BALD Laplace wins 9 of 12 comparisons and all four metrics at $k=8$. Third, in a single-seed transfer probe, stronger SAM backbones did not automatically improve active prompting. SAM1 benefits most from BALD Laplace, SAM2 reaches the best final IoU with only a small endpoint gain, and SAM3 degrades despite the strongest first-prompt IoU under the same SAM1-trained head. These results indicate that active prompting is both model-dependent and interaction-dependent, governed by the acquisition rule, the backbone's response to additional prompts, and the preservation of useful epistemic uncertainty. Future work should measure how this uncertainty evolves as SAM backbones become stronger or more domain-adapted, for example by tracking BALD acquisition maps across domain-specific fine-tuned models. This would clarify when active prompting remains useful and when uncertainty has been absorbed into adapted weights. Further directions include mixed prompt modalities such as box initialization followed by informative point queries, stronger domain shifts including seismic imagery, camouflaged object segmentation, and additional medical modalities, real human-in-the-loop studies evaluating annotator-provided labels, user acceptance or override, and the practical acceptability of the prompt-selection latency, and a mechanistic analysis of acquisition maps across the candidate grid, boundary complexity, contrast, and prompting steps.

\appendices

\section{Representative Morphology-Strata Examples}
\label{app:strata-examples}

\begin{table}[!t]
\caption{Morphological strata definitions used in
Table~\ref{tab:strata_dominance}. Perimeter $P$ is approximated by binary
erosion as the number of pixels removed from the original mask, and area $A$
is the foreground-pixel count. All measurements are computed on masks and
images resized to $256\times256$ pixels. Contrast is computed as
$C=|\operatorname{mean}(\mathrm{FG})-\operatorname{mean}(\mathrm{BG})|$
using grayscale intensities. Thresholds were fixed a priori and were not
estimated from the test data. The test set contains 1,297 images, and counts
sum to 1,297 within each stratum family. The thin and complex-boundary strata
each contain only 22 samples (1.7\%); therefore, results for these strata
should be interpreted with caution.}
\label{tab:strata_defs}
\centering
\scriptsize
\setlength{\tabcolsep}{2.5pt}
\renewcommand{\arraystretch}{1.12}

\begin{tabularx}{\columnwidth}{@{}
    >{\raggedright\arraybackslash}p{1.37cm}
    >{\raggedright\arraybackslash}p{1.03cm}
    >{\raggedright\arraybackslash}X
    >{\raggedleft\arraybackslash}p{1.35cm}
@{}}
\toprule
\textbf{Stratum} &
\textbf{Level} &
\textbf{Threshold} &
$\bm{n}$ \textbf{(\%)} \\
\midrule

\multirow{3}{1.37cm}{Thinness\\[-1pt]
$\frac{P^2}{4\pi A}$}
& Compact  & $\leq 2.0$       & 1,231 (94.9) \\
& Moderate & $(2.0,\,5.0]$    & 44 (3.4) \\
& Thin     & $>5.0$           & 22 (1.7) \\
\midrule

\multirow{3}{1.37cm}{Boundary\\complexity\\[-1pt]
$\frac{P}{\sqrt{A}}$}
& Simple   & $\leq 5.0$       & 1,230 (94.8) \\
& Moderate & $(5.0,\,10.0]$   & 45 (3.5) \\
& Complex  & $>10.0$          & 22 (1.7) \\
\midrule

\multirow{3}{1.37cm}{Object size\\$A$ (px)}
& Small    & $<1{,}000$             & 123 (9.5) \\
& Medium   & $[1{,}000,\,10{,}000]$ & 706 (54.4) \\
& Large    & $>10{,}000$            & 468 (36.1) \\
\midrule

\multirow{3}{1.37cm}{Contrast\\$C$}
& Low      & $<30$            & 523 (40.3) \\
& Medium   & $[30,\,80]$      & 686 (52.9) \\
& High     & $>80$            & 88 (6.8) \\
\bottomrule
\end{tabularx}
\end{table}

\begin{figure*}[t]
\centering
\subfloat[Boundary Complexity: left-to-right examples for Moderate, Simple, and Complex. Top row shows the medical image; bottom row shows the corresponding ground-truth mask.\label{fig:strata_boundary_complexity_examples}]{\includegraphics[width=0.49\textwidth]{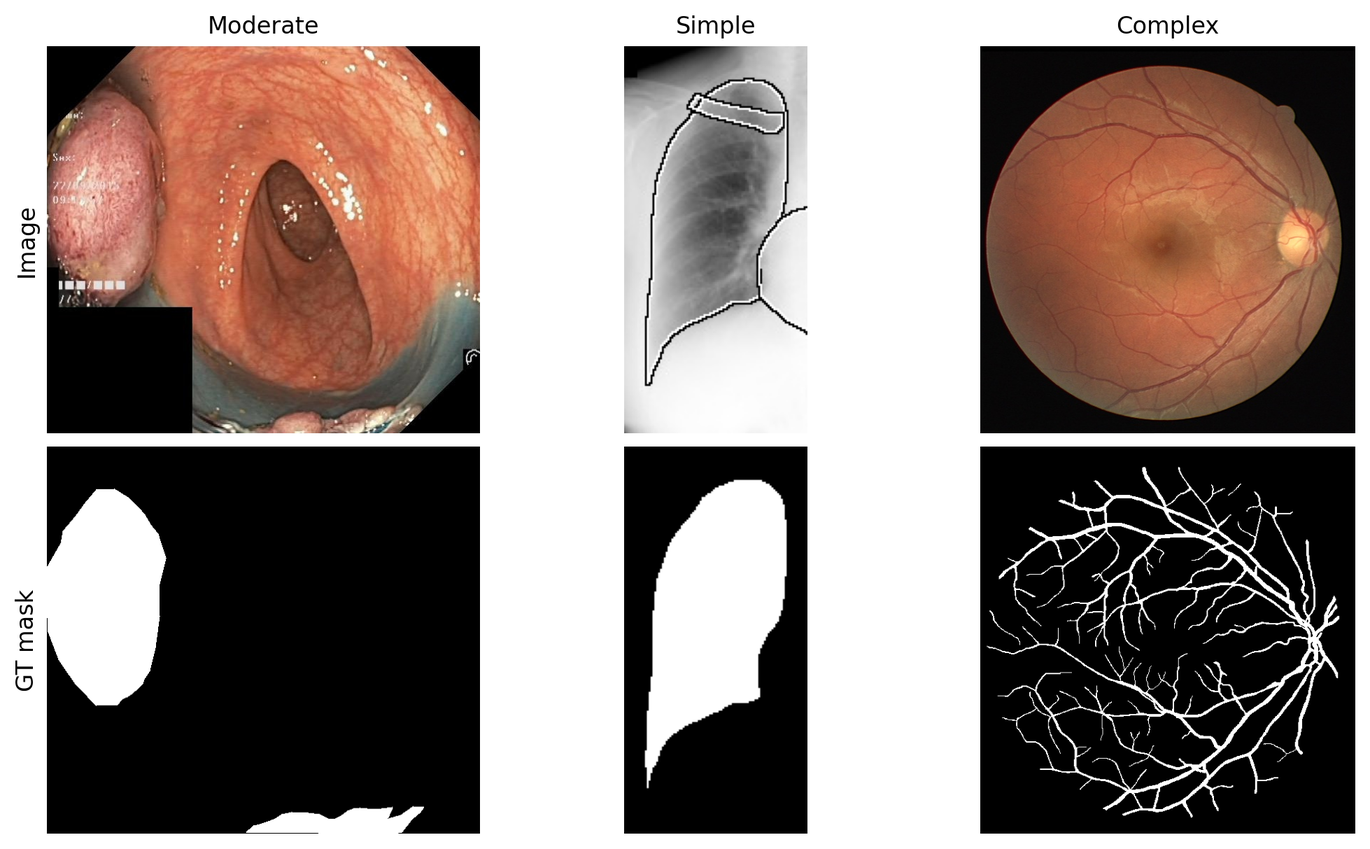}}
\hfill
\subfloat[Contrast: left-to-right examples for Low, Medium, and High. Top row shows the medical image; bottom row shows the corresponding ground-truth mask.\label{fig:strata_contrast_examples}]{\includegraphics[width=0.49\textwidth]{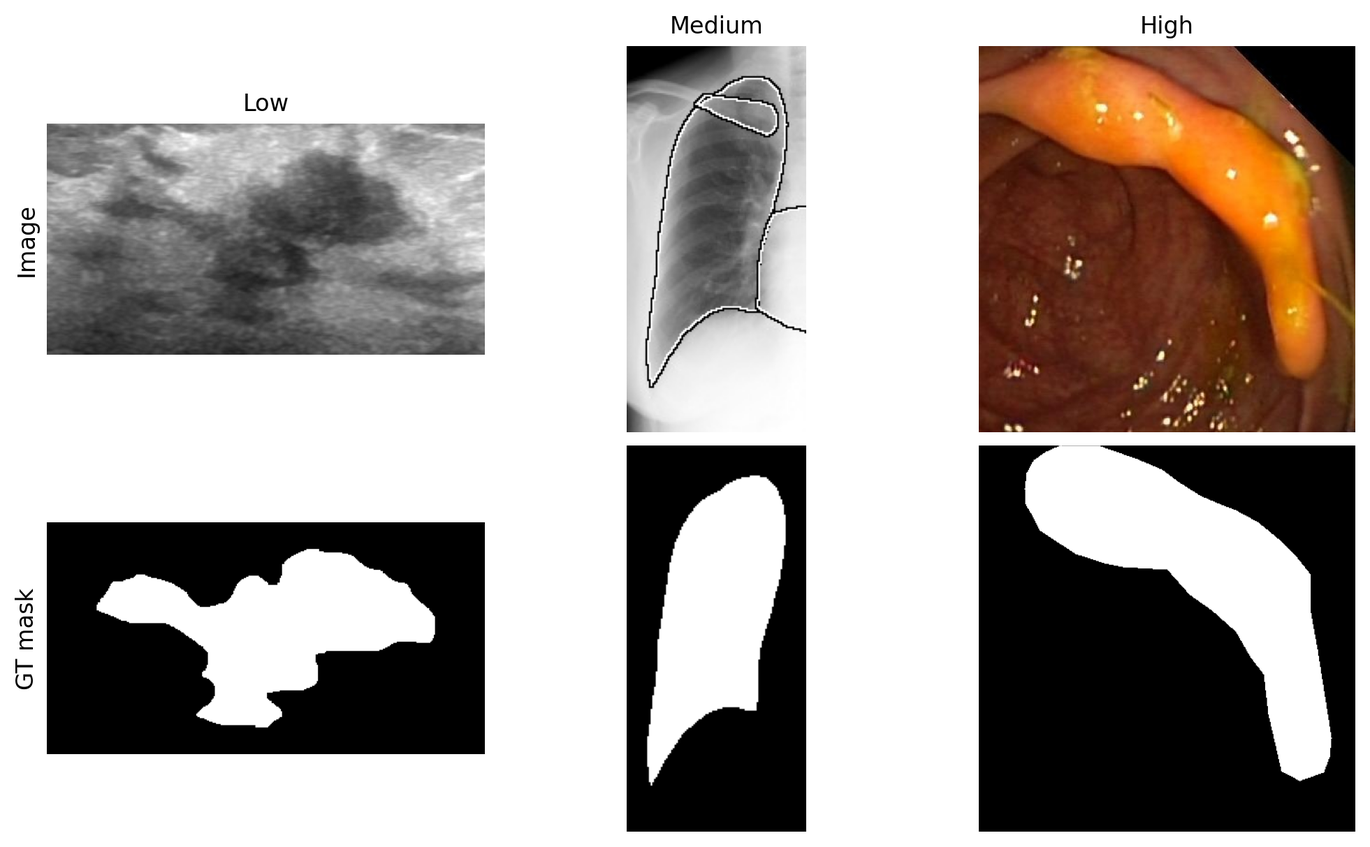}}
\par\vspace{0.75em}
\subfloat[Thinness: left-to-right examples for Compact, Moderate, and Thin. Top row shows the medical image; bottom row shows the corresponding ground-truth mask.\label{fig:strata_thinness_examples}]{\includegraphics[width=0.49\textwidth]{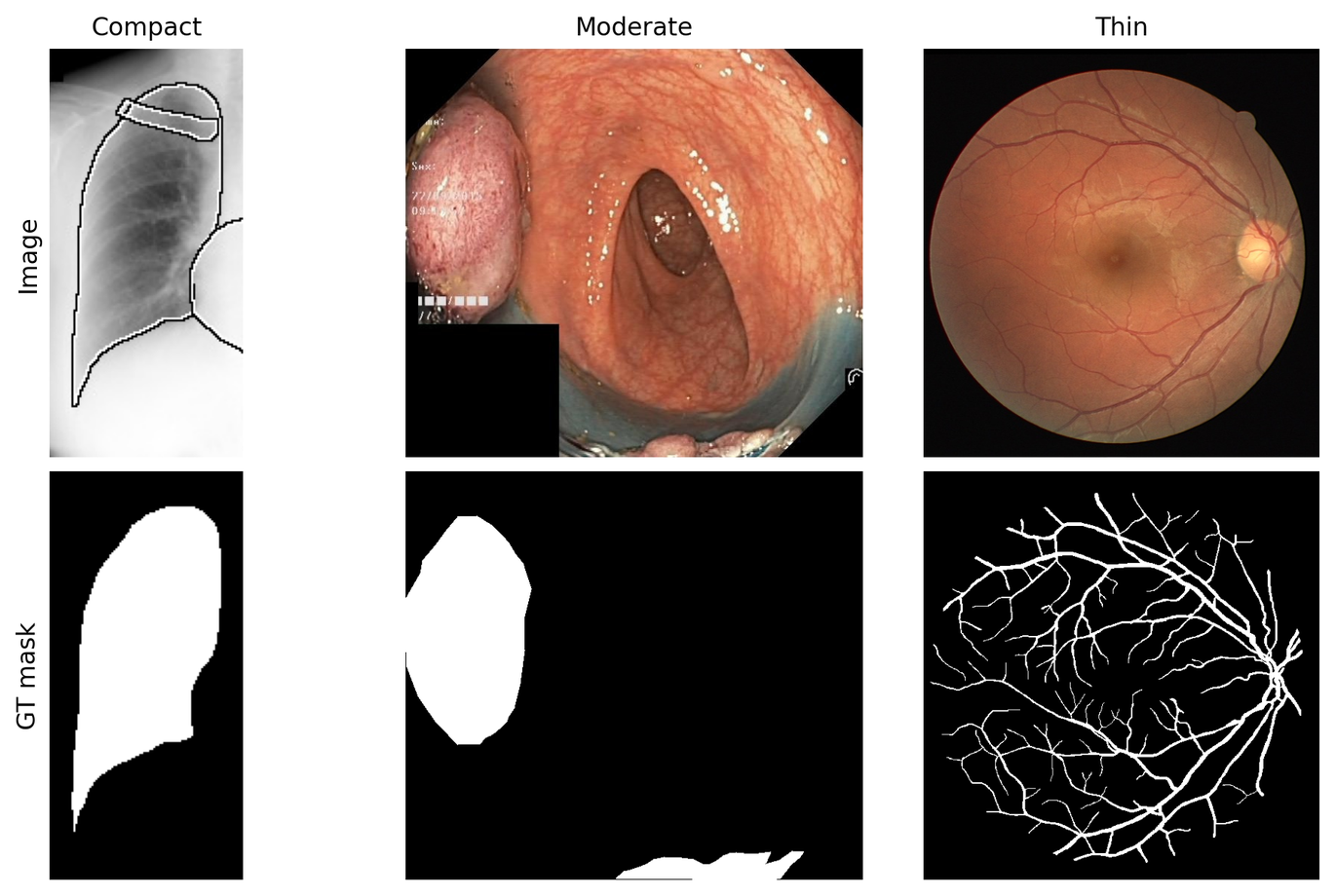}}
\hfill
\subfloat[Object Size: left-to-right examples for Medium, Small, and Large. Top row shows the medical image; bottom row shows the corresponding ground-truth mask.\label{fig:strata_object_size_examples}]{\includegraphics[width=0.49\textwidth]{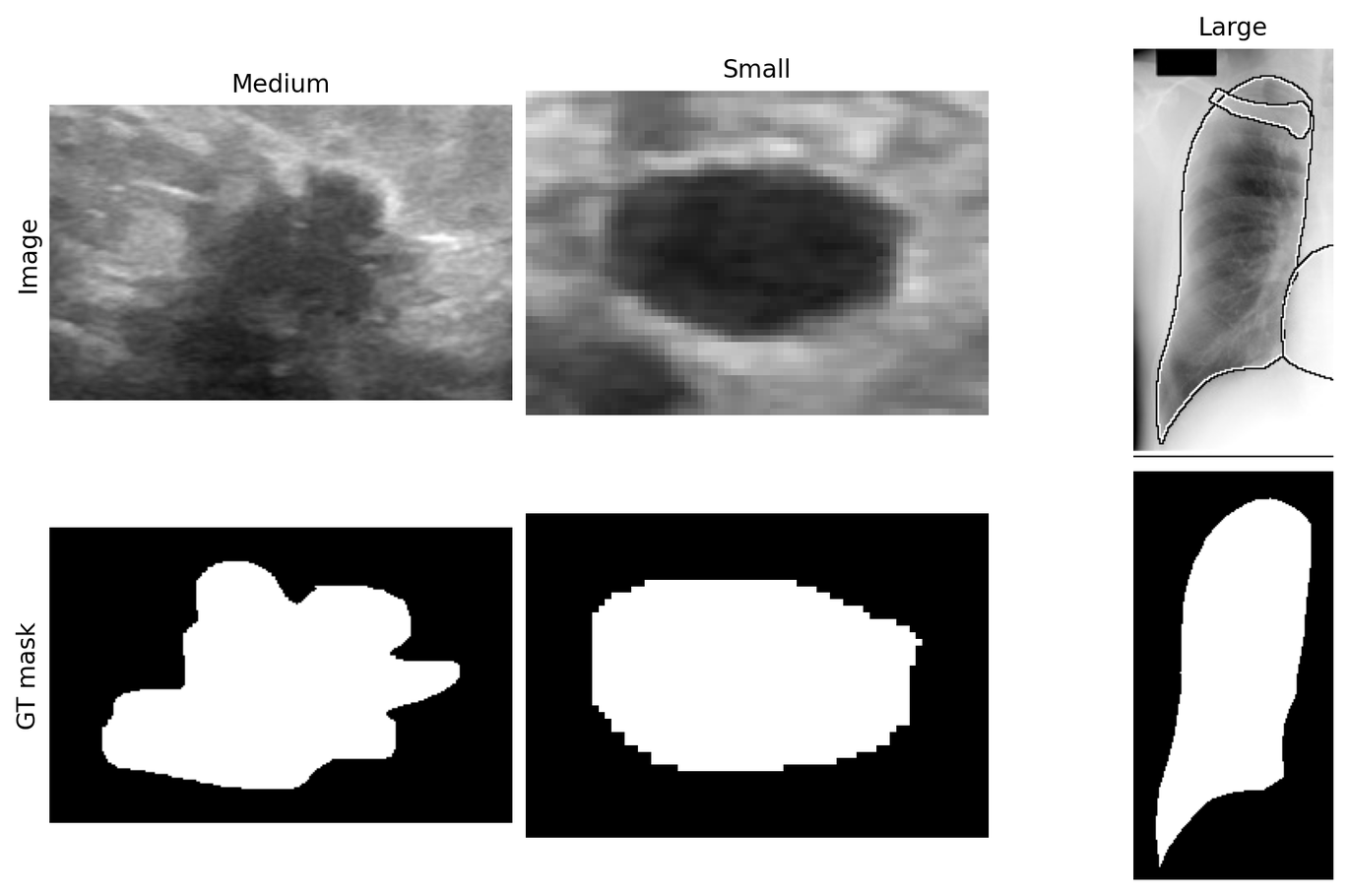}}
\par\vspace{0.75em}
\caption{Representative medical examples for each morphological stratum level used in the strata analysis. For each family, columns correspond to the listed levels, with the raw image on the top row and the ground-truth mask on the bottom row.}
\label{fig:medical_strata_examples}
\end{figure*}

Fig.~\ref{fig:medical_strata_examples} shows representative medical examples for each morphology stratum used in the strata analysis. For each stratum family, the columns correspond to the three discrete levels used in the main experiments, with the raw image shown on the top row and the corresponding ground-truth mask shown on the bottom row. These examples are included to make the operational meaning of the strata labels visually explicit. In particular, the boundary-complexity examples illustrate the difference between smooth, moderately irregular, and highly intricate contours; the contrast examples show how foreground-background separability varies across datasets; the thinness examples distinguish compact objects from elongated thin structures; and the object-size examples illustrate the range from small localized targets to large anatomical regions.

These qualitative references provide context for interpreting Table~\ref{tab:strata_dominance}. The strata analysis is not intended as a semantic taxonomy of datasets, but as a morphology-oriented grouping of samples according to geometric and appearance properties that are known to affect interactive segmentation difficulty. The examples therefore serve as a visual anchor for the gain-based stratum summaries reported in the appendix and help clarify why certain acquisition strategies dominate in some strata but not others.

\section{Per-Dataset Results}
\label{app:per-dataset-results}

\begin{table*}[!t]
\centering
\caption{Per-dataset IoU endpoint improvement (step 0 to full budget, direction-corrected) for all strategies across the five medical datasets. The pooled mean weights datasets proportionally to test-set image count (BUSI 48.6\%, CheXmask 19.0\%, ISIC 2018 15.4\%, Kvasir-SEG 15.4\%, DRIVE 1.5\%). The equal-weight mean treats each dataset equally regardless of size. CheXmask yields negative improvement for every strategy, likely because SAM already achieves IoU@0 of 0.909 from the first prompt, leaving little room for further gain; additional prompts appear to degrade the already near-optimal segmentation of chest X-ray anatomy. BALD-MCD$^*$ denotes the properly trained dropout head. Bold = best non-GT strategy per row.}
\label{tab:per_dataset_iou}
\small
\setlength{\tabcolsep}{6pt}
\renewcommand{\arraystretch}{1.1}
\begin{tabular}{@{}lc cccccc !{\color{black!20}\vrule} c@{}}
\toprule
Dataset & $n$ & BALD-Lap. & BALD-MCD & BALD-MCD$^*$ & Entropy & Random & MaxDist & GT-based \\
\midrule
BUSI        & 630  & \bfseries +0.129 & +0.073 & $-0.119$ & $-0.059$ & $-0.167$ & $-0.203$ & $-0.012$ \\
CheXmask    & 247  & $-0.133$         & $-0.344$ & $-0.087$ & $\bm{-0.006}$ & $-0.156$ & $-0.269$ & $-0.072$ \\
DRIVE       & 20   & $-0.002$ & +0.000 & \bfseries +0.001 & +0.000 & $-0.010$ & $-0.009$ & $-0.013$ \\
ISIC 2018    & 200  & \bfseries +0.079 & +0.047 & $-0.124$ & +0.005 & +0.068 & $-0.061$ & +0.084 \\
Kvasir-SEG   & 200  & \bfseries +0.207 & +0.141 & +0.048 & +0.180 & +0.179 & +0.006 & +0.230 \\
\midrule
Pooled      & 1297 & \bfseries +0.081 & $-0.001$ & $-0.086$ & $-0.001$ & $-0.073$ & $-0.158$ & +0.029 \\
Equal-wt.   & ---  & \bfseries +0.056 & $-0.016$ & $-0.056$ & +0.024 & $-0.017$ & $-0.107$ & +0.043 \\
\bottomrule
\end{tabular}
\vspace{4pt}

\begin{tabular}{@{}lc cccccc@{}}
\toprule
\multicolumn{8}{l}{\textit{Per-dataset ECE endpoint improvement (direction-corrected, higher is better)}} \\
\midrule
Dataset & $n$ & BALD-Lap. & BALD-MCD & BALD-MCD$^*$ & Entropy & Random & MaxDist \\
\midrule
BUSI        & 630  & \bfseries +0.156 & +0.095 & +0.002 & +0.041 & +0.029 & +0.015 \\
CheXmask    & 247  & $-0.065$         & $-0.106$ & $-0.033$ & $\bm{-0.015}$ & $-0.083$ & $-0.090$ \\
DRIVE       & 20   & \bfseries +0.180 & +0.033 & +0.075 & +0.087 & +0.177 & +0.136 \\
ISIC 2018    & 200  & \bfseries +0.110 & +0.052 & $-0.046$ & +0.054 & +0.100 & +0.061 \\
Kvasir-SEG   & 200  & \bfseries +0.089 & +0.054 & $-0.001$ & +0.059 & +0.083 & +0.015 \\
\midrule
Pooled      & 1297 & \bfseries +0.097 & +0.043 & $-0.011$ & +0.036 & +0.029 & +0.004 \\
\bottomrule
\end{tabular}
\end{table*}

\begin{table*}[!t]
\centering
\caption{Per-dataset BALD Laplace final IoU and endpoint improvement for SAM1, SAM2, and SAM3. All SAM versions use the same head trained on SAM1-precomputed logits. IoU@0 is the mean IoU after the first prompt, placed at the ground-truth foreground centroid (before BALD starts). $\Delta$ is endpoint improvement from IoU@0 to final IoU at full budget. Bold = largest $\Delta$ per dataset.}
\label{tab:per_dataset_sam123}
\small
\setlength{\tabcolsep}{5pt}
\renewcommand{\arraystretch}{1.1}
\begin{tabular}{@{}lc cc cc cc@{}}
\toprule
& & \multicolumn{2}{c}{SAM1} & \multicolumn{2}{c}{SAM2} & \multicolumn{2}{c}{SAM3} \\
\cmidrule(lr){3-4}\cmidrule(lr){5-6}\cmidrule(lr){7-8}
Dataset & $n$ & IoU@0 & $\Delta$ & IoU@0 & $\Delta$ & IoU@0 & $\Delta$ \\
\midrule
BUSI        & 630  & 0.431 & \bfseries +0.091 & 0.667 & +0.000 & 0.651 & $-0.239$ \\
CheXmask    & 247  & 0.909 & $-0.077$         & 0.823 & \bfseries $-0.032$ & 0.901 & $-0.510$ \\
DRIVE       & 20   & 0.124 & $-0.002$         & 0.090 & +0.009 & 0.103 & \bfseries +0.022 \\
ISIC 2018    & 200  & 0.511 & \bfseries +0.084 & 0.638 & +0.037 & 0.711 & $-0.151$ \\
Kvasir-SEG   & 200  & 0.461 & \bfseries +0.287 & 0.669 & +0.088 & 0.774 & $-0.352$ \\
\midrule
Overall     & 1297 & 0.534 & \bfseries +0.087 & 0.684 & +0.013 & 0.718 & $-0.291$ \\
\bottomrule
\end{tabular}
\end{table*}

Table~\ref{tab:per_dataset_iou} shows per-dataset IoU endpoint improvement. The pooled mean weights datasets by test-set image count (BUSI 48.6\%, CheXmask 19.0\%, ISIC 2018 15.4\%, Kvasir-SEG 15.4\%, DRIVE 1.5\%). BALD Laplace has positive improvement on three of five datasets (BUSI $+0.129$, ISIC 2018 $+0.079$, Kvasir-SEG $+0.207$) and is the best non-GT strategy on each of them. CheXmask is negative for every strategy (BALD Laplace $-0.133$, Random $-0.156$, BALD-MCD $-0.344$). SAM achieves IoU@0 of 0.909 on CheXmask from the first prompt; the initial segmentation is already near-optimal, and additional prompts appear to reduce rather than improve mask quality on chest X-ray anatomy, consistent with a performance-ceiling effect rather than a BALD-specific failure. BALD-MCD$^*$ achieves positive IoU improvement only on DRIVE ($+0.001$, the best non-GT value on that dataset) and Kvasir-SEG ($+0.048$), with negative improvement on the remaining three datasets. The equal-weight dataset mean is $+0.056$ for BALD Laplace versus $-0.016$ for BALD-MCD, $-0.056$ for BALD-MCD$^*$, $+0.024$ for Entropy, $-0.017$ for Random, and $-0.107$ for MaxDist, so BALD Laplace leads under both pooled and equal-weight aggregation. Table~\ref{tab:per_dataset_sam123} shows the per-dataset SAM version breakdown; the SAM3 degradation is concentrated in BUSI ($-0.239$), CheXmask ($-0.510$), and Kvasir-SEG ($-0.352$), while DRIVE is the only dataset where SAM3 gains ($+0.022$).

\begin{figure*}[!t]
\centering
\subfloat[Good case: Baseball\_bat/sample\_000376 (final IoU $=0.941$).]{\includegraphics[width=\textwidth]{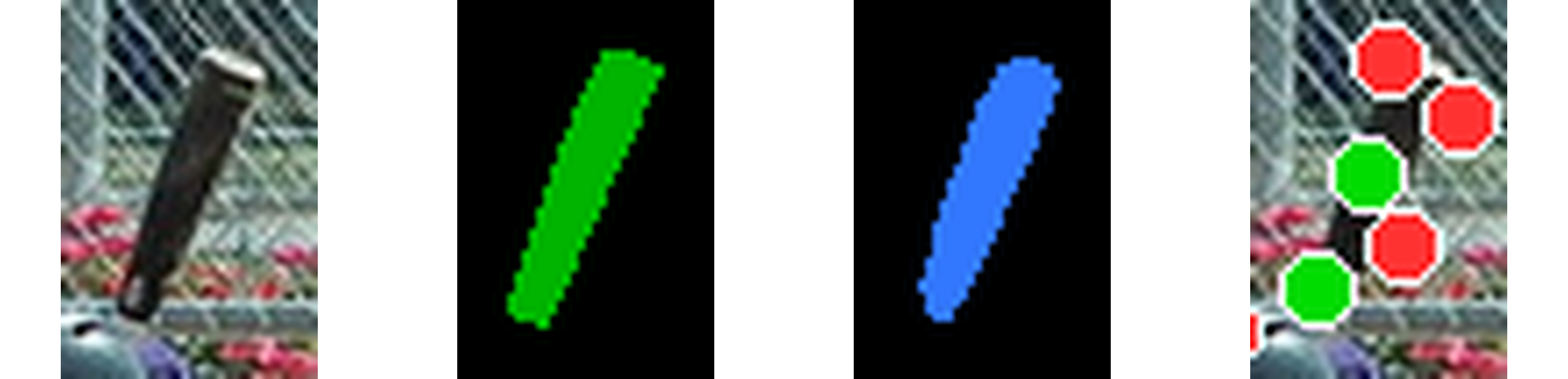}}
\par\vspace{0.5em}
\subfloat[Good case: Stop\_sign/sample\_000046 (final IoU $=0.932$).]{\includegraphics[width=\textwidth]{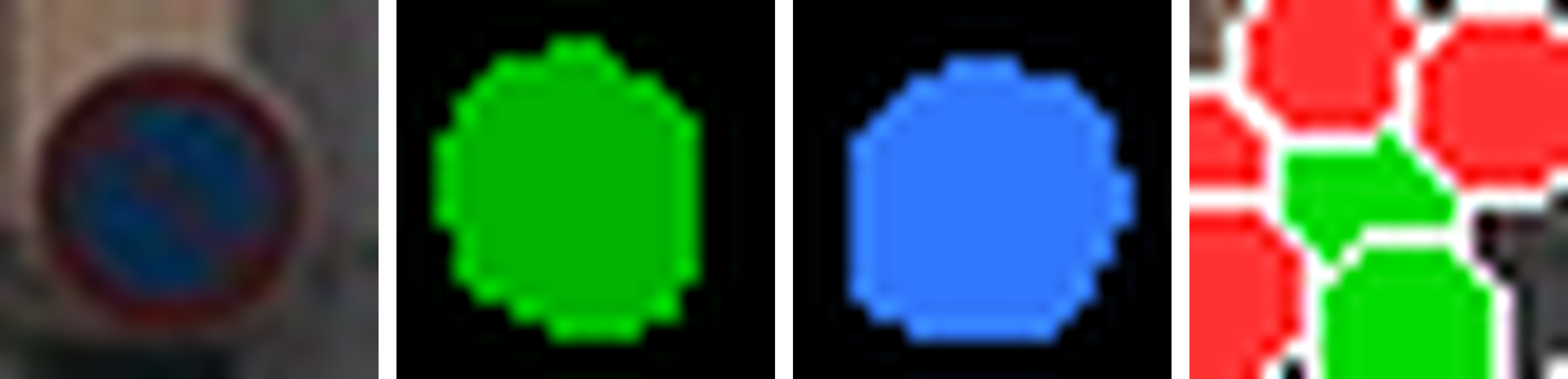}}
\par\vspace{0.5em}
\subfloat[Failure case: Clock/sample\_000204 (final IoU $=0.094$).]{\includegraphics[width=\textwidth]{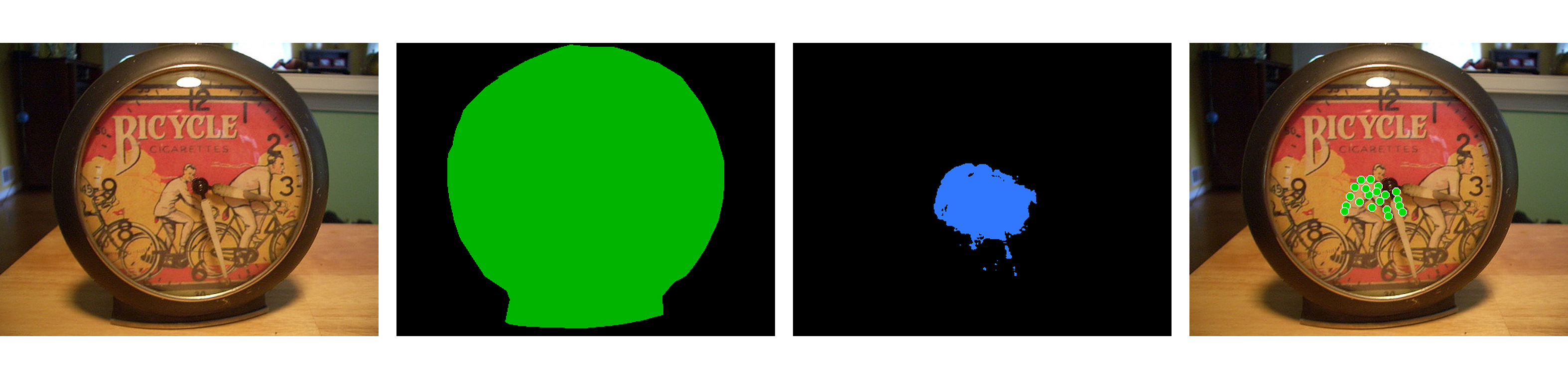}}
\par\vspace{0.5em}
\subfloat[Failure case: Bus/sample\_000092 (final IoU $=0.022$).]{\includegraphics[width=\textwidth]{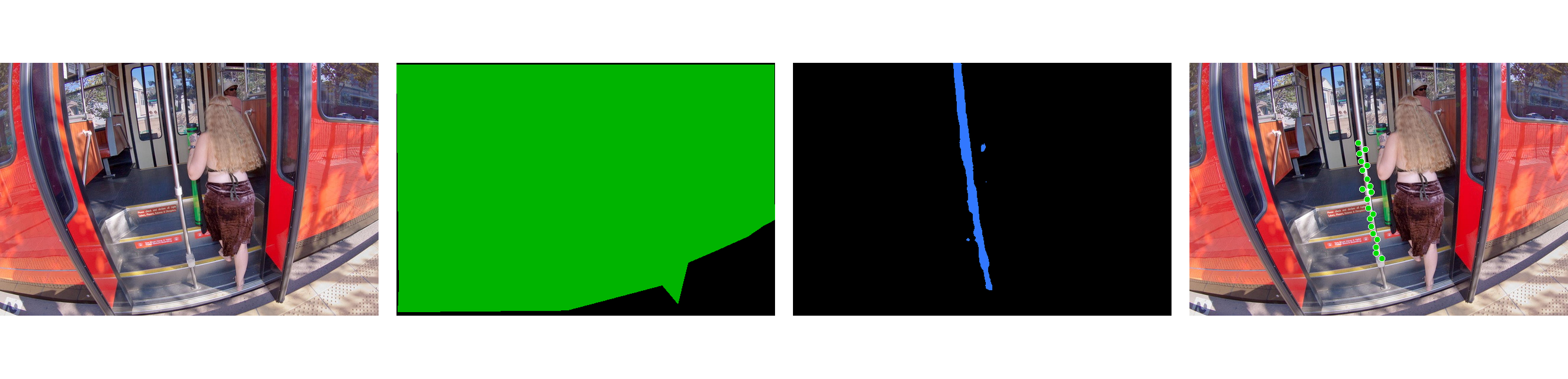}}
\caption{Representative PointPrompt active-prompting examples. Two successful cases and two failure cases are shown. From left to right, the columns show the original image crop, the ground-truth mask, the final BALD-Laplace prediction, and the final prompt set, with foreground prompts shown in green and background prompts shown in red.}
\label{fig:pointprompt_qualitative_reconstructed}
\end{figure*}

\section{Qualitative Results: PointPrompt Examples}
\label{subsec:qualitative-pointprompt}

The PointPrompt examples in Fig.~\ref{fig:pointprompt_qualitative_reconstructed} show the same underlying principle, but with a stronger dependence on object morphology. For thin or compact natural-image objects, the acquisition process often succeeds not by densely sampling the object interior, but by combining a small number of positive anchors with many negative exclusion points around the object boundary. This behavior is clear in the baseball bat example (Fig.~\ref{fig:pointprompt_qualitative_reconstructed}, top row), where the thin bat is defined by a few foreground prompts on the object itself and multiple negative prompts placed around it. The final prediction is successful because the prompt set separates the thin structure from the surrounding clutter. In the stop sign example, the prompts surround the sign rather than collapsing into the center, which allows the model to recover the outer octagonal extent instead of only an interior patch.

The failures are visually intuitive and especially informative. In the clock example, the prompts anchor onto the textured central region of the clock face rather than the outer circular boundary, so the model explains the ambiguous interior detail instead of segmenting the full object. In the bus example, the prompt sequence collapses onto the central vertical bar inside the scene, which is a narrow but visually salient structure. As a result, the interaction process becomes trapped on that bar rather than spreading outward to define the much larger bus region. These failures suggest that the main weakness of active prompting is not simply inaccurate uncertainty estimation in the abstract, but spatial mislocalization of the acquisition rule onto a locally salient yet globally uninformative structure.

\begin{figure*}[!t]
\centering
\subfloat[Good case: BUSI/benign\_benign\_390 (final IoU $=0.950$).]{\includegraphics[width=\textwidth]{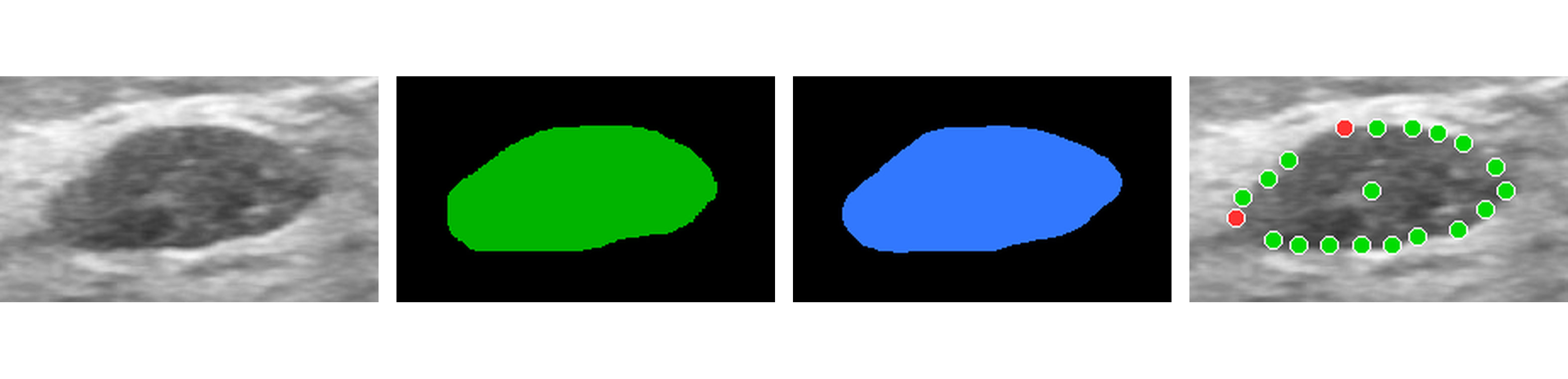}}
\par\vspace{0.5em}
\subfloat[Good case: KvasirSEG/cju2hewssldzx0835ep795xu0 (final IoU $=0.939$).]{\includegraphics[width=\textwidth]{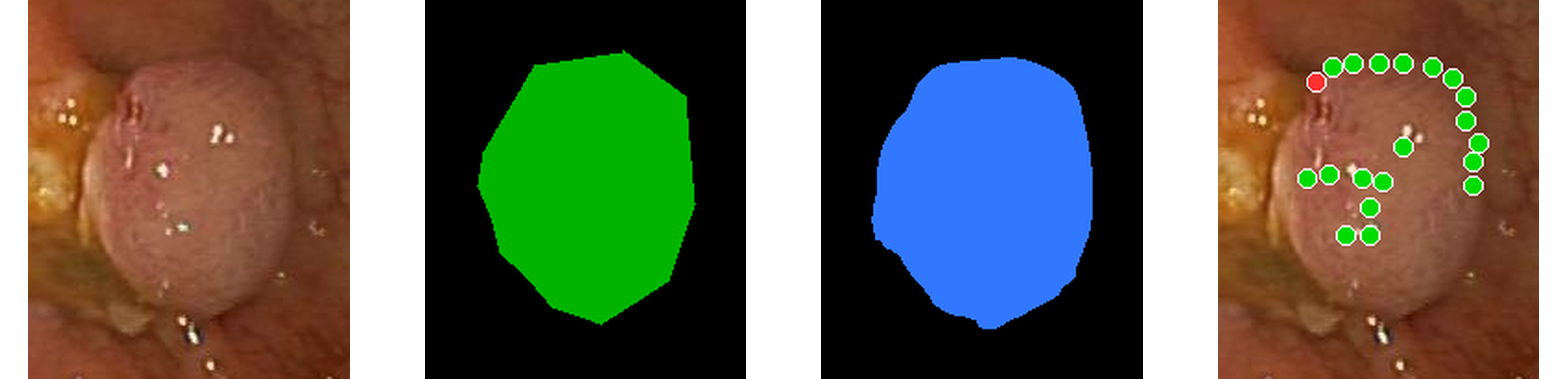}}
\par\vspace{0.5em}
\subfloat[Failure case: ISIC2018/ISIC\_0014784 (final IoU $=0.013$).]{\includegraphics[width=\textwidth]{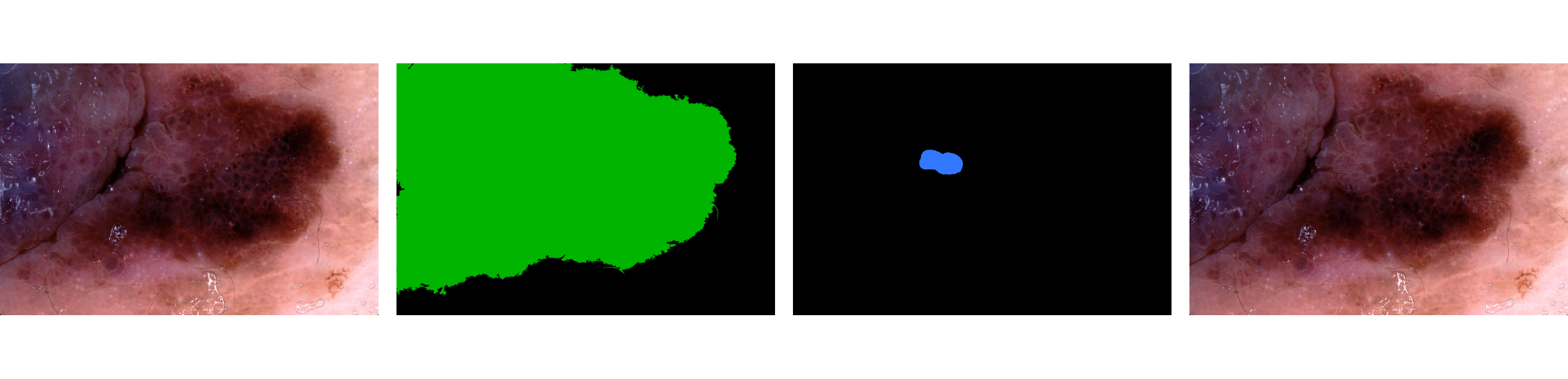}}
\par\vspace{0.5em}
\subfloat[Failure case: CheXmask/JPCLN113 (final IoU $=0.576$).]{\includegraphics[width=\textwidth]{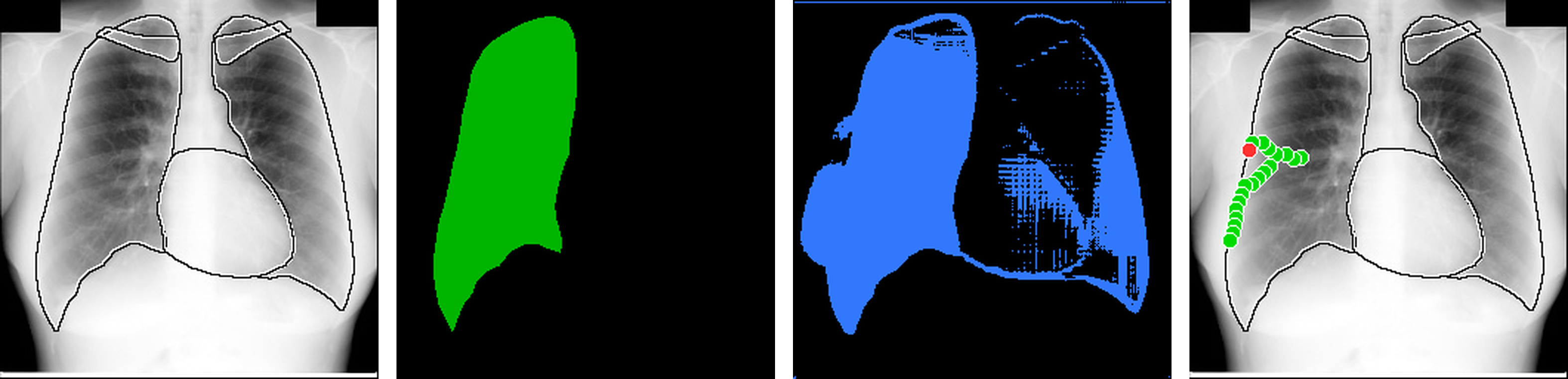}}
\caption{Representative medical active-prompting examples on baseline SAM1. Two successful cases and two failure cases are shown. From left to right, the columns show the original image crop, the ground-truth mask, the final BALD-Laplace prediction, and the final prompt set, with foreground prompts shown in green and background prompts shown in red.}
\label{fig:medical_qualitative_reconstructed}
\end{figure*}

\section{Qualitative Results: Medical Examples}
\label{subsec:qualitative-medical}

The medical examples in Fig.~\ref{fig:medical_qualitative_reconstructed} illustrate both the geometric strengths and the failure modes of uncertainty-guided prompting. In the successful cases, the prompt set provides broad spatial coverage of the target rather than collapsing to a single interior region. In the BUSI benign lesion example (Fig.~\ref{fig:medical_qualitative_reconstructed}, top row), the positive prompts trace the lesion perimeter almost circumferentially, whereas the small number of negative prompts provide local corrective context outside the mass. This effectively brackets the boundary from multiple directions, so the final prediction recovers the lesion shape with high fidelity. A similar pattern appears in the Kvasir-SEG polyp example, where the prompts distribute around the polyp contour instead of concentrating on only one side of the structure. In both cases, the final mask is visually consistent with the target because the prompt geometry constrains the object extent rather than merely confirming a local interior region.

The failure cases expose the opposite behavior. In the ISIC 2018 dermoscopy example, the prompts remain anchored in the lesion interior and fail to bracket the outer contour, so the model converges to a small central fragment rather than expanding to the full lesion extent. In the CheXmask chest X-ray example, the prompts emphasize only part of the thoracic structure and do not adequately constrain the opposite side or the surrounding anatomical boundary, producing a one-sided and underconstrained segmentation. The visual lesson from these medical cases is that successful prompting is associated with spatially distributed boundary support, whereas failure occurs when the prompt trajectory stays confined to a partial or locally ambiguous subregion of the object.

\section{Overall Visual Conclusions}
\label{subsec:qualitative-overall}

Taken together, the qualitative results show that active prompting succeeds when uncertainty leads to geometrically complementary prompts that constrain the object from multiple sides. In the best cases, the prompts either wrap around the target boundary or combine inclusion and exclusion cues in a way that exposes the full object extent. In the failure cases, the prompt sequence collapses onto redundant interior evidence, one-sided anatomical structure, or distracting high-salience image elements such as the clock center or the bus's central bar. The examples therefore suggest that uncertainty is most useful when it is spatially organized around the true object geometry, and least useful when it repeatedly reinforces a partial or misleading local cue.

\section*{Acknowledgment}
This work was supported by the ML4Seismic Industry Partners at the Georgia Institute of Technology.

\bibliographystyle{IEEEtran}
\bibliography{References/IEEEtran_control,References/references}

\begin{thebibliography}{10}
\providecommand{\url}[1]{#1}
\csname url@samestyle\endcsname
\providecommand{\newblock}{\relax}
\providecommand{\bibinfo}[2]{#2}
\providecommand{\BIBentrySTDinterwordspacing}{\spaceskip=0pt\relax}
\providecommand{\BIBentryALTinterwordstretchfactor}{4}
\providecommand{\BIBentryALTinterwordspacing}{\spaceskip=\fontdimen2\font plus
\BIBentryALTinterwordstretchfactor\fontdimen3\font minus
  \fontdimen4\font\relax}
\providecommand{\BIBforeignlanguage}[2]{{%
\expandafter\ifx\csname l@#1\endcsname\relax
\typeout{** WARNING: IEEEtran.bst: No hyphenation pattern has been}%
\typeout{** loaded for the language `#1'. Using the pattern for}%
\typeout{** the default language instead.}%
\else
\language=\csname l@#1\endcsname
\fi
#2}}
\providecommand{\BIBdecl}{\relax}
\BIBdecl

\bibitem{wang2018interactive}
G.~Wang \emph{et~al.}, ``{Interactive medical image segmentation using deep
  learning with image-specific fine tuning},'' \emph{IEEE Trans. Med. Imag.},
  vol.~37, no.~7, pp. 1562--1573, Jul. 2018, doi: 10.1109/TMI.2018.2791721.

\bibitem{waldeland2018convolutional}
A.~U. Waldeland, A.~C. Jensen, L.-J. Gelius, and A.~H.~S. Solberg,
  ``{Convolutional neural networks for automated seismic interpretation},''
  \emph{Leading Edge}, vol.~37, no.~7, pp. 529--537, Jul. 2018, doi:
  10.1190/tle37070529.1.

\bibitem{quesada2025large}
J.~Quesada \emph{et~al.}, ``{A large-scale benchmark on geological fault
  delineation models: Domain shift, training dynamics, generalizability,
  evaluation, and inferential behavior},'' \emph{IEEE Access}, vol.~13, pp.
  215\,110--215\,131, 2025, doi: 10.1109/ACCESS.2025.3643616.

\bibitem{pedersen2019detection}
M.~Pedersen, J.~B. Haurum, R.~Gade, and T.~B. Moeslund, ``{Detection of marine
  animals in a new underwater dataset with varying visibility},'' in
  \emph{Proc. IEEE/CVF Conf. Comput. Vis. Pattern Recognit. Workshops (CVPRW)},
  Long Beach, CA, USA, 2019, pp. 18--26.

\bibitem{lin2014microsoft}
T.-Y. Lin \emph{et~al.}, ``{Microsoft COCO: Common objects in context},'' in
  \emph{Proc. Eur. Conf. Comput. Vis. (ECCV)}, Zurich, Switzerland, 2014, pp.
  740--755, doi: 10.1007/978-3-319-10602-1\_48.

\bibitem{kirillov2023segment}
A.~Kirillov \emph{et~al.}, ``{Segment anything},'' in \emph{Proc. IEEE/CVF Int.
  Conf. Comput. Vis. (ICCV)}, Paris, France, 2023, pp. 3992--4003, doi:
  10.1109/ICCV51070.2023.00371.

\bibitem{wu2025medical}
J.~Wu \emph{et~al.}, ``{Medical SAM adapter: Adapting segment anything model
  for medical image segmentation},'' \emph{Med. Image Anal.}, vol. 102, May
  2025, {A}rt. no. 103547, doi: 10.1016/j.media.2025.103547.

\bibitem{ren2024segment}
S.~Ren \emph{et~al.}, ``{Segment anything, from space?}'' in \emph{Proc.
  IEEE/CVF Winter Conf. Appl. Comput. Vis. (WACV)}, Waikoloa, HI, USA, 2024,
  pp. 8340--8350, doi: 10.1109/WACV57701.2024.00817.

\bibitem{wang2024sam}
Z.~Peng, Z.~Xu, Z.~Zeng, X.~Yang, and W.~Shen, ``{SAM-PARSER: Fine-tuning SAM
  efficiently by parameter space reconstruction},'' in \emph{Proc. AAAI Conf.
  Artif. Intell.}, vol.~38, no.~5, Vancouver, BC, Canada, 2024, pp. 4515--4523,
  doi: 10.1609/aaai.v38i5.28250.

\bibitem{liu2024grounding}
S.~Liu \emph{et~al.}, ``{Grounding DINO: Marrying DINO with grounded
  pre-training for open-set object detection},'' in \emph{Proc. Eur. Conf.
  Comput. Vis. (ECCV)}, Milan, Italy, 2024, pp. 38--55, doi:
  10.1007/978-3-031-72970-6\_3.

\bibitem{gu2023systematic}
J.~Gu \emph{et~al.}, ``{A systematic survey of prompt engineering on
  vision-language foundation models},'' arXiv:2307.12980, 2023.

\bibitem{huang2024alignsam}
D.~Huang \emph{et~al.}, ``{AlignSAM: Aligning segment anything model to open
  context via reinforcement learning},'' in \emph{Proc. IEEE/CVF Conf. Comput.
  Vis. Pattern Recognit. (CVPR)}, Seattle, WA, USA, 2024, pp. 3205--3215, doi:
  10.1109/CVPR52733.2024.00309.

\bibitem{szeto2017click}
R.~Szeto and J.~J. Corso, ``{Click here: Human-localized keypoints as guidance
  for viewpoint estimation},'' in \emph{Proc. IEEE Int. Conf. Comput. Vis.
  (ICCV)}, Venice, Italy, 2017, pp. 1604--1613, doi: 10.1109/ICCV.2017.177.

\bibitem{quesada2024pointprompt}
J.~Quesada, M.~Alotaibi, M.~Prabhushankar, and G.~AlRegib, ``{PointPrompt: A
  multi-modal prompting dataset for segment anything model},'' in \emph{Proc.
  IEEE/CVF Conf. Comput. Vis. Pattern Recognit. Workshops (CVPRW)}, Seattle,
  WA, USA, 2024, pp. 1604--1610, doi: 10.1109/CVPRW63382.2024.00167.

\bibitem{settles2009active}
B.~Settles, ``{Active learning literature survey},'' Dept. Comput. Sci., Univ.
  Wisconsin--Madison, Madison, WI, USA, Comput. Sci. Tech. Rep. 1648, 2009.

\bibitem{gal2017deep}
Y.~Gal, R.~Islam, and Z.~Ghahramani, ``{Deep Bayesian active learning with
  image data},'' in \emph{Proc. 34th Int. Conf. Mach. Learn. (ICML)}, Sydney,
  NSW, Australia, 2017, pp. 1183--1192.

\bibitem{sener2017active}
O.~Sener and S.~Savarese, ``{Active learning for convolutional neural networks:
  A core-set approach},'' in \emph{Proc. Int. Conf. Learn. Represent. (ICLR)},
  Vancouver, BC, Canada, 2018.

\bibitem{benkert2023gaussian}
R.~Benkert, M.~Prabhushankar, G.~AlRegib, A.~Pacharmi, and E.~Corona,
  ``{Gaussian switch sampling: A second-order approach to active learning},''
  \emph{IEEE Trans. Artif. Intell.}, vol.~5, no.~1, pp. 38--50, Jan. 2024, doi:
  10.1109/TAI.2023.3246959.

\bibitem{benkert2024effective}
R.~Benkert, M.~Prabhushankar, and G.~AlRegib, ``{Effective data selection for
  seismic interpretation through disagreement},'' \emph{IEEE Trans. Geosci.
  Remote Sens.}, vol.~62, pp. 1--12, 2024, doi: 10.1109/TGRS.2024.3409476.

\bibitem{long2015fully}
J.~Long, E.~Shelhamer, and T.~Darrell, ``{Fully convolutional networks for
  semantic segmentation},'' in \emph{Proc. IEEE/CVF Conf. Comput. Vis. Pattern
  Recognit. (CVPR)}, Boston, MA, USA, 2015, pp. 3431--3440, doi:
  10.1109/CVPR.2015.7298965.

\bibitem{ronneberger2015u}
O.~Ronneberger, P.~Fischer, and T.~Brox, ``{U-Net: Convolutional networks for
  biomedical image segmentation},'' in \emph{Proc. Int. Conf. Med. Image
  Comput. Comput.-Assist. Intervent. (MICCAI)}, Munich, Germany, 2015, pp.
  234--241, doi: 10.1007/978-3-319-24574-4\_28.

\bibitem{chen2017rethinking}
L.-C. Chen, G.~Papandreou, F.~Schroff, and H.~Adam, ``{Rethinking atrous
  convolution for semantic image segmentation},'' arXiv:1706.05587, 2017.

\bibitem{zhao2017pyramid}
H.~Zhao, J.~Shi, X.~Qi, X.~Wang, and J.~Jia, ``{Pyramid scene parsing
  network},'' in \emph{Proc. IEEE/CVF Conf. Comput. Vis. Pattern Recognit.
  (CVPR)}, Honolulu, HI, USA, 2017, pp. 6230--6239, doi: 10.1109/CVPR.2017.660.

\bibitem{everingham2010pascal}
M.~Everingham, L.~Van~Gool, C.~K. Williams, J.~Winn, and A.~Zisserman, ``{The
  PASCAL visual object classes (VOC) challenge},'' \emph{Int. J. Comput. Vis.},
  vol.~88, no.~2, pp. 303--338, Jun. 2010, doi: 10.1007/s11263-009-0275-4.

\bibitem{zhou2017scene}
B.~Zhou, H.~Zhao, X.~Puig, S.~Fidler, A.~Barriuso, and A.~Torralba, ``{Scene
  parsing through ADE20K dataset},'' in \emph{Proc. IEEE/CVF Conf. Comput. Vis.
  Pattern Recognit. (CVPR)}, Honolulu, HI, USA, 2017, pp. 5122--5130, doi:
  10.1109/CVPR.2017.544.

\bibitem{isensee2021nnu}
F.~Isensee, P.~F. Jaeger, S.~A. Kohl, J.~Petersen, and K.~H. Maier-Hein,
  ``{nnU-Net: A self-configuring method for deep learning-based biomedical
  image segmentation},'' \emph{Nature Methods}, vol.~18, no.~2, pp. 203--211,
  Feb. 2021, doi: 10.1038/s41592-020-01008-z.

\bibitem{yuan2021multi}
Y.~Yuan, L.~Zhang, L.~Wang, and H.~Huang, ``{Multi-level attention network for
  retinal vessel segmentation},'' \emph{IEEE J. Biomed. Health Inform.},
  vol.~26, no.~1, pp. 312--323, Jan. 2022, doi: 10.1109/JBHI.2021.3089201.

\bibitem{wu2019faultseg3d}
X.~Wu, L.~Liang, Y.~Shi, and S.~Fomel, ``{FaultSeg3D: Using synthetic data sets
  to train an end-to-end convolutional neural network for 3D seismic fault
  segmentation},'' \emph{Geophysics}, vol.~84, no.~3, pp. IM35--IM45, May 2019,
  doi: 10.1190/geo2018-0646.1.

\bibitem{wu2020building}
X.~Wu, Z.~Geng, Y.~Shi, N.~Pham, S.~Fomel, and G.~Caumon, ``{Building realistic
  structure models to train convolutional neural networks for seismic
  structural interpretation},'' \emph{Geophysics}, vol.~85, no.~4, pp.
  WA27--WA39, Jul. 2020, doi: 10.1190/geo2019-0375.1.

\bibitem{kemker2018algorithms}
R.~Kemker, C.~Salvaggio, and C.~Kanan, ``{Algorithms for semantic segmentation
  of multispectral remote sensing imagery using deep learning},'' \emph{ISPRS
  J. Photogramm. Remote Sens.}, vol. 145, pp. 60--77, Nov. 2018, doi:
  10.1016/j.isprsjprs.2018.04.014.

\bibitem{maggiori2016convolutional}
E.~Maggiori, Y.~Tarabalka, G.~Charpiat, and P.~Alliez, ``{Convolutional neural
  networks for large-scale remote-sensing image classification},'' \emph{IEEE
  Trans. Geosci. Remote Sens.}, vol.~55, no.~2, pp. 645--657, Feb. 2017, doi:
  10.1109/TGRS.2016.2612821.

\bibitem{zou2023segment}
X.~Zou \emph{et~al.}, ``{Segment everything everywhere all at once},'' in
  \emph{Proc. Adv. Neural Inf. Process. Syst. (NeurIPS)}, vol.~36, New Orleans,
  LA, USA, 2023, pp. 19\,769--19\,782, doi: 10.52202/075280-0868.

\bibitem{li2024segment}
F.~Li \emph{et~al.}, ``{Segment and recognize anything at any granularity},''
  in \emph{Proc. Eur. Conf. Comput. Vis. (ECCV)}, Milan, Italy, 2024, pp.
  467--484, doi: 10.1007/978-3-031-73195-2\_27.

\bibitem{wang2023seggpt}
X.~Wang, X.~Zhang, Y.~Cao, W.~Wang, C.~Shen, and T.~Huang, ``{SegGPT:
  Segmenting everything in context},'' in \emph{Proc. IEEE/CVF Int. Conf.
  Comput. Vis. (ICCV)}, Paris, France, 2023, pp. 1130--1140, doi:
  10.1109/ICCV51070.2023.00110.

\bibitem{dosovitskiy2020image}
A.~Dosovitskiy \emph{et~al.}, ``{An image is worth $16\times16$ words:
  Transformers for image recognition at scale},'' in \emph{Proc. Int. Conf.
  Learn. Represent. (ICLR)}, 2021.

\bibitem{ma2024segment}
J.~Ma, Y.~He, F.~Li, L.~Han, C.~You, and B.~Wang, ``{Segment anything in
  medical images},'' \emph{Nature Commun.}, vol.~15, no.~1, Jan. 2024, {A}rt.
  no. 654, doi: 10.1038/s41467-024-44824-z.

\bibitem{zhang2023customized}
K.~Zhang and D.~Liu, ``{Customized segment anything model for medical image
  segmentation},'' arXiv:2304.13785, 2023.

\bibitem{huang2024segment}
Y.~Huang \emph{et~al.}, ``{Segment anything model for medical images?}''
  \emph{Med. Image Anal.}, vol.~92, Feb. 2024, {A}rt. no. 103061, doi:
  10.1016/j.media.2023.103061.

\bibitem{zhang2024rs}
E.~Zhang \emph{et~al.}, ``{RS-SAM: Integrating multi-scale information for
  enhanced remote sensing image segmentation},'' in \emph{Proc. Asian Conf.
  Comput. Vis. (ACCV)}, Hanoi, Vietnam, 2024, pp. 994--1010.

\bibitem{li2019seismic}
S.~Li, C.~Yang, H.~Sun, and H.~Zhang, ``{Seismic fault detection using an
  encoder--decoder convolutional neural network with a small training set},''
  \emph{J. Geophys. Eng.}, vol.~16, no.~1, pp. 175--189, Feb. 2019, doi:
  10.1093/jge/gxy015.

\bibitem{ravi2024sam}
N.~Ravi \emph{et~al.}, ``{SAM 2: Segment anything in images and videos},'' in
  \emph{Proc. Int. Conf. Learn. Represent. (ICLR)}, Singapore, 2025.

\bibitem{carion2025sam}
N.~Carion \emph{et~al.}, ``{SAM 3: Segment anything with concepts},'' in
  \emph{Proc. Int. Conf. Learn. Represent. (ICLR)}, Rio de Janeiro, Brazil,
  2026.

\bibitem{wei2022chain}
J.~Wei \emph{et~al.}, ``{Chain-of-thought prompting elicits reasoning in large
  language models},'' in \emph{Proc. Adv. Neural Inf. Process. Syst.
  (NeurIPS)}, vol.~35, New Orleans, LA, USA, 2022, pp. 24\,824--24\,837, doi:
  10.52202/068431-1800.

\bibitem{brown2020language}
T.~Brown \emph{et~al.}, ``{Language models are few-shot learners},'' in
  \emph{Proc. Adv. Neural Inf. Process. Syst. (NeurIPS)}, vol.~33, 2020, pp.
  1877--1901.

\bibitem{ouyang2022training}
L.~Ouyang \emph{et~al.}, ``{Training language models to follow instructions
  with human feedback},'' in \emph{Proc. Adv. Neural Inf. Process. Syst.
  (NeurIPS)}, vol.~35, New Orleans, LA, USA, 2022, pp. 27\,730--27\,744, doi:
  10.52202/068431-2011.

\bibitem{rose2023visual}
D.~Rose \emph{et~al.}, ``{Visual chain of thought: Bridging logical gaps with
  multimodal infillings},'' arXiv:2305.02317, 2023.

\bibitem{brooks2023instructpix2pix}
T.~Brooks, A.~Holynski, and A.~A. Efros, ``{InstructPix2Pix: Learning to follow
  image editing instructions},'' in \emph{Proc. IEEE/CVF Conf. Comput. Vis.
  Pattern Recognit. (CVPR)}, Vancouver, BC, Canada, 2023, pp. 18\,392--18\,402,
  doi: 10.1109/CVPR52729.2023.01764.

\bibitem{diao2024active}
S.~Diao, P.~Wang, Y.~Lin, R.~Pan, X.~Liu, and T.~Zhang, ``{Active prompting
  with chain-of-thought for large language models},'' in \emph{Proc. 62nd Annu.
  Meeting Assoc. Comput. Linguistics (ACL)}, Bangkok, Thailand, 2024, pp.
  1330--1350, doi: 10.18653/v1/2024.acl-long.73.

\bibitem{zhang2023personalize}
R.~Zhang \emph{et~al.}, ``{Personalize segment anything model with one shot},''
  in \emph{Proc. Int. Conf. Learn. Represent. (ICLR)}, Vienna, Austria, 2024.

\bibitem{ren2024grounded}
T.~Ren \emph{et~al.}, ``{Grounded SAM: Assembling open-world models for diverse
  visual tasks},'' arXiv:2401.14159, 2024.

\bibitem{liu2023matcher}
Y.~Liu, M.~Zhu, H.~Li, H.~Chen, X.~Wang, and C.~Shen, ``{Matcher: Segment
  anything with one shot using all-purpose feature matching},'' in \emph{Proc.
  Int. Conf. Learn. Represent. (ICLR)}, Vienna, Austria, 2024.

\bibitem{chen2024rsprompter}
K.~Chen \emph{et~al.}, ``{RSPrompter: Learning to prompt for remote sensing
  instance segmentation based on visual foundation model},'' \emph{IEEE Trans.
  Geosci. Remote Sens.}, vol.~62, pp. 1--17, 2024, doi:
  10.1109/TGRS.2024.3356074.

\bibitem{zhang2023segment}
Y.~Zhang and R.~Jiao, ``{How segment anything model (SAM) boost medical image
  segmentation: A survey},'' SSRN preprint 4495221, 2023, doi:
  10.2139/ssrn.4495221.

\bibitem{mazurowski2023segment}
M.~A. Mazurowski, H.~Dong, H.~Gu, J.~Yang, N.~Konz, and Y.~Zhang, ``{Segment
  anything model for medical image analysis: An experimental study},''
  \emph{Med. Image Anal.}, vol.~89, Oct. 2023, {A}rt. no. 102918, doi:
  10.1016/j.media.2023.102918.

\bibitem{li2024polyp}
Y.~Li, M.~Hu, and X.~Yang, ``{Polyp-SAM: Transfer SAM for polyp
  segmentation},'' in \emph{Proc. SPIE Med. Imag.: Comput.-Aided Diagnosis},
  vol. 12927, San Diego, CA, USA, 2024, {A}rt. no. 1292735, doi:
  10.1117/12.3006809.

\bibitem{wang2024empirical}
Y.~Wang, Y.~Zhao, and L.~Petzold, ``{An empirical study on the robustness of
  the segment anything model (SAM)},'' \emph{Pattern Recognit.}, vol. 155, Nov.
  2024, {A}rt. no. 110685, doi: 10.1016/j.patcog.2024.110685.

\bibitem{shan2023robustness}
X.~Shan and C.~Zhang, ``{Robustness of segment anything model (SAM) for
  autonomous driving in adverse weather conditions},'' arXiv:2306.13290, 2023.

\bibitem{zhou2023can}
T.~Zhou, Y.~Zhang, Y.~Zhou, Y.~Wu, and C.~Gong, ``{Can SAM segment polyps?}''
  arXiv:2304.07583, 2023.

\bibitem{scheffer2001active}
T.~Scheffer, C.~Decomain, and S.~Wrobel, ``{Active hidden Markov models for
  information extraction},'' in \emph{Proc. Int. Symp. Intell. Data Anal.
  (IDA)}, Cascais, Portugal, 2001, pp. 309--318, doi:
  10.1007/3-540-44816-0\_31.

\bibitem{shannon1948mathematical}
C.~E. Shannon, ``{A mathematical theory of communication},'' \emph{Bell Syst.
  Tech. J.}, vol.~27, no.~3, pp. 379--423, Jul. 1948, doi:
  10.1002/j.1538-7305.1948.tb01338.x.

\bibitem{seung1992query}
H.~S. Seung, M.~Opper, and H.~Sompolinsky, ``{Query by committee},'' in
  \emph{Proc. 5th Annu. Workshop Comput. Learn. Theory (COLT)}, Pittsburgh, PA,
  USA, 1992, pp. 287--294, doi: 10.1145/130385.130417.

\bibitem{benkert2024targeting}
R.~Benkert, M.~Prabhushankar, and G.~AlRegib, ``{Targeting negative flips in
  active learning using validation sets},'' in \emph{Proc. IEEE Int. Conf. Big
  Data (BigData)}, Washington, DC, USA, 2024, pp. 820--829, doi:
  10.1109/BigData62323.2024.10825140.

\bibitem{ash2019deep}
J.~T. Ash, C.~Zhang, A.~Krishnamurthy, J.~Langford, and A.~Agarwal, ``{Deep
  batch active learning by diverse, uncertain gradient lower bounds},'' in
  \emph{Proc. Int. Conf. Learn. Represent. (ICLR)}, 2020.

\bibitem{prabhushankar2022introspective}
M.~Prabhushankar and G.~AlRegib, ``{Introspective learning: A two-stage
  approach for inference in neural networks},'' in \emph{Proc. Adv. Neural Inf.
  Process. Syst. (NeurIPS)}, vol.~35, New Orleans, LA, USA, 2022, pp.
  12\,126--12\,140, doi: 10.52202/068431-0881.

\bibitem{kokilepersaud2023focal}
K.~Kokilepersaud \emph{et~al.}, ``{FOCAL: A cost-aware video dataset for active
  learning},'' in \emph{Proc. IEEE Int. Conf. Big Data (BigData)}, Sorrento,
  Italy, 2023, pp. 1269--1278, doi: 10.1109/BigData59044.2023.10386242.

\bibitem{fowler2023clinical}
Z.~Fowler, K.~P. Kokilepersaud, M.~Prabhushankar, and G.~AlRegib, ``{Clinical
  trial active learning},'' in \emph{Proc. 14th ACM Int. Conf. Bioinf., Comput.
  Biol., Health Informat. (BCB)}, Houston, TX, USA, 2023, pp. 1--10, doi:
  10.1145/3584371.3612961.

\bibitem{houlsby2011bayesian}
N.~Houlsby, F.~Husz{\'a}r, Z.~Ghahramani, and M.~Lengyel, ``{Bayesian active
  learning for classification and preference learning},'' arXiv:1112.5745,
  2011.

\bibitem{gal2016dropout}
Y.~Gal and Z.~Ghahramani, ``{Dropout as a Bayesian approximation: Representing
  model uncertainty in deep learning},'' in \emph{Proc. 33rd Int. Conf. Mach.
  Learn. (ICML)}, New York, NY, USA, 2016, pp. 1050--1059.

\bibitem{beluch2018power}
W.~H. Beluch, T.~Genewein, A.~N{\"u}rnberger, and J.~M. K{\"o}hler, ``{The
  power of ensembles for active learning in image classification},'' in
  \emph{Proc. IEEE/CVF Conf. Comput. Vis. Pattern Recognit. (CVPR)}, Salt Lake
  City, UT, USA, 2018, pp. 9368--9377, doi: 10.1109/CVPR.2018.00976.

\bibitem{ritter2018scalable}
H.~Ritter, A.~Botev, and D.~Barber, ``{A scalable Laplace approximation for
  neural networks},'' in \emph{Proc. Int. Conf. Learn. Represent. (ICLR)},
  Vancouver, BC, Canada, 2018.

\bibitem{mackowiak2018cereals}
R.~Mackowiak, P.~Lenz, O.~Ghori, F.~Diego, O.~Lange, and C.~Rother,
  ``{CEREALS---Cost-effective region-based active learning for semantic
  segmentation},'' in \emph{Proc. Brit. Mach. Vis. Conf. (BMVC)}, Newcastle,
  U.K., 2018.

\bibitem{kao2018localization}
C.-C. Kao, T.-Y. Lee, P.~Sen, and M.-Y. Liu, ``{Localization-aware active
  learning for object detection},'' in \emph{Proc. Asian Conf. Comput. Vis.
  (ACCV)}, Perth, WA, Australia, 2018, pp. 506--522, doi:
  10.1007/978-3-030-20876-9\_32.

\bibitem{yang2017suggestive}
L.~Yang, Y.~Zhang, J.~Chen, S.~Zhang, and D.~Z. Chen, ``{Suggestive annotation:
  A deep active learning framework for biomedical image segmentation},'' in
  \emph{Proc. Int. Conf. Med. Image Comput. Comput.-Assist. Intervent.
  (MICCAI)}, Quebec City, QC, Canada, 2017, pp. 399--407, doi:
  10.1007/978-3-319-66179-7\_46.

\bibitem{trotter2020ndd20}
C.~Trotter \emph{et~al.}, ``{NDD20: A large-scale few-shot dolphin dataset for
  coarse and fine-grained categorisation},'' arXiv:2005.13359, 2020.

\bibitem{al2020dataset}
W.~Al-Dhabyani, M.~Gomaa, H.~Khaled, and A.~Fahmy, ``{Dataset of breast
  ultrasound images},'' \emph{Data Brief}, vol.~28, Feb. 2020, {A}rt. no.
  104863, doi: 10.1016/j.dib.2019.104863.

\bibitem{jha2020kvasir}
D.~Jha \emph{et~al.}, ``{Kvasir-SEG: A segmented polyp dataset},'' in
  \emph{Proc. Int. Conf. MultiMedia Model. (MMM)}, Daejeon, South Korea, 2020,
  pp. 451--462, doi: 10.1007/978-3-030-37734-2\_37.

\bibitem{codella2019skin}
N.~Codella \emph{et~al.}, ``{Skin lesion analysis toward melanoma detection
  2018: A challenge hosted by the International Skin Imaging Collaboration
  (ISIC)},'' arXiv:1902.03368, 2019.

\bibitem{alaudah2019machine}
Y.~Alaudah, P.~Micha{\l}owicz, M.~Alfarraj, and G.~AlRegib, ``{A
  machine-learning benchmark for facies classification},''
  \emph{Interpretation}, vol.~7, no.~3, pp. SE175--SE187, Aug. 2019, doi:
  10.1190/INT-2018-0249.1.

\bibitem{gaggion2024chexmask}
N.~Gaggion \emph{et~al.}, ``{CheXmask: A large-scale dataset of anatomical
  segmentation masks for multi-center chest X-ray images},'' \emph{Sci. Data},
  vol.~11, no.~1, May 2024, {A}rt. no. 511, doi: 10.1038/s41597-024-03358-1.

\bibitem{staal2004ridge}
J.~Staal, M.~D. Abr{\`a}moff, M.~Niemeijer, M.~A. Viergever, and B.~van
  Ginneken, ``{Ridge-based vessel segmentation in color images of the
  retina},'' \emph{IEEE Trans. Med. Imag.}, vol.~23, no.~4, pp. 501--509, Apr.
  2004, doi: 10.1109/TMI.2004.825627.

\bibitem{tschandl2018ham10000}
P.~Tschandl, C.~Rosendahl, and H.~Kittler, ``{The HAM10000 dataset, a large
  collection of multi-source dermatoscopic images of common pigmented skin
  lesions},'' \emph{Sci. Data}, vol.~5, no.~1, Aug. 2018, {A}rt. no. 180161,
  doi: 10.1038/sdata.2018.161.

\bibitem{chakrabarty2025comparing}
S.~Chakrabarty and R.~Soni, ``{Comparing SAM 2 and SAM 3 for zero-shot
  segmentation of 3D medical data},'' arXiv:2511.21926, 2025.

\bibitem{sengupta2025sam}
S.~Sengupta, S.~Chakrabarty, and R.~Soni, ``{Is SAM 2 better than SAM in
  medical image segmentation?}'' in \emph{Proc. SPIE Med. Imag.: Image
  Process.}, vol. 13406, San Diego, CA, USA, 2025, {A}rt. no. 134062N, doi:
  10.1117/12.3047370.

\bibitem{ma2025medsam2}
J.~Ma \emph{et~al.}, ``{MedSAM2: Segment anything in 3D medical images and
  videos},'' arXiv:2504.03600, 2025.

\end{thebibliography}

\begin{IEEEbiography}[{\includegraphics[width=1in,height=1.25in,clip,keepaspectratio]{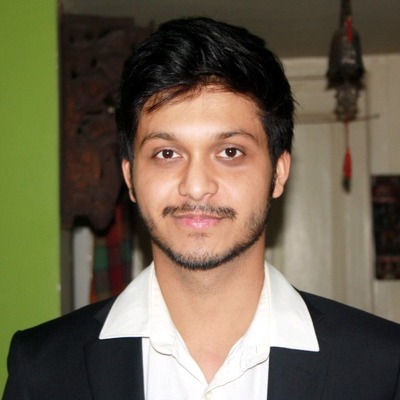}}]{Prithwijit Chowdhury} (Student Member, IEEE) received the B.Tech. degree from KIIT University, Bhubaneswar, India, in 2020. He joined the School of Electrical and Computer Engineering, Georgia Institute of Technology, Atlanta, GA, USA, as an M.S. student in 2021, where he is currently pursuing the Ph.D. degree. He is a Researcher with the Center for Energy and Geo Processing (CeGP) and a member of the Omni Lab for Intelligent Visual Engineering and Science (OLIVES). His research interests include digital signal and image processing and machine learning with applications to geophysics. He has presented several works at the IMAGE conference and NeurIPS workshops and has published in \emph{Geophysics}.
\end{IEEEbiography}

\begin{IEEEbiography}[{\includegraphics[width=1in,height=1.25in,clip,keepaspectratio]{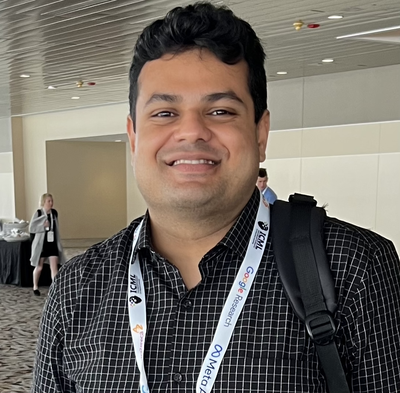}}]{Mohit Prabhushankar} (Member, IEEE) received the Ph.D. degree in electrical engineering from the Georgia Institute of Technology (Georgia Tech), Atlanta, GA, USA, in 2021. He is currently a Postdoctoral Research Fellow with the School of Electrical and Computer Engineering, Georgia Institute of Technology, in the Omni Lab for Intelligent Visual Engineering and Science (OLIVES). His research interests include image processing, machine learning, active learning, healthcare, and robust and explainable AI. He was a recipient of the Best Paper Award at ICIP 2019 and the Top Viewed Special Session Paper Award at ICIP 2020. He also received the ECE Outstanding Graduate Teaching Award, the CSIP Research Award, and the Roger P. Webb ECE Graduate Research Assistant Excellence Award, all in 2022. He has delivered short courses and tutorials at IEEE IV'23, ICIP'23, BigData'23, WACV'24, and AAAI'24.
\end{IEEEbiography}

\begin{IEEEbiography}[{\includegraphics[width=1in,height=1.25in,clip,keepaspectratio]{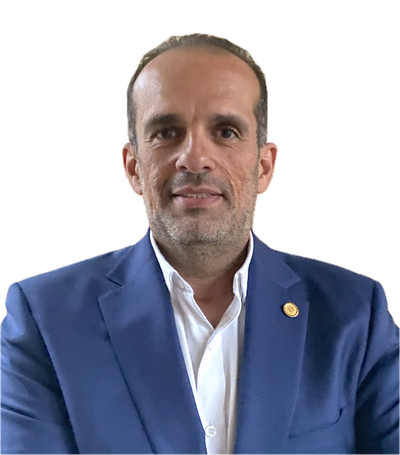}}]{Ghassan AlRegib} (Fellow, IEEE) is currently the John and Marilu McCarty Chair Professor with the School of Electrical and Computer Engineering, Georgia Institute of Technology, Atlanta, GA, USA. In the Omni Lab for Intelligent Visual Engineering and Science (OLIVES), he and his group work on robust and interpretable machine learning algorithms, uncertainty and trust, and human-in-the-loop algorithms. The group has demonstrated its work on a wide range of applications, such as autonomous systems, medical imaging, and subsurface imaging. The group is interested in advancing the fundamentals as well as the deployment of such systems in real-world scenarios. He has been issued several U.S. patents and invention disclosures. He is a Fellow of the IEEE. Prof. AlRegib is active in the IEEE. He served on the editorial board of several transactions and served as the TPC Chair for ICIP 2020, ICIP 2024, and GlobalSIP 2014. He was an Area Editor of \emph{IEEE Signal Processing Magazine}. In 2008, he received the ECE Outstanding Junior Faculty Member Award. In 2017, he received the Denning Faculty Award for Global Engagement. He received the 2024 ECE Distinguished Faculty Achievement Award at Georgia Tech. He and his students received the Best Paper Award at ICIP 2019 and the 2023 EURASIP Best Paper Award for \emph{Image Communication}. In addition, one of their papers was the Best Paper Runner-Up at BigData 2024. In 2024, he co-founded the AI Makerspace at Georgia Tech, where any student and any community member can access and utilize AI regardless of their background. 
\end{IEEEbiography}

\EOD
\vfill
\end{document}